\documentclass{article}

 \usepackage[preprint]{neurips_2026}

\usepackage[utf8]{inputenc} % allow utf-8 input
\usepackage[T1]{fontenc}    % use 8-bit T1 fonts
\usepackage{hyperref}       % hyperlinks
\usepackage{url}            % simple URL typesetting
\usepackage{booktabs}       % professional-quality tables
\usepackage{amsfonts}       % blackboard math symbols
\usepackage{nicefrac}       % compact symbols for 1/2, etc.
\usepackage{microtype}      % microtypography
\usepackage[utf8]{inputenc} % allow utf-8 input
\usepackage[T1]{fontenc}    % use 8-bit T1 fonts
\usepackage{hyperref}       % hyperlinks
\hypersetup{
    colorlinks=true,
    citecolor=green,
    urlcolor=blue,
}
\usepackage{url}            % simple URL typesetting
\usepackage{booktabs}       % professional-quality tables
\usepackage{amsfonts}       % blackboard math symbols
\usepackage{nicefrac}       % compact symbols for 1/2, etc.
\usepackage{microtype}      % microtypography
\usepackage{xcolor}         % colors
\usepackage{graphicx}
\usepackage{amsmath}
\usepackage{xcolor}         % colors
\usepackage{algorithm}
\usepackage{algpseudocode}
\usepackage{amssymb}
\usepackage{colortbl}
\usepackage{fontawesome}
\usepackage{xspace}
\usepackage{pifont}
\usepackage{multirow}
\usepackage{enumitem}
\usepackage{listings}
\usepackage{caption}
\usepackage[most]{tcolorbox}
\newcommand{\bench}{Edit-Compass}
\newcommand{\rmbench}{EditReward-Compass}

\definecolor{blue1}{HTML}{196ab1}
\definecolor{blue2}{HTML}{4886c1}
\definecolor{blue3}{HTML}{5e9bd6}
\definecolor{blue4}{HTML}{77b1e2}
\definecolor{blue5}{HTML}{bdd930}
\definecolor{blue6}{HTML}{dfebf6}

\definecolor{red1}{HTML}{de512c}
\definecolor{red2}{HTML}{f2642d}
\definecolor{red3}{HTML}{f68f58}
\definecolor{red4}{HTML}{febf92}
\definecolor{red5}{HTML}{f8e9c8}
\definecolor{darkorange}{rgb}{1.0, 0.55, 0.0}
\definecolor{forestgreen}{rgb}{0.0, 0.5, 0.0}
\definecolor{ashgrey}{rgb}{0.7, 0.75, 0.71}
\newcommand{\icono}{\textcolor{ashgrey}{\faTimesCircle}\xspace}
\newcommand{\icoyes}{\textcolor{forestgreen}{\faCheckCircle}\xspace}
\newcommand{\icohalf}{\textcolor{darkorange}{\ding{51}\kern-0.65em\ding{55}}}
\newcommand{\icohalfno}{\ding{51}\kern-0.65em\ding{55}}
\newcommand{\gain}[1]{\textcolor{green!50!black}{\small$\uparrow$#1}}

\newtcolorbox[auto counter]{promptbox}[2][]{
  colback=gray!5,
  colframe=gray!40,
  title=\textbf{Box~\thetcbcounter: #2},
  fonttitle=\bfseries,
  breakable,
  sharp corners,
  boxrule=0.8pt,
  #1
}
\title{\bench~\&~\rmbench: A Unified Benchmark for Image Editing and Reward Modeling}

\author{%
    Xuehai Bai$^{1}$\thanks{Equal Contribution} \,\,\,
    Yang Shi$^{2,3}$\footnotemark[1] \,\,\,
    Yi-Fan Zhang$^{4}$\footnotemark[1]\,\,\thanks{Project Lead} \,\,\,
    Xuanyu Zhu$^{2}$ \,
    Yuran Wang$^{2}$ \,
    \\
    \bfseries
    Yifan Dai$^{3}$ \,
    Xinyu Liu$^{3}$ \,
    Yiyan Ji$^{3}$ \,
    Xiaoling Gu$^{1}$\thanks{Corresponding Author} \,\,\,
    Yuanxing Zhang$^{3}$\footnotemark[3] \,\,\,
    \\ 
    $^1$HDU\quad
    $^2$PKU\quad
    $^3$Kling Team\quad
    $^4$CASIA\quad
    \\
    {\centering}
    \url{https://github.com/bxhsort/Edit-Compass-and-EditReward-Compass}
}

\begin{document}

\maketitle

\begin{abstract}
Recent image editing models have achieved remarkable progress in instruction following, multimodal understanding, and complex visual editing. However, existing benchmarks often fail to faithfully reflect human judgment, especially for strong frontier models, due to limited task difficulty and coarse-grained evaluation protocols. In parallel, reward models have become increasingly important for RL-based image editing optimization, yet existing reward model benchmarks still rely on unrealistic evaluation settings that deviate from practical RL scenarios. These limitations hinder reliable assessment of both image editing models and reward models.
To address these challenges, we introduce \textbf{{\bench}} and \textbf{{\rmbench}}, a unified evaluation suite for image editing and reward modeling. {\bench} contains $2,388$ carefully annotated instances spanning six progressively challenging task categories, covering capabilities such as world knowledge reasoning, visual reasoning, and multi-image editing. Beyond broad task coverage, {\bench} adopts a fine-grained multidimensional evaluation framework based on structured reasoning and carefully designed scoring rubrics. In parallel, {\rmbench} contains $2,251$ preference pairs that simulate realistic reward modeling scenarios during RL optimization.
We conduct extensive evaluations on $29$ frontier image editing models and $21$ reward models. The results reveal a substantial gap between proprietary and open-source systems, while also exposing persistent weaknesses in world knowledge understanding, visual reasoning, and multi-image editing. Moreover, native multimodal large language models outperform existing open-source reward models, including models explicitly trained on preference data. Overall, our benchmark suite provides a comprehensive and human-aligned framework for evaluating frontier image editing systems and reward models.
\end{abstract}

\section{Introduction}
\begin{table}[t]
\caption{\textbf{Comparison between \bench\ and existing image editing benchmarks.} 
\texttt{AVR} denotes Algorithm Visual Reasoning, \texttt{MIA} denotes Multi-Image Awareness, \texttt{WKR} denotes World Knowledge Reasoning, \texttt{DM} denotes Dynamic Manipulation, and \texttt{HP} denotes Human Preference evaluation.}
\vspace{0.5em}
\label{tab:compare_image_editing_bench}
\centering
\resizebox{ 0.95\linewidth}{!}{%
\begin{tabular}{lcccccccc}
\toprule
\textbf{Benchmark} & \textbf{Size} & \textbf{Tasks} & \textbf{Lang.} & \textbf{AVR} & \textbf{MIA} & \textbf{WKR} & \textbf{DM}& \textbf{HP} \\
\midrule

Magicbrush~\cite{zhang2023magicbrush} & 1,053 & 5 & EN & \icono & \icono &\icono & \icono &  Low \\
Emu Edit~\cite{cui2025emu3} & 3,055 & 7 & EN & \icono & \icono & \icono & \icono  & Low \\
AnyEdit-Bench~\cite{yu2025anyedit} & 1,250 & 25 & EN & \icono & \icono & \icono & \icono & Low \\
ImgEdit-Bench~\cite{ye2025imgedit} & 811 & 11 & EN & \icono & \icono & \icono & \icono & Mid \\
GEdit-Bench~\cite{liu2025step1x} & 606 & 14 & EN/ZH & \icono & \icono & \icono & \icono & Mid \\
ICE-Bench~\cite{pan2025ice} & 6,538 & 31 & EN & \icono & \icono & \icono & \icono & Low \\
RISE-Bench\cite{zhao2025envisioning} & 360 & 4 & EN & \icono & \icono & \icohalf & \icono  & Mid \\
UniREditBench~\cite{han2025unireditbench} & 2,700 & 13 & EN & \icono & \icono & \icono & \icohalf  & Mid \\
WiseEdit~\cite{pan2025wiseedit} & 1,220 & 13 & EN/ZH & \icono & \icohalf  & \icoyes & \icono  & Mid \\
\midrule
\bench~(Ours) & 2,388 & 36 & EN/ZH & \icoyes & \icoyes & \icoyes & \icoyes  & High \\
\bottomrule
\end{tabular}%
}
% \vspace{-1.70em}
\end{table}

Recent image editing models~\cite{brooks2023instructpix2pix, chen2025opengpt,labs2025flux, wang2025scone,tong2026cof,zhu2026layermultilayerrepresentationfusion,zhao2024ultraedit,bai2026think,yu2025anyedit} have achieved remarkable progress, evolving from simple instruction-driven editing toward more advanced capabilities involving multimodal understanding, complex reasoning, and multi-image editing. 
As frontier models continue to improve, accurately evaluating their editing quality becomes increasingly challenging.
However, existing benchmarks~\cite{ye2025imgedit,liu2025step1x,pan2025ice} often exhibit a noticeable discrepancy between benchmark scores and human judgment, particularly for strong frontier models.
This limitation mainly stems from insufficient task difficulty and coarse-grained evaluation protocols, making it difficult to reliably distinguish subtle capability differences among advanced models.
Accurate evaluation is also crucial for reinforcement learning (RL) based image editing optimization.
Recent works such as EditScore~\cite{luo2025editscore} and EditReward~\cite{wu2025editreward} train reward models to support FlowGRPO-based~\cite{liu2025flow} image editing optimization.
However, existing reward model benchmarks often suffer from a distribution mismatch between evaluation samples and the edited images encountered during RL training, limiting their ability to faithfully assess reward model quality in realistic optimization settings.
Together, these limitations hinder a deeper understanding of frontier image editing models and their corresponding reward models, highlighting the need for a more comprehensive benchmark for faithful image editing evaluation.

To address these challenges, we introduce \textbf{{\bench}} and \textbf{{\rmbench}}, a unified evaluation suite for image editing models and reward models.
As illustrated in Figure~\ref{fig:gallery}, {\bench} contains $2,388$ carefully annotated instances spanning six progressively challenging task categories.
These tasks cover a diverse range of capabilities, including general editing, world perception, dynamic manipulation, visual reasoning, and multi-image understanding.
Beyond broad task coverage, {\bench} further adopts a fine-grained and multi-dimensional evaluation framework.
Each editing result is evaluated through chain-of-thought reasoning guided by well-defined scoring rubrics, enabling more reliable and interpretable assessment in complex editing scenarios.
This design better aligns benchmark evaluation with human judgment while improving evaluation consistency and sensitivity for frontier models.
In parallel, {\rmbench} contains $2,251$ preference pairs that simulate realistic decision-making scenarios encountered by reward models during RL optimization.
Together, this unified benchmark suite enables systematic evaluation of both frontier image editing models and reward models.
It further provides a realistic testbed for analyzing the effectiveness of reward-guided optimization in RL-based image editing.

To validate the effectiveness and difficulty of {\bench} and {\rmbench}, we conduct extensive evaluations on a broad range of frontier models, including $29$ image editing models and $21$ reward models.
For image editing, our evaluation covers state-of-the-art proprietary models, such as Nano-Banana Pro~\cite{nanobananapro}, Wan2.7-Image~\cite{wan}, and Seedream 4.5~\cite{seedream2025seedream}, as well as leading open-source models including Qwen-Image-Edit~\cite{wu2025qwen} and Joy-Image-Edit~\cite{joyaiimage2026}.
The results reveal a substantial performance gap between closed-source and open-source models.
The best proprietary model achieves an overall score of $3.99$, while the strongest open-source model, Qwen-Image-Edit~\cite{wu2025qwen}, reaches only $2.69$.
Beyond overall performance, fine-grained analysis further reveals clear weaknesses in multi-image understanding, world knowledge awareness, and visual reasoning, even for frontier models.
On the reward modeling side, native multimodal large language models~\cite{qwen3.5,su2026omnipack,qwen36_35b_a3b,wang2026beacon,zhang2025debiasing,wang2025monet,qwen3.6-27b,shi2025mavors} achieve stronger overall performance than existing open-source reward models, including models explicitly trained on preference data.
This finding suggests that current reward models remain limited in evaluating visual consistency and perceptual quality under complex editing scenarios.
Overall, our results reveal a fundamental limitation of current image editing systems: while existing models perform reasonably well on shallow perception-level editing tasks, they still struggle with deeper reasoning, world knowledge understanding, and complex multi-image editing.

\section{Related Work}

\subsection{Benchmarks for Image Editing}

\begin{table}[t]
 
\caption{\textbf{Comparison between \rmbench\ and existing image editing reward benchmarks.}
\texttt{AVR} denotes Algorithm Visual Reasoning, \texttt{DM} denotes Dynamic Manipulation, \texttt{WKR} denotes World Knowledge Reasoning, and \texttt{CP} denotes Complex Paint.}
\vspace{0.5em}
\label{tab:compare_image_editing_reward_bench}
\centering
\resizebox{0.95\linewidth}{!}{%
\begin{tabular}{lccccccccc}
\toprule
\textbf{Benchmark} & \textbf{Size} & \textbf{Tasks} & \textbf{Sampling Strategy}  & \textbf{AVR} & \textbf{DM} & \textbf{WK} & \textbf{CP} \\
\midrule
GenAI-Bench~\cite{li2024genai} & 919 & 7 & Cross-model  & \icono & \icono & \icono & \icono  \\
EditReward-Bench~\cite{luo2025editscore} & 3,072 & 13 & Cross-model & \icono & \icono & \icono & \icono  \\
EditReward-Bench~\cite{wu2025editreward} & 1,500 & 8 & Cross-model  & \icono & \icono & \icono & \icono  \\
MMBench2~\cite{hu2025multimodal} & 1,000 & - & Cross-model   & \icono & \icono & \icohalf & \icono  \\
\midrule
\rmbench~(Ours) & 2,251 & 36 & Cross-/Intra-model   & \icoyes & \icoyes & \icoyes & \icoyes  \\

\bottomrule
\end{tabular}%
}

\end{table}

Existing image editing benchmarks face two major limitations: limited task coverage and insufficient evaluation reliability. As shown in Table~\ref{tab:compare_image_editing_bench}, early benchmarks~\cite{zhang2023magicbrush,sheynin2024emu,yu2025anyedit,pan2025ice} mainly focus on narrow editing tasks and rely on automated metrics such as CLIP-I and DINO-I. However, these metrics often fail to capture fine-grained editing quality, especially for tasks involving world knowledge, visual consistency, and complex instruction following.
Recent benchmarks~\cite{ye2025imgedit,liu2025step1x,zhao2025envisioning,zhang2026well} adopt powerful MLLMs as judges for more flexible evaluation. Nevertheless, their reliance on simple judging prompts can lead to unstable assessments and misalignment with human judgment in complex scenarios.
To address these limitations, we propose \textbf{{\bench}}, a comprehensive benchmark covering $36$ fine-grained tasks across six categories. Beyond broad task coverage, {\bench} introduces human-aligned evaluation prompts with structured reasoning and carefully designed scoring rubrics, enabling more accurate, reliable, and interpretable assessment of image editing models.

\subsection{Benchmarks for Image Editing Reward Model}

With the rapid progress of image generation and editing, reward models have become increasingly important for improving instruction following and visual consistency through reinforcement learning (RL). Accordingly, reliable evaluation of image editing reward models has attracted growing attention.
As shown in Table~\ref{tab:compare_image_editing_reward_bench}, existing benchmarks~\cite{luo2025editscore,wu2025editreward,hu2025multimodal} typically construct preference pairs from limited editing tasks or from outputs generated by different models. However, such settings often deviate from practical RL scenarios, where reward models are required to compare candidate outputs produced by the same editing model under the same instruction. This mismatch limits faithful assessment of reward model quality and training effectiveness.
Recent efforts~\cite{zhao2025envisioning,deng2025emerging} have expanded evaluation coverage to more diverse tasks, including world knowledge and visual reasoning. Nevertheless, existing benchmarks still lack realistic and controlled preference construction, particularly in balancing task diversity and comparison consistency.
To bridge this gap, we propose \textbf{{\rmbench}}, a comprehensive benchmark for evaluating image editing reward models. {\rmbench} constructs preference pairs under more realistic and controlled settings, enabling multidimensional analysis of reward models in terms of instruction following, visual consistency, perceptual quality, and reasoning-aware editing preference.

% \begin{table}[t]
 
% \caption{\textbf{Comparison between \rmbench\ and existing image editing reward benchmarks.}
% \texttt{AVR} denotes Algorithm Visual Reasoning, \texttt{DM} denotes Dynamic Manipulation, \texttt{WKR} denotes World Knowledge Reasoning, and \texttt{CP} denotes Complex Paint.}
% \vspace{0.5em}
% \label{tab:compare_image_editing_reward_bench}
% \centering
% \resizebox{0.65\linewidth}{!}{%
% \begin{tabular}{lccccccccc}
% \toprule
% \textbf{Benchmark} & \textbf{Size} & \textbf{Tasks} & \textbf{Sampling Strategy}  & \textbf{AVR} & \textbf{DM} & \textbf{WK} & \textbf{CP} \\
% \midrule
% GenAI-Bench~\cite{li2024genai} & 919 & 7 & Cross-model  & \icono & \icono & \icono & \icono  \\
% EditReward-Bench~\cite{luo2025editscore} & 3,072 & 13 & Cross-model & \icono & \icono & \icono & \icono  \\
% EditReward-Bench~\cite{wu2025editreward} & 1,500 & 8 & Cross-model  & \icono & \icono & \icono & \icono  \\
% MMBench2~\cite{hu2025multimodal} & 1,000 & - & Cross-model   & \icono & \icono & \icohalf & \icono  \\
% \midrule
% \rmbench~(Ours) & 2,251 & 36 & Cross-/Intra-model   & \icoyes & \icoyes & \icoyes & \icoyes  \\

% \bottomrule
% \end{tabular}%
% }
% \vspace{-1.5em}
% \end{table}

\section{\bench}

\begin{figure}[t]
  \centering
  \includegraphics[width=0.8\linewidth]{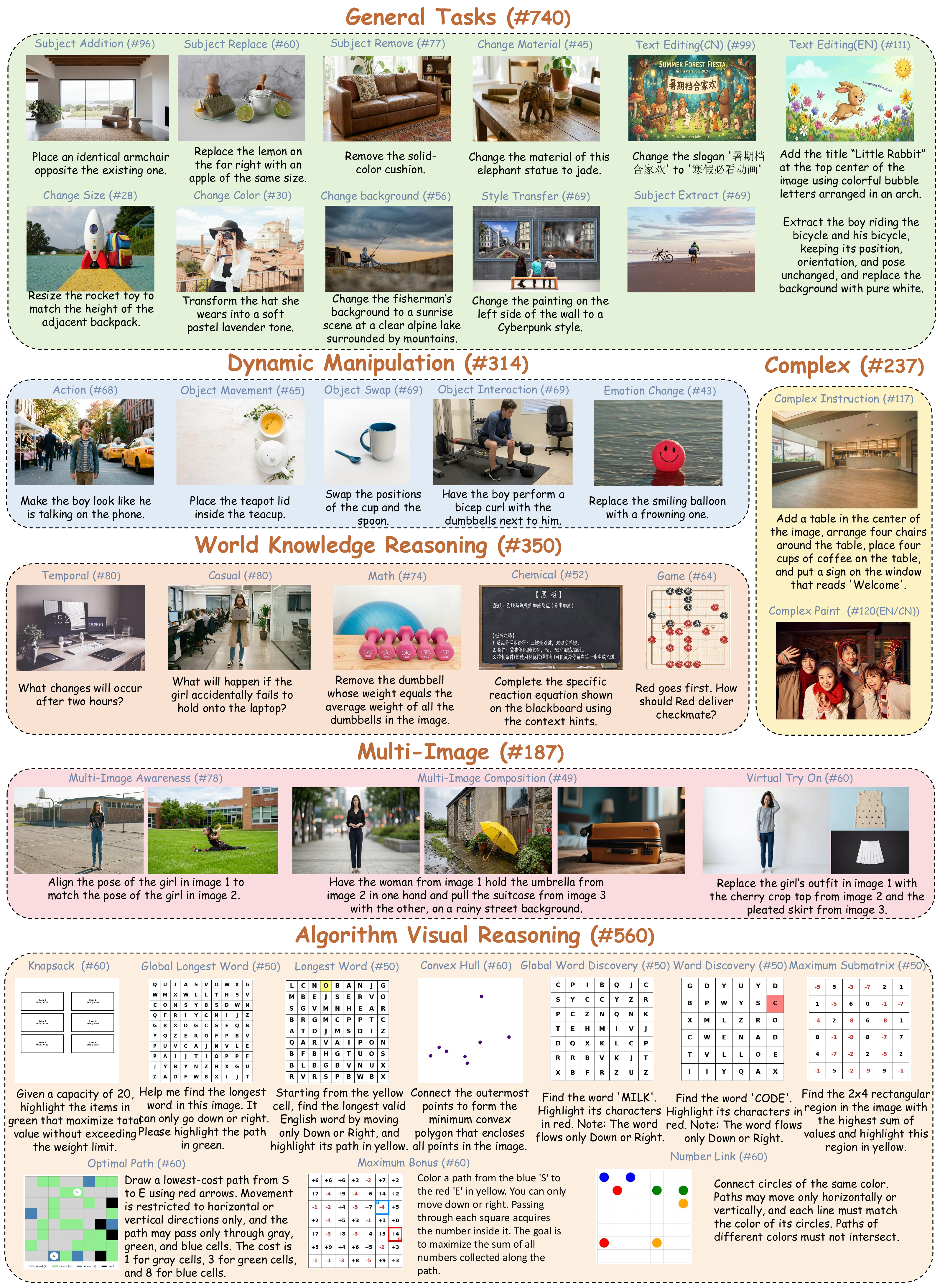}
   
  \caption{\textbf{{\bench}} covers 36 diverse image editing tasks, spanning single-image and multi-image settings as well as general editing and algorithmic visual reasoning. Each panel shows a representative example for a task type, with the number of examples (\#) indicated.}
  \label{fig:gallery}
  \vspace{-1.0em}
\end{figure}

\subsection{Task Taxonomy}

\textbf{General Tasks.} 
General tasks evaluate the fundamental image editing capabilities of models, focusing on instruction understanding and accurate execution across both global and local editing scenarios. Global editing includes tasks such as \textit{style transfer} and \textit{background transformation}, while local editing extends beyond conventional operations like addition, removal, and replacement to more fine-grained edits.
As illustrated in Figure~\ref{fig:gallery}, we introduce a novel \textit{Copy} task, which requires models to duplicate an existing object within the input image while preserving its visual attributes and maintaining spatial coherence. We further include challenging tasks such as \textit{change size}, which evaluate the ability to manipulate object scale and spatial relationships. Together, these tasks provide a comprehensive evaluation of general image editing capabilities at both global and object levels.

\textbf{Dynamic Manipulation Tasks.}
Dynamic Manipulation tasks evaluate a model's ability to perform object-level dynamic edits in real-world scenes, focusing on actions, movements, emotional changes, and inter-object interactions. Unlike general editing tasks, this category emphasizes dynamic scene understanding and interaction modeling.
Specifically, this category includes five subtasks: (1) \textit{Action}, which modifies object motion; (2) \textit{Emotion Change}, which alters object expressions or affective states; (3) \textit{Object Movement}, which repositions objects within the scene; (4) \textit{Object Swap}, which exchanges attributes such as appearance, color, or state between objects; and (5) \textit{Object Interaction}, which evaluates the modeling of interactions among multiple objects.

\textbf{World Knowledge Reasoning Tasks.}
These tasks evaluate a model's ability to leverage real-world knowledge to infer and execute intended edits. We define five representative subtasks: (1) \textit{Temporal Reasoning}, which involves reasoning about past and future changes over time; (2) \textit{Causal Reasoning}, which evaluates understanding of object changes under external conditions; (3) \textit{Game Reasoning}, which requires reasoning about game rules and states; (4) \textit{Math Reasoning}, which tests mathematical reasoning ability; and (5) \textit{Chemical Reasoning}, which involves understanding chemical phenomena and reactions.
These tasks evaluate models' abilities in temporal, causal, and domain-specific reasoning for complex image editing scenarios.

\textbf{Algorithmic Visual Reasoning Tasks.}  
Algorithmic visual reasoning tasks evaluate whether models can interpret visual inputs and perform multi-step reasoning to execute corresponding edits. This category includes ten task types, such as \textit{Optimal Path Identification}, \textit{Convex Hull Identification}, \textit{Maximum Submatrix Sum Identification}, and \textit{Knapsack Selection}. These tasks require models to understand visual structures, reason over them, and faithfully render the results through image editing, providing a challenging benchmark for deep visual reasoning in image editing.

\textbf{Multi-Image Tasks.} 
Multi-image tasks evaluate a model's ability to understand and integrate multiple input images for image editing. Beyond \textit{Multi-Image Composition} and \textit{Virtual Try-On}, we introduce a novel task termed \textit{Multi-Image-Aware Editing}, where models edit a target image based on fine-grained attributes extracted from reference images, such as object properties, actions, orientations, and colors. These tasks comprehensively evaluate models' abilities to understand, transfer, and manipulate visual information across multiple images.

\textbf{Complex Tasks.} 
Complex tasks evaluate a model's ability to handle compound instructions involving multiple editing intents. Unlike single-step editing tasks, these tasks require coherent execution of multiple edits within the source image.
We further introduce \textit{Complex Paint}, a multimodal editing task that incorporates visual guidance directly into the source image through cues such as arrows, circles, and cross marks. This setting better reflects real-world interactive editing scenarios, where users combine textual instructions with visual indications to specify complex edits.
These tasks provide a more rigorous evaluation of compositional and multimodal image editing capabilities.

\begin{figure}[t]
  \centering
  \includegraphics[width=0.95\textwidth]{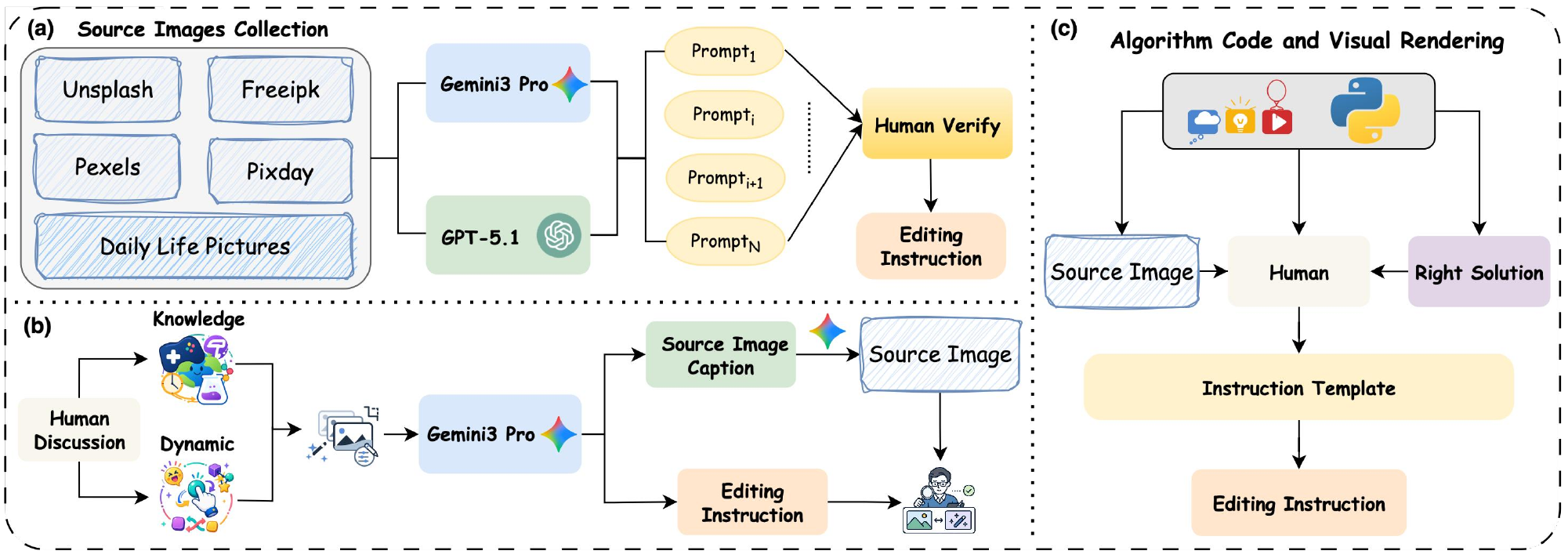}
   
  \caption{\textbf{Overview of the source data construction pipelines in \bench.}
(a) General and Complex tasks.
(b) Dynamic Manipulation, World Knowledge Reasoning, and Multi-Image tasks.
(c) Algorithmic Visual Reasoning tasks.}
  \label{Fig: Data Construction}
  \vspace{-1em}
\end{figure}

% Summary
\subsection{Benchmark Construction}
The source data in {\bench} consists of original images and executable editing instructions. As illustrated in Figure~\ref{Fig: Data Construction}, we adopt three data construction strategies tailored to different task categories.
For General and Complex tasks, original images are collected from online resources and real-world photographs, while editing instructions are generated with Gemini 3 Pro~\cite{gemini3pro} and GPT-5.1~\cite{gpt4o20250325}, followed by human verification. For Dynamic Manipulation, World Knowledge Reasoning, and Multi-Image tasks, image-editing experts design challenging yet realistic scenarios, describe the desired source images, and construct bilingual editing instructions in Chinese and English. The source images are then generated from enhanced prompts refined by Gemini 3 Pro~\cite{gemini3pro}. For Algorithmic Visual Reasoning tasks, we programmatically generate source images using Python and derive ground-truth annotations from algorithmic solutions.
To ensure consistency and clarity, we design unified instruction templates for each task category that specify task requirements and intended outcomes. All samples are further reviewed by multiple human experts to ensure data quality. More details are provided in Appendix~\ref{supp: Edit-Compass Data Construction}.

\subsection{Evaluation Pipeline}
Accurately evaluating diverse image editing tasks in a human-aligned manner remains challenging, especially for instruction adherence and visual consistency. To address this, we structure the evaluation around three core dimensions: \textbf{Instruction Awareness}, \textbf{Visual Consistency}, and \textbf{Visual Quality}. Based on these dimensions, we design an MLLM-as-judge evaluation pipeline that produces both scalar scores and fine-grained rationales. For reasoning-intensive tasks, the rationale further includes the expected ground-truth outcome, improving evaluation accuracy and interpretability. More details are provided in Appendix~\ref{supp: Evaluation pipeline}.

\textbf{Dimension 1: Instruction Awareness.}
This dimension evaluates whether the edited image correctly follows the instruction and reflects the intended change. It consists of two dynamic subcomponents: \textit{Instruction Following} and \textit{World Knowledge Awareness}. Instruction following assesses whether the model correctly identifies the target object, applies the required attribute or spatial modification, and satisfies explicit constraints. World knowledge awareness evaluates whether the model incorporates relevant real-world knowledge and visual cues to infer implicit editing intent.

\textbf{Dimension 2: Visual Consistency.}
This dimension measures whether visual content unrelated to the requested edit is preserved. It includes \textit{Unedited Region Consistency (URC)} and \textit{Identity Consistency}. URC evaluates whether non-edited regions remain unchanged at both local and global levels. Identity consistency assesses whether the edited object preserves attributes irrelevant to the requested modification, avoiding unintended changes in appearance, structure, or identity.

\textbf{Dimension 3: Visual Quality.}
This dimension evaluates whether the edited image is visually plausible, coherent, and artifact-free. It considers naturalness, structural fidelity, artifact severity, distortion, and text legibility when applicable.

\section{\rmbench}
{\rmbench} is designed to systematically evaluate reward models for image editing. It contains $2,251$ preference pairs, each consisting of an editing instruction and two candidate edited images. We evaluate reward models using the same rubric-based judging framework as {\bench}, enabling consistent assessment across image editing models and reward models. This also allows us to examine the robustness and generality of our evaluation prompts. The construction of {\rmbench} follows two stages: sampling (Section~\ref{Sampling Stage}) and human annotation (Section~\ref{Reward Model Benchmark Human Annotation}).

\subsection{Sampling Stage}
\label{Sampling Stage}
We use {\bench} as the source data for constructing {\rmbench}, as its diverse and executable editing instructions provide broad coverage of realistic editing scenarios. To better reflect reward modeling during RL optimization, we simulate the sampling process with a FlowGRPO-inspired strategy~\cite{liu2025flow} and introduce stochasticity through stochastic differential equations~\cite{song2020score}.
Specifically, we sample candidate outputs from six image editing models and control the denoising steps to ensure visually clear and valid results. For tasks involving world knowledge or complex reasoning, where open-source models often show limited capability, we further expand the sampling pool to ten diverse open-source and proprietary models to improve task diversity and coverage. Additional implementation details are provided in Appendix~\ref{supp: EditReward-Compass Data Construction}.

\subsection{Human Annotation Stage}
\label{Reward Model Benchmark Human Annotation}
To ensure the quality of {\rmbench}, we employ a two-stage human annotation pipeline to select preference pairs along multiple dimensions, including instruction adherence, visual consistency, and visual quality. Given the complexity of image editing evaluation, {\rmbench} places particular emphasis on instruction adherence and visual consistency. The annotation process involves eight human experts in image editing.
In the first stage, three annotators independently review sampled outputs to construct candidate preference pairs. Ambiguous cases are flagged and resolved through discussion, leading to either consensus decisions or sample removal. In the second stage, five annotators conduct fine-grained verification of the selected pairs, checking both task validity and preference correctness. A pair is retained only when all five annotators reach unanimous agreement, ensuring high annotation consistency.

\begin{table*}[t]
    \centering
    
    \caption{
    \textbf{Main results on {\bench} under English instructions.} The best results are marked in \textbf{bold} for
    {\color[HTML]{F88825} open-} and {\color[HTML]{319B62} closed-} models, respectively.}
    \label{tab:Image Editing Bench Main Results_EN}
    \resizebox{\linewidth}{!}{
        \begin{tabular}{lc|ccc|ccc|ccc|ccc|ccc|ccc|c}
            \toprule
            \multirow{2}{*}{Model} & \multirow{2}{*}{Multi-Img}
            & \multicolumn{3}{c}{\textbf{General}}
            & \multicolumn{3}{c}{\textbf{Dynamic Manipulation}}
            & \multicolumn{3}{c}{\textbf{World Knowledge}}
            & \multicolumn{3}{c}{\textbf{Visual Reasoning}}
            & \multicolumn{3}{c}{\textbf{Multi Image }}
            & \multicolumn{3}{c}{\textbf{Complex}}
            & \textbf{Overall} \\
            &
            & $\mathcal{IA}\uparrow$ & $\mathcal{VC}\uparrow$ & $\mathcal{VQ}\uparrow$
            & $\mathcal{IA}\uparrow$ & $\mathcal{VC}\uparrow$ & $\mathcal{VQ}\uparrow$
            & $\mathcal{IA}\uparrow$ & $\mathcal{VC}\uparrow$ & $\mathcal{VQ}\uparrow$
            & $\mathcal{IA}\uparrow$ & $\mathcal{VC}\uparrow$ & $\mathcal{VQ}\uparrow$
            & $\mathcal{IA}\uparrow$ & $\mathcal{VC}\uparrow$ & $\mathcal{VQ}\uparrow$
            & $\mathcal{IA}\uparrow$ & $\mathcal{VC}\uparrow$ & $\mathcal{VQ}\uparrow$
            & \cellcolor[HTML]{C8C8C8}\textbf{AVG} \\ \hline

InstructPix2Pix~\cite{brooks2023instructpix2pix} & \ding{55}
& 1.98 & 1.34 & 2.33 & 1.72 & 1.61 & 2.23 & 1.51 & 1.56
& 2.57 & 1.09 & 1.66 & 3.26 & - & - & - & 1.91 & 1.15 &  2.07
& \cellcolor[HTML]{C8C8C8}1.19 \\

UltraEdit~\cite{zhao2024ultraedit} & \ding{55}
& 2.23 & 1.80 & 2.44 & 1.82 & 2.37 & 2.44 & 1.56 & 2.32 & 2.89
& 1.04 & 3.34 & 3.71 & - & - & - & 1.82  & 1.31 & 2.11
& \cellcolor[HTML]{C8C8C8}1.28 \\

AnyEdit~\cite{yu2025anyedit} & \icohalfno 
& 2.05 & 2.36 & 2.47  & 1.61 & 2.73 & 2.48 & 1.45 & 2.81
& 2.95 & 1.03 & 3.21 & 3.44 & - & - & - & 1.34 & 1.78 & 2.07
& \cellcolor[HTML]{C8C8C8}1.31 \\

MagicBrush~\cite{zhang2023magicbrush} & \ding{55}
& 2.21  & 2.11 & 2.28 & 1.75 & 2.42 & 2.29 & 1.51 & 2.58 & 2.49  & 1.05
& 1.69 & 2.37 & - & - & - & 1.53 & 1.66 & 2.10
& \cellcolor[HTML]{C8C8C8}1.33 \\

FLUX.1 Kontext Dev~\cite{labs2025flux} & \ding{55}
& 3.53 & 2.98 & 3.09 & 2.41 & 2.93 & 2.81 & 1.65 & 3.22 & 3.32 & 1.27
& 3.98 & \color[HTML]{F88825}\textbf{4.78} & - & - & - & 2.41 & 2.32 & 2.63
& \cellcolor[HTML]{C8C8C8}1.93 \\

FLUX.2 Dev~\cite{flux-2-2025} & $\checkmark$
& 4.24 & 4.16 & \color[HTML]{F88825}\textbf{3.96} & 3.23 & \color[HTML]{F88825}\textbf{4.36} & \color[HTML]{F88825}\textbf{3.36} &  2.05 & 3.85 & \color[HTML]{F88825}\textbf{3.71}
& 1.42 & 2.85 & 4.56 & 3.16 & 4.13 & \color[HTML]{F88825}\textbf{3.14} & 2.62 & \color[HTML]{F88825}\textbf{4.32} & \color[HTML]{F88825}\textbf{3.54}
& \cellcolor[HTML]{C8C8C8} 2.61 \\
\hline

OneCAT~\cite{li2025onecat} & \ding{55}
& 2.23 & 1.12 & 2.12
& 1.67 & 1.26 & 2.15
& 1.41 & 1.29 & 2.09
& 1.02 & 1.05 & 2.06
& - & - & -
& 1.79 & 1.03 & 1.99
& \cellcolor[HTML]{C8C8C8}1.21 \\

Lumina-DiMOO~\cite{xin2025lumina} & \ding{55}
& 2.37 & 2.00 & 2.33
& 1.73 & 2.31 & 2.40
& 1.50 & 2.89 & 2.93
& 1.09 & 2.08 & 2.83
& - & - & -
& 1.70 & 1.54 & 2.08
& \cellcolor[HTML]{C8C8C8}1.33 \\

Nextstep-V1~\cite{han2025nextstep} & \ding{55}
& 3.31 & 1.73 & 2.29
& 2.35 & 1.81 & 2.23
& 1.74 & 1.81 & 2.28
& 1.11 & 1.02 & 2.08
& - & - & -
& 2.14 & 1.22 & 2.09
& \cellcolor[HTML]{C8C8C8}1.45 \\

InternVL-U~\cite{tian2026internvl} & \ding{55}
& 3.84 & 2.13 & 2.69
& 2.54 & 1.91 & 2.40
& 1.82 & 2.35 & 2.69
& 1.18 & 1.91 & 2.66
& - & - & -
& 2.42 & 1.38 & 2.22
& \cellcolor[HTML]{C8C8C8}1.59 \\

UniWorld-V1~\cite{lin2025uniworld} & $\checkmark$
& 2.94 & 3.19 & 3.09 & 1.85 & 3.34 & 2.93 & 1.47 & 3.17 & 3.27
& 1.06 & 1.59 & 3.59 & 1.74 & 2.02 & 2.81 & 1.74 & 2.65 & 2.41
& \cellcolor[HTML]{C8C8C8}1.71 \\

HiDream-E1~\cite{cai2025hidream} & \ding{55}
& 3.29 & 2.42 & 2.75
& 2.66 & 2.65 & 2.68
& 1.77 & 2.54 & 2.80
& 1.22 & 2.55 & 3.75
& - & - & -
& 2.26 & 1.59 & 2.14
& \cellcolor[HTML]{C8C8C8}1.76 \\

ChronoEdit~\cite{wu2025chronoedit} & \ding{55}
& 2.63 & 3.63 & 3.49
& 2.85 & 3.84 & 3.22
& 1.61 & 3.26 & 3.35
& 1.15 & 1.87 & 3.45
& - & - & -
& 1.97 & 3.00 & 2.83
& \cellcolor[HTML]{C8C8C8}1.85 \\

OmniGen2~\cite{wu2025omnigen2} & $\checkmark$
& 3.33 & 3.34 & 3.12
& 2.47 & 3.70 & 3.19
& 1.53 & 3.33 & 3.55
& 1.11 & 2.51 & 3.69
& 2.55 & 1.77 & 2.65
& 2.27 & 2.73 & 2.73
& \cellcolor[HTML]{C8C8C8}1.88 \\

DeepGen 1.0~\cite{wang2026deepgen} & \ding{55}
& 3.85 & 2.45 & 2.94
& 2.88 & 2.50 & 2.62
& 2.30 & 2.68 & 3.13
& \color[HTML]{F88825}\textbf{1.57} & \color[HTML]{F88825}\textbf{4.33} & 4.44
& - & - & -
& 2.66 & 1.68 & 2.28
& \cellcolor[HTML]{C8C8C8}1.91 \\

UniReason1.0~\cite{wang2026unireason} & $\checkmark$
& 3.58 & 2.82 & 2.81
& 3.31 & 3.09 & 2.71
& 2.32 & 3.05 & 2.90
& 1.42 & 4.08 & 3.20
& 1.90 & 1.59 & 2.77
& 2.55 & 1.52 & 2.30
& \cellcolor[HTML]{C8C8C8}1.92 \\

Bagel~\cite{deng2025emerging} & $\checkmark$
& 3.60 & 3.79 & 3.03
& 2.46 & 3.65 & 2.82
& 1.78 & 3.59 & 3.38
& 1.21 & 3.28 & 3.83
& 1.86 & 2.03 & 3.06
& 2.30 & 2.99 & 2.65
& \cellcolor[HTML]{C8C8C8}2.08 \\

Bagel-Think~\cite{deng2025emerging} & $\checkmark$
& 3.52 & 3.92 & 3.14
& 2.49 & 3.73 & 2.97
& 2.22 & 3.09 & 3.27
& 1.21 & 3.25 & 3.48
& 1.85 & 2.01 & 3.03
& 2.19 & 3.35 & 2.89
& \cellcolor[HTML]{C8C8C8}2.09 \\

Unipic3~\cite{wei2026skywork} & $\checkmark$
& 4.22 & 3.85 & 3.52
& 3.25 & 3.58 & 3.03
& 2.02 & 3.37 & 3.07
& 1.20 & 2.47 & 3.73
& 3.00 & 3.04 & 2.72
& 2.76 & 2.63 & 2.74
& \cellcolor[HTML]{C8C8C8}2.35 \\

UniWorld-V2~\cite{li2025uniworld} & $\checkmark$
& 4.39 & 4.04 & 3.74
& 3.64 & 3.64 & 3.03
& 2.22 & 3.27 & 3.25
& 1.21 & 2.27 & 3.64
& 3.05 & 3.13 & 2.82
& 2.91 & 2.73 & 2.82
& \cellcolor[HTML]{C8C8C8}2.53 \\

Step1X-Edit-v1p2~\cite{liu2025step1x} & \ding{55}
& 4.26 & 4.31 & 3.64
& 3.16 & 4.15 & 3.12
& 2.30 & \color[HTML]{F88825}\textbf{4.09} & 3.49
& 1.26 & 2.72 & 4.41
& - & - & -
& 2.94 & 3.33 & 2.81
& \cellcolor[HTML]{C8C8C8}2.58 \\

Longcat-Image-Edit~\cite{team2025longcat} & \ding{55}
& 4.51 & \color[HTML]{F88825}\textbf{4.48} & 3.90
& 3.54 & 4.14 & 3.38
& 1.98 & 3.94 & 3.49
& 1.31 & 3.26 & 4.21
& - & - & -
& 2.72 & 3.31 & 3.19
& \cellcolor[HTML]{C8C8C8}2.65 \\

EMU3.5~\cite{cui2025emu3} & $\checkmark$
& 4.45 & 4.00 & 3.68 & 3.78 & 3.79 & 3.23 & \color[HTML]{F88825}\textbf{2.60} & 3.55 & 3.33
& 1.43 & 3.54 & 3.83 & \color[HTML]{F88825}\textbf{3.46} & 3.43 & 2.89 & 2.71 & 3.28 & 2.93
& \cellcolor[HTML]{C8C8C8}2.66 \\

JoyAI-Image-Edit~\cite{joyaiimage2026} & \ding{55}
& 4.56 & 4.35 & 3.61
& 3.65 & 4.14 & 3.17
& 2.35 & 3.87 & 3.46
& 1.41 & 2.56 & 4.38
& - & - & -
& \color[HTML]{F88825}\textbf{3.00} & 3.41 & 2.87
& \cellcolor[HTML]{C8C8C8}2.68 \\

Qwen-Image-Edit-2511~\cite{wu2025qwen} & $\checkmark$
& \color[HTML]{F88825}\textbf{4.61} & 4.39 & 3.75
& \color[HTML]{F88825}\textbf{3.81} & 3.93 & 3.16
& 2.33 & 3.56 & 3.25
& 1.26 & 2.53 & 4.09
& 3.27 & \color[HTML]{F88825}\textbf{3.55} & 2.92
& 2.80 & 3.81 & 3.02
& \cellcolor[HTML]{C8C8C8}\color[HTML]{F88825}\textbf{2.69} \\
\hline
% \noalign{\vskip 3pt}

Seedream 4.5~\cite{seedream2025seedream} & $\checkmark$
& 4.66 & 4.36 & 4.13
& 4.36 & 4.15 &  3.89
& 3.58 & 4.07 & 4.04
& 1.60 & 2.90 & 4.50
& 4.34 & 4.13 & 3.42
& 4.04 & 4.11 & 3.44
& \cellcolor[HTML]{C8C8C8}3.22 \\

Wan2.7-image~\cite{wan} & $\checkmark$
& 4.60 & 4.36 & 4.07
& 4.40 & 4.11 & \color[HTML]{319B62}\textbf{3.92}
& 3.65 & 4.16 & 4.01
& 1.61 & 2.81 & 4.49
& 4.25 & 4.23 & 3.41
& 3.97 & 4.04 & 3.41
& \cellcolor[HTML]{C8C8C8} 3.23  \\

Nano Banana 2~\cite{google2026gemini31flashimagepreview}  & $\checkmark$
& \color[HTML]{319B62}\textbf{4.79} & 4.54 &  \color[HTML]{319B62}\textbf{4.14}
& 4.50 & 4.33 & 3.71
& 4.20 & 4.35 & 4.19
& 2.77 & 4.03 & 4.63
&  \color[HTML]{319B62}\textbf{4.50} &  \color[HTML]{319B62}\textbf{4.37} & 3.41
& \color[HTML]{319B62}\textbf{4.37}  &  \color[HTML]{319B62}\textbf{4.46} & 3.50
& \cellcolor[HTML]{C8C8C8} 3.74 \\

Nano Banana Pro~\cite{gemini3pro} & $\checkmark$
& 4.76 & \color[HTML]{319B62}\textbf{4.70} & 4.11
&  \color[HTML]{319B62}\textbf{4.54} &  \color[HTML]{319B62}\textbf{4.58} & 3.79
&  \color[HTML]{319B62}\textbf{4.33} &  \color[HTML]{319B62}\textbf{4.49} &  \color[HTML]{319B62}\textbf{4.28}
&  \color[HTML]{319B62}\textbf{3.61} &  \color[HTML]{319B62}\textbf{4.25} &  \color[HTML]{319B62}\textbf{4.73}
& 4.43 & 4.29 &  \color[HTML]{319B62}\textbf{3.44}
&  4.28& 4.40 &  \color[HTML]{319B62}\textbf{3.53}
& \cellcolor[HTML]{C8C8C8} \color[HTML]{319B62}\textbf{3.99} \\
 
            \bottomrule
        \end{tabular}}
\end{table*}

\section{Experiments}
\label{sec:experiments}
\subsection{Experimental Setup}
For the image editing evaluation, we benchmark a total of $29$ models, comprising $25$ open-source models and $4$ proprietary models, thereby covering a broad range of recent image editing paradigms.
The open-source models span diverse architectural families. Diffusion-based methods include InstructPix2Pix~\cite{brooks2023instructpix2pix}, MagicBrush~\cite{zhang2023magicbrush}, AnyEdit~\cite{yu2025anyedit}, UltraEdit~\cite{zhao2024ultraedit}, and Flux-Kontext~\cite{labs2025flux}. 
Unified multimodal models include  EMU3.5~\cite{cui2025emu3}, OneCAT~\cite{li2025onecat}, NextStep-V1~\cite{han2025nextstep}, BAGEL~\cite{deng2025emerging}, Qwen-Image-Edit~\cite{wu2025qwen}, Step1X-Edit-v1.2~\cite{liu2025step1x}, UniWorld-V1~\cite{lin2025uniworld},
UniWorld-V2~\cite{li2025uniworld},
DeepGen1.0~\cite{wang2026deepgen}, UniPic3~\cite{wei2026skywork}, UniReason~\cite{wang2026unireason}, and OmniGen2~\cite{wu2025omnigen2}.
The proprietary models include Nano Banana Pro\cite{gemini3pro}, Nano Banana 2~\cite{google2026gemini31flashimagepreview}, Wan2.7-Image\cite{wan}, and Seedream4.5\cite{seedream2025seedream}, which are incorporated to provide a more comprehensive evaluation of state-of-the-art systems. 
In addition, we evaluate three categories of reward models for image editing, covering open-source general-purpose multimodal models, image-editing-specific reward models trained on human preference data, and proprietary models. The open-source general-purpose multimodal models include Qwen2.5-VL~\cite{wang2024qwen2}, Qwen3-VL~\cite{bai2025qwen3}, native multimodal models such as Qwen3.5~\cite{qwen3.5} and Qwen3.6~\cite{qwen3.6-27b,qwen36_35b_a3b}, as well as Gemma3~\cite{Gemma3} and Gemma4~\cite{google2026gemma4}. 
The image-editing-specific reward models include EditScore~\cite{luo2025editscore} and EditReward~\cite{wu2025editreward}. 
For comparison, we also include proprietary models from the GPT-4.1~\cite{gpt4o20250325}, Gemini 3 Flash~\cite{google2026gemini31flashimagepreview}, and Gemini 3.1 Pro~\cite{google2026gemini31propreview}.

\subsection{Main Results}
\textbf{Image Editing Model Results.}
Tables~\ref{tab:Image Editing Bench Main Results_EN} and~\ref{tab:Image Editing Bench Main Results_CN} report the results on {\bench} under English and Chinese instructions, respectively. Overall, the benchmark reveals clear performance differences across models, task categories, and evaluation dimensions.
Among open-source models, Qwen-Image-Edit~\cite{wu2025qwen} achieves the best overall performance under both English and Chinese instructions, likely benefiting from its integration of a 20B diffusion transformer with a 7B Qwen-VL model. Among closed-source models, Nano Banana Pro~\cite{nanobananapro} performs best and shows consistent advantages across task categories.
The progressive task design of {\bench} further reveals where open-source models remain competitive and where they still lag behind. On relatively basic categories such as General and Dynamic Manipulation, open-source models including Qwen-Image-Edit~\cite{wu2025qwen} and Longboat-Image-Edit~\cite{team2025longcat} achieve performance comparable to closed-source systems such as Seedream 4.5~\cite{seedream2025seedream} and Wan2.7-Image~\cite{wan}. However, a substantial gap remains on more challenging categories, including World Knowledge Reasoning, Algorithmic Visual Reasoning, Multi-Image, and Complex Tasks. For example, on World Knowledge Reasoning, Nano Banana Pro achieves a score of $3.89$, while Qwen-Image-Edit obtains only $1.74$, indicating that current open-source models still struggle with knowledge-intensive and reasoning-intensive editing.
Additional experimental results are provided in Appendix~\ref{supp:Evaluation Details}.

\textbf{Reward Model Results.}

As shown in Table~\ref{tab:RMBench Main Result}, {\rmbench} compares diverse reward-model candidates across Instruction Awareness, Visual Consistency , and Visual Quality. Among open-source models, we observe a clear scaling trend within the same model families, where larger models generally perform better across dimensions. Visual Consistency and Visual Quality remain more challenging than Instruction Awareness, suggesting that fine-grained visual preservation and perceptual quality assessment are still difficult for current models.
Native multimodal models show strong potential, especially the Qwen3.5~\cite{qwen3.5} and Qwen3.6~\cite{qwen3.6-27b,qwen36_35b_a3b} series. Notably, Qwen3.5-9B outperforms Gemma3-27B~\cite{google2026gemma4} and performs comparably to the much larger Qwen3-VL-32B~\cite{bai2025qwen3}.
Among preference-trained reward models, EditReward outperforms EditScore under the same Qwen2.5-VL-7B backbone on both Instruction Awareness and Visual Quality.

\begin{table*}[t]
    \centering
    
    \caption{
    \textbf{Main results on {\bench} under Chinese instructions.} The best results are marked in \textbf{bold} for
    {\color[HTML]{F88825} open-} and {\color[HTML]{319B62} closed-} models, respectively.}
    \label{tab:Image Editing Bench Main Results_CN}
    \resizebox{\linewidth}{!}{
        \begin{tabular}{lc|ccc|ccc|ccc|ccc|ccc|ccc|c}
            \toprule
            \multirow{2}{*}{Model} & \multirow{2}{*}{Multi-Img}
            & \multicolumn{3}{c}{\textbf{General}}
            & \multicolumn{3}{c}{\textbf{Dynamic Manipulation}}
            & \multicolumn{3}{c}{\textbf{World Knowledge}}
            & \multicolumn{3}{c}{\textbf{Visual Reasoning}}
            & \multicolumn{3}{c}{\textbf{Multi Image }}
            & \multicolumn{3}{c}{\textbf{Complex}}
            & \textbf{Overall} \\
            &
            & $\mathcal{IA}\uparrow$ & $\mathcal{VC}\uparrow$ & $\mathcal{VQ}\uparrow$
            & $\mathcal{IA}\uparrow$ & $\mathcal{VC}\uparrow$ & $\mathcal{VQ}\uparrow$
            & $\mathcal{IA}\uparrow$ & $\mathcal{VC}\uparrow$ & $\mathcal{VQ}\uparrow$
            & $\mathcal{IA}\uparrow$ & $\mathcal{VC}\uparrow$ & $\mathcal{VQ}\uparrow$
            & $\mathcal{IA}\uparrow$ & $\mathcal{VC}\uparrow$ & $\mathcal{VQ}\uparrow$
            & $\mathcal{IA}\uparrow$ & $\mathcal{VC}\uparrow$ & $\mathcal{VQ}\uparrow$
            & \cellcolor[HTML]{C8C8C8}\textbf{AVG} \\ \hline

AnyEdit~\cite{yu2025anyedit} & \icohalfno
& 1.28 & 2.50 & 2.65 & 1.48 & 3.01 & 2.56 & 1.30 & 2.91 & 3.13
& 1.00 & 2.75 & 3.13 & - & - & - & 1.15 & 1.89 & 2.15
& \cellcolor[HTML]{C8C8C8}1.13 \\

MagicBrush~\cite{zhang2023magicbrush} & \ding{55}
& 1.47  & 1.94 & 2.31 & 1.27 & 1.42 & 2.22 & 2.28 & 2.00 & 2.54  & 1.00
& 1.30 & 2.34 & - & - & - & 1.29 & 1.62 & 2.09
& \cellcolor[HTML]{C8C8C8}1.14 \\

UltraEdit~\cite{zhao2024ultraedit} & \ding{55}
& 2.23 & 1.80 & 2.44 & 1.82 & 2.37 & 2.44 & 1.56 & 2.32 & 2.89
& 1.04 & 3.34 & 3.71 & - & - & - & 1.82 & 1.31 & 2.11
& \cellcolor[HTML]{C8C8C8}1.15 \\

InstructPix2Pix~\cite{brooks2023instructpix2pix} & \ding{55}
& 1.48 & 2.40 & 2.68 & 1.44 & 2.8 & 2.69  & 1.37 & 2.85
& 3.22 & 1.00 & 1.65 & 2.64 & - & - & - & 1.19 & 1.98 &  2.14
& \cellcolor[HTML]{C8C8C8}1.17 \\

FLUX.1 Kontext Dev~\cite{labs2025flux} & \ding{55}
& 1.47 & 3.05 & 3.14 & 1.35 & 3.02 & 2.91 & 1.29 & 3.33 & 3.51
& 1.01 & \color[HTML]{F88825}\textbf{4.43} & 4.85 & - & - & - & 1.16 & 2.93 & 2.67
& \cellcolor[HTML]{C8C8C8}1.18 \\

FLUX.2 Dev~\cite{flux-2-2025} & $\checkmark$
& 4.21 & 4.09 & \color[HTML]{F88825}\textbf{3.97} & 3.36 & 4.21 & \color[HTML]{F88825}\textbf{3.44} &  2.17  & 3.85  & \color[HTML]{F88825}\textbf{3.77}
& 1.24 & 2.75 & 4.41 & 3.27 & \color[HTML]{F88825}\textbf{4.17} & \color[HTML]{F88825}\textbf{3.20} &  2.55  & \color[HTML]{F88825}\textbf{4.27} & \color[HTML]{F88825}\textbf{3.44}
&  \cellcolor[HTML]{C8C8C8}2.60  \\
\hline

OneCAT~\cite{li2025onecat} & \ding{55}
& 2.22 & 1.08 & 2.09
& 1.63 & 1.24 & 2.11
& 1.45 & 1.26 & 2.12
& 1.05 & 1.04 & 2.13
& - & - & -
& 1.75 & 1.05 & 2.00
& \cellcolor[HTML]{C8C8C8}1.12 \\

UniWorld-V1~\cite{lin2025uniworld} & $\checkmark$
& 1.41 & 3.09 & 3.34
& 1.40 & 3.51 & 3.13
& 1.30 & 3.31 & 3.40
& 1.02 & 1.92 & 3.80
& 1.66 & 2.38 & 2.77
& 1.25 & 3.00 & 2.57
& \cellcolor[HTML]{C8C8C8}1.21 \\

Lumina-DiMOO~\cite{xin2025lumina} & \ding{55}
& 2.21 & 1.97 & 2.34
& 1.59 & 2.21 & 2.38
& 1.47 & 2.73 & 2.86
& 1.02 & 2.49 & 2.89
& - & - & -
& 1.57 & 1.56 & 2.09
& \cellcolor[HTML]{C8C8C8}1.32 \\\

Nextstep-V1~\cite{han2025nextstep} & \ding{55}
& 3.22 & 1.77 & 2.28
& 2.33 & 1.74 & 2.25
& 1.70 & 1.81 & 2.25
& 1.10 & 1.04 & 2.07
& - & - & -
& 2.12 & 1.21 & 2.10
& \cellcolor[HTML]{C8C8C8}1.44 \\

InternVL-U~\cite{tian2026internvl} & \ding{55}
& 3.89 & 1.98 & 2.85
& 2.75 & 1.78 & 2.53
& 1.86 & 2.10 & 2.67
& 1.21 & 2.25 & 2.76
& - & - & -
& 2.41 & 1.36 & 2.30
& \cellcolor[HTML]{C8C8C8}1.59 \\

HiDream-E1~\cite{cai2025hidream} & \ding{55}
& 3.09 & 2.20 & 2.53
& 2.47 & 2.21 & 2.42
& 1.72 & 2.20 & 2.55
& 1.22 & 2.48 & 3.60
& - & - & -
& 2.20 & 1.30 & 2.08
& \cellcolor[HTML]{C8C8C8}1.61 \\

UniReason1.0~\cite{wang2026unireason} & $\checkmark$
& 2.94 & 2.50 & 2.76
& 2.73 & 2.91 & 2.59
& 2.16 & 2.82 & 2.76
& 1.24 & 3.18 & 2.77
& 1.81 & 1.70 & 2.84
& 2.18 & 1.47 & 2.24
& \cellcolor[HTML]{C8C8C8}1.67 \\

DeepGen 1.0~\cite{wang2026deepgen} & \ding{55}
& 3.71 & 1.96 & 2.76
& 2.80 & 1.98 & 2.45
& 2.16 & 2.31 & 2.85
& \color[HTML]{F88825}\textbf{1.53} & 4.32  & \color[HTML]{F88825}\textbf{4.50}
& - & - & -
& 2.68 & 1.50 & 2.24
& \cellcolor[HTML]{C8C8C8}1.70 \\

OmniGen2~\cite{wu2025omnigen2} & $\checkmark$
& 3.32 & 3.29 & 3.18
& 2.53 & 3.73 & 3.13
& 1.45 & 3.63 & 3.64
& 1.06 & 2.25 & 3.62
& 2.40 & 1.68 & 2.64
& 2.19 & 2.86 & 2.78
& \cellcolor[HTML]{C8C8C8}1.88 \\

ChronoEdit~\cite{wu2025chronoedit} & \ding{55}
& 2.74 & 3.52 & 3.42
& 2.84 & 3.70 & 3.16
& 1.66 & 3.43 & 3.15
& 1.10 & 1.83 & 3.34
& - & - & -
& 2.01 & 3.12 & 2.81
& \cellcolor[HTML]{C8C8C8}1.90 \\

Bagel-Think~\cite{deng2025emerging} & $\checkmark$
& 3.62 & 3.94 & 3.16
& 2.55 & 3.89 & 2.91
& 2.27 & 3.07 & 3.34
& 1.20 & 3.12 & 3.48
& 1.78 & 1.75 & 2.93
& 2.07 & 3.35 & 2.82
& \cellcolor[HTML]{C8C8C8}2.13 \\

Bagel~\cite{deng2025emerging} & $\checkmark$
& 3.68 & 3.99 & 3.11
& 2.43 & 3.82 & 2.87
& 1.78 & 3.59 & 3.38
& 1.20 & 3.28 & 3.83
& 1.86 & 2.03 & 3.06
& 2.30 & 2.99 & 2.65
& \cellcolor[HTML]{C8C8C8}2.15 \\

Unipic3~\cite{wei2026skywork} & $\checkmark$
& 4.28 & 4.11 & 3.65
& 3.37 & 3.94 & 3.09
& 2.16 & 3.51 & 3.16
& 1.15 & 2.39 & 3.56
& 2.76 & 2.86 & 2.64
& 2.76 & 2.98 & 2.74
& \cellcolor[HTML]{C8C8C8}2.47 \\

Step1X-Edit-v1p2~\cite{liu2025step1x} & \ding{55}
& 4.20 & 4.35 & 3.66
& 3.16 & 4.14 & 3.12
& 2.14 & \color[HTML]{F88825}\textbf{3.93} & 3.47
& 1.22 & 2.60 & 4.38
& - & - & -
& 2.92 & 3.32 & 2.79
& \cellcolor[HTML]{C8C8C8}2.53 \\

UniWorld-V2~\cite{li2025uniworld} & $\checkmark$
& 4.42 & 4.02 & 3.73
& 3.76 & 3.61 & 3.06
& 2.29 & 2.96 & 3.26
& 1.15 & 2.03 & 3.53
& 3.07 & 3.06 & 2.90
& 2.92 & 2.78 & 2.81
& \cellcolor[HTML]{C8C8C8}2.53 \\

EMU3.5~\cite{cui2025emu3} & $\checkmark$
& 4.48 & 4.00 & 3.68 & 3.81 & 3.66 & 3.19 & \color[HTML]{F88825}\textbf{2.54} & 3.51 & 3.39
& 1.42 & 3.59 & 3.94 & \color[HTML]{F88825}\textbf{3.39} & 3.36 & 2.84 & 2.71 & 3.31 & 2.89
& \cellcolor[HTML]{C8C8C8}2.63 \\

JoyAI-Image-Edit~\cite{joyaiimage2026} & \ding{55}
& 4.54 & 4.31 & 3.68
& 3.59 & 4.20 & 3.15
& 2.26 & 3.89 & 3.46
& 1.35 & 2.50 & 4.38
& - & - & -
& \color[HTML]{F88825}\textbf{3.02} & 3.31 & 2.90
& \cellcolor[HTML]{C8C8C8}2.63 \\

Longcat-Image-Edit~\cite{team2025longcat} & \ding{55}
& 4.53 & \color[HTML]{F88825}\textbf{4.47} & 3.85
& 3.58 & \color[HTML]{F88825}\textbf{4.26} & 3.36
& 2.11 & 3.88 & 3.53
& 1.33 & 3.37 & 4.36
& - & - & -
& 2.81 & 3.34 & 3.19
& \cellcolor[HTML]{C8C8C8}2.68 \\

Qwen-Image-Edit-2511~\cite{wu2025qwen} & $\checkmark$
& \color[HTML]{F88825}\textbf{4.61} & 4.44 & 3.75
& \color[HTML]{F88825}\textbf{3.94} & 4.05 & 3.17
& 2.38 & 3.54 & 3.36
& 1.28 & 2.62 & 4.25
& 3.26 & 3.47 & 2.84
& 2.80 & 3.69 & 3.05
& \cellcolor[HTML]{C8C8C8}{\color[HTML]{F88825}\textbf{ 2.73}} \\

\hline
% \noalign{\vskip 3pt}

Wan2.7-image~\cite{wan} & $\checkmark$
& 4.57 & 4.39 & 4.05
& 4.42 & 4.01 & 3.83
& 3.62 & 3.91 & 4.06
& 1.52 & 2.94 & 4.51
& 4.25 & 4.25 & 3.29
& 4.07 & 4.08 & \color[HTML]{319B62}\textbf{3.50}
& \cellcolor[HTML]{C8C8C8}3.21 \\

Seedream 4.5~\cite{seedream2025seedream} & $\checkmark$
& 4.62 & 4.40 & 4.05
& 4.39 & 4.05 & \color[HTML]{319B62}\textbf{3.84}
& 3.65 & 4.12 & 4.02
& 1.48 & 2.89 & 4.58
& 4.29 & 4.24 & 3.39
& 4.03 & 4.21 & 3.48
& \cellcolor[HTML]{C8C8C8}3.23 \\

Nano Banana 2~\cite{google2026gemini31flashimagepreview}  & $\checkmark$
& 4.75 & 4.52 & \color[HTML]{319B62}\textbf{4.15}
& \color[HTML]{319B62}\textbf{4.55} & 4.33 & 3.77
& 4.24 & 4.25 & 4.19
& 2.77 & 3.96 & 4.65
& 4.35 & 4.30 & \color[HTML]{319B62}\textbf{3.45}
& \color[HTML]{319B62}\textbf{4.49} & \color[HTML]{319B62}\textbf{4.38} & 3.46
& \cellcolor[HTML]{C8C8C8} 3.71  \\

Nano Banana Pro~\cite{gemini3pro} & $\checkmark$
& \color[HTML]{319B62}\textbf{4.81} & \color[HTML]{319B62}\textbf{4.69} & 4.10
& 4.49 & \color[HTML]{319B62}\textbf{4.55} & \color[HTML]{319B62}\textbf{3.84}
& \color[HTML]{319B62}\textbf{4.36} & \color[HTML]{319B62}\textbf{4.47} & \color[HTML]{319B62}\textbf{4.20}
& \color[HTML]{319B62}\textbf{3.49} & \color[HTML]{319B62}\textbf{4.19} & \color[HTML]{319B62}\textbf{4.72}
& \color[HTML]{319B62}\textbf{4.41} & \color[HTML]{319B62}\textbf{4.35} & 3.43
& 4.39 & 4.36 & 3.43
& \cellcolor[HTML]{C8C8C8}\color[HTML]{319B62}\textbf{3.95} \\

            \bottomrule
        \end{tabular}}
   
\end{table*}

\begin{table}[]
\centering
\caption{\textbf{Main results on {\rmbench}.} 
$^\dagger$ indicates Qwen2.5-VL-7B as the baseline, \textsuperscript{\S} indicates Qwen3-VL-8B as the baseline, and $^\ddagger$ denotes the thinking-enabled version.}
\label{tab:RMBench Main Result}
\resizebox{0.8\textwidth}{!}{%
\begin{tabular}{lcccc}
\toprule
\textbf{Method} & \textbf{Instruction Awareness} & \textbf{Visual Consistency} & \textbf{Visual Quality} & \textbf{AVG} \\
\midrule
\rowcolor{red4!20}
\multicolumn{5}{c}{\textit{Open-source Multimodel Large Language Models}} \\
Gemma 3 12B~\citep{Gemma3} & 0.4301 & 0.2871 & 0.2635 & 0.3799\\
Gemma 3 27B~\citep{Gemma3} & 0.5909 & 0.3300 & 0.2905 & 0.4996\\
Gemma 4 26B A4B~\citep{google2026gemma4} & 0.6947 & 0.3960 & 0.4392 & 0.5960 \\
Gemma 4 31B ~\citep{google2026gemma4} & 0.7527 & 0.5165 & 0.4932 & 0.6709 \\
Qwen2.5-VL-7B ~\citep{wang2024qwen2} & 0.4272 & 0.2165 & 0.3151 & 0.3621 \\
Qwen3-VL-4B ~\citep{bai2025qwen3} & 0.5209 & 0.2413 &0.3378 & 0.4322 \\
Qwen3-VL-8B ~\citep{bai2025qwen3} & 0.5646 & 0.2541 & 0.3446 & 0.4650 \\
Qwen3-VL-30B A3B ~\citep{bai2025qwen3} & 0.5633 & 0.3086 & 0.3378 & 0.4787 \\
Qwen3-VL-32B ~\citep{bai2025qwen3} & 0.6763 & 0.3960 & 0.3649 & 0.5790 \\

Qwen3.5-2B ~\citep{qwen3.5} & 0.4162 & 0.3279 & 0.2500 & 0.3811\\
Qwen3.5-2B$^\ddagger$ ~\cite{qwen3.5} & 0.4220 & 0.2804 & 0.4466 & 0.3848\\ % think

Qwen3.5-9B ~\citep{qwen3.5} & 0.585 & 0.452 & 0.372 & 0.535 \\
Qwen3.5-9B$^\ddagger$ ~\cite{qwen3.5} & 0.7615 & 0.4860 & 0.4898 & 0.6681\\ % think

Qwen3.5-27B ~\citep{qwen3.5} & 0.7322 & \color[HTML]{F88825}\textbf{0.5850} & 0.5381 & 0.6693\\
Qwen3.5-27B$^\ddagger$ ~\cite{qwen3.5} & 0.7674 & 0.5637 & \color[HTML]{F88825}\textbf{0.5878} & 0.6998\\% think

Qwen3.5-35B-A3B  ~\citep{qwen3.5} & 0.7279 & 0.5479 & 0.4205 & 0.6318\\
Qwen3.5-35B-A3B$^\ddagger$ ~\cite{qwen3.5} &  \color[HTML]{F88825}\textbf{0.8074} & 0.5073 & 0.5608 & 0.7089\\% think

Qwen3.6-27B ~\citep{qwen3.6-27b} & 0.7147 & 0.4966 & 0.4184 & 0.6328\\
Qwen3.6-27B$^\ddagger$ ~\cite{qwen3.6-27b} & 0.7961 & 0.5656 & 0.5743 & \color[HTML]{F88825}\textbf{0.7183}\\ % think

Qwen3.6-35B-A3B~\citep{qwen36_35b_a3b} & 0.6824 & 0.4558 & 0.4344 & 0.5995\\
Qwen3.6-35B-A3B$^\ddagger$~\citep{qwen36_35b_a3b} &0.7921  & 0.5300 & 0.5608 & 0.7051\\% think

\midrule

\rowcolor{blue4!20}
\multicolumn{5}{c}{\textit{Proprietary Models}} \\
GPT-4.1~\cite{gpt41} & 0.7471 & 0.4845 & 0.5338 & 0.6611 \\
Gemini 3 Flash ~\cite{gemini3pro}& 0.8042 & 0.5981 & 0.4865 & 0.7268 \\
Gemini 3.1 Pro~\cite{google2026gemini31propreview} & 0.8324 & 0.6002 & 0.4459 & 0.7433\\

\midrule

\rowcolor[HTML]{F5FFFA}
\multicolumn{5}{c}{\textit{Models Trained on Human Preference Pairs}} \\
EditReward$^\dagger$(Point-wise)~\cite{wu2025editreward} & 0.5524 & - & 0.6369 & 0.5601\\
EditReward$^\dagger$(Pair-wise)~\cite{wu2025editreward} & 0.5490 & - & 0.6301 & 0.5564\\
EditScore$^\dagger$~\cite{luo2025editscore} & 0.5092 & 0.4160 & 0.5890 & 0.4912\\
EditScore\textsuperscript{\S}~\cite{wu2025editreward} & 0.5736 & 0.5222 & 0.5616 & 0.5587 \\
\bottomrule
\end{tabular}%
}
\vspace{-0.5em}
\end{table}

\begin{table*}[t]
\centering
\caption{\textbf{System prompt ablation and visual perception analysis.}
(a) visual perception comparison on single-image and multi-image tasks.
(b) compares different system prompts on {\rmbench}, where colored $\uparrow$ denotes the absolute gain over the corresponding EditScore prompt.}
\label{tab:combined_results}

\begin{minipage}{0.38\textwidth}
\centering
(b) Visual perception analysis\\[0.3em]
\setlength{\tabcolsep}{4pt}
\small
\resizebox{\linewidth}{!}{%
\begin{tabular}{lcccc}
\toprule
Model & Movement & Swap & CP & MIA \\
\midrule
FLUX2.Dev       & 2.98 & 1.95 & 1.04 & 1.80 \\
Qwen-Image-Edit & 3.59 & 2.32 & 1.23 & 1.90 \\
Bagel           & 2.39 & 1.64 & 1.03 & 1.13 \\
EMU3.5          & 3.66 & 2.02 & 1.09 & 2.26 \\
\midrule
Seedream4.5     & 3.56 & 4.05 & 3.26 & 3.31 \\
Nano-Banana-Pro & \textbf{4.03} & \textbf{4.22} & \textbf{3.65} & \textbf{3.60} \\
\bottomrule
\end{tabular}%
}
\end{minipage}
\hfill
\begin{minipage}{0.58\textwidth}
\centering
(a) System prompt ablation \\[0.3em]
\resizebox{\linewidth}{!}{%
\begin{tabular}{lcccc}
\toprule
\textbf{Method} & \textbf{Instruction Awareness} & \textbf{Visual Consistency} & \textbf{Visual Quality} & \textbf{AVG} \\
\midrule
\rowcolor{blue4!20}
\multicolumn{5}{c}{\textit{System Prompts for EditScore}} \\
Qwen3-VL-4B ~\citep{bai2025qwen3} & 0.418 & 0.1849 & 0.3287 & 0.3500 \\
Qwen3-VL-8B ~\citep{bai2025qwen3} & 0.44 & 0.1198 & 0.2397 & 0.3415 \\
Qwen3-VL-32B ~\citep{bai2025qwen3} & 0.6344 & 0.1952 & 0.2876 & 0.4912 \\
\midrule

\rowcolor{red4!20}
\multicolumn{5}{c}{\textit{System Prompts for {\rmbench}}} \\
Qwen3-VL-4B ~\citep{bai2025qwen3} & 0.525(\gain{0.107}) & 0.2483(\gain{0.0634}) & 0.3356(\gain{0.0069}) & 0.4367(\gain{0.0885}) \\
Qwen3-VL-8B ~\citep{bai2025qwen3} & 0.5682(\gain{0.1282}) & 0.2583(\gain{0.1385}) & 0.3423(\gain{0.1026}) & 0.4684(\gain{0.1293}) \\
Qwen3-VL-32B ~\citep{bai2025qwen3} & 0.6786(\gain{0.0442}) & 0.3944(\gain{0.1992}) & 0.363(\gain{0.0754}) & 0.5800(\gain{0.0888}) \\
\bottomrule
\end{tabular}%
}
\end{minipage}

\vspace{-1em}
\end{table*}

\subsection{Analysis and Findings for {\bench}}

\textbf{Human-Aligned Evaluation Protocol.}
We evaluate the reliability of our evaluation protocol from both benchmark-level and model-level perspectives. At the benchmark level, we sample instances from ImgEdit-Bench~\cite{ye2025imgedit}, GEdit-Bench~\cite{liu2025step1x}, RISE-Bench~\cite{zhao2025envisioning}, and {\bench}. We generate edited results using OmniGen2 and ask human experts to provide preference rankings. As shown in Figure~\ref{User_Study}(b), {\bench} achieves stronger agreement with human preferences than existing benchmarks.
At the model level, we randomly sample test instances, collect editing results from different models~\cite{wu2025qwen,flux-2-2025,nanobananapro,deng2025emerging}, and compute the Pearson correlation between human ratings and MLLM-based scores. The resulting correlation of Figure~\ref{User_Study}(a) demonstrates that our protocol provides a reliable automatic evaluation tool for image editing.

\textbf{Visual Perception Ability.}
Table~\ref{tab:combined_results}(a) reports the performance of representative models on visual perception tasks. Open-source models perform well on relatively basic tasks such as Object Movement, but their performance drops notably on more complex tasks, especially Object Swap and Complex Paint. This gap is particularly clear in Complex Paint, where models must interpret both textual instructions and in-image visual annotations. For multi-image perception, closed-source models show a clear advantage, indicating that cross-image understanding remains a major challenge for open-source models.

\textbf{Algorithmic Visual Reasoning Ability.}
Table~\ref{tab:Algorithm Visual Reasoning} provides a detailed comparison across Algorithmic Visual Reasoning sub-tasks. The results show that open-source models still struggle to perform visual reasoning and faithfully execute the derived edits, leading to poor performance in this category. Although closed-source models show some potential on certain sub-tasks, their overall performance remains limited, indicating that algorithmic visual reasoning is still a major challenge for current image editing models.

\textbf{Cross-Lingual Performance.}
Some models~\cite{li2025uniworld,labs2025flux} show clear cross-lingual imbalance, performing better under English instructions than Chinese ones. In contrast, many advanced unified models, including both open-source and closed-source systems, exhibit only marginal differences between the two languages. This suggests that robust cross-lingual image editing requires strong language-vision understanding and balanced multilingual training data.

\subsection{Analysis and Findings for {\rmbench}}

\textbf{Impact of System Prompts.}
Table~\ref{tab:combined_results}(b) compares the system prompts used in {\rmbench} with those adopted by EditScore across different models. For a fair comparison, we evaluate them on the corresponding single-image subsets of {\rmbench}. Our prompts consistently improve performance across all evaluation dimensions, with the largest gain of $12.93$\% on Qwen3-VL-8B~\cite{bai2025qwen3}.

\textbf{Effect of Thinking-Enabled Inference.}
We further study the effect of thinking-enabled inference on reward model evaluation. As shown in Table~\ref{tab:RMBench Main Result}, enabling thinking consistently improves performance. Among medium-sized dense models, Qwen3.5-9B~\cite{qwen3.5} achieves the largest gain, improving by $9.83$ points over its non-thinking counterpart. Among sparse MoE models, Qwen3.6-35B-A3B~\citep{qwen36_35b_a3b} shows the greatest improvement, with a gain of $10.56$ points.

\section{Conclusion, Discussion, and Limitations}
\label{sec:conclusion}
We introduce {\bench} and {\rmbench}, a unified benchmark suite for evaluating frontier image editing systems and reward models. {\bench} includes 2,388 carefully annotated instances across 36 progressively challenging task categories, covering general editing, world knowledge reasoning, visual reasoning, dynamic manipulation, and multi-image editing. It further adopts a fine-grained multidimensional evaluation framework with structured reasoning and scoring rubrics. Complementarily, {\rmbench} provides 2,251 preference pairs that simulate realistic reward modeling scenarios during RL optimization. Extensive evaluations on 29 image editing models and 21 reward models reveal substantial gaps between proprietary and open-source systems, persistent weaknesses in reasoning-intensive and multi-image editing, and the potential of native multimodal large language models as reward models.
A limitation of our current evaluation protocol is its reliance on API-based MLLM judges. Although our structured rubrics improve interpretability and alignment with human judgments, the scores may still be affected by judge capabilities and version updates, limiting accessibility. In future work, we plan to develop a dedicated image-editing judge model for more stable and transparent evaluation without proprietary APIs.

\section{Acknowledgements}
We sincerely thank Zhuoran Zhang, Qixun Wang, Yuqi Tang, Tengfei Liu, Haotian Wang, Bohan Zeng, Xinlong Chen, Yue Ding, Chengzhuo Tong, Bozhou Li, Ruizhe Chen, Shilin Yan, Xuelong Li, Yunshu Wang, Huanyu Zhang, Dianyi Wang, Liuling Dong, Siqi Yin, Saikun Sun, Jiafeng Chen, and Shengqi Wu for their support of \bench\ and \rmbench. 

\bibliographystyle{plain}

\bibliography{neurips_2026}

\newpage

\appendix

% \section{Appendix}
\section{{\bench} Data Construction}
\label{supp: Edit-Compass Data Construction}
As shown in Figure~\ref{Fig: Data Construction}, the construction of {\bench} consists of three main components.

\subsection{General and  Complex tasks.}
For the General and Complex task categories, multiple human experts collect real, high-quality images from Unsplash, Pexels, Pixabay, and Freepik under permissive licenses. The collected images are then reviewed by five human experts from multiple perspectives, including safety, image quality, and suitability for editing. An image is retained only if all reviewers vote to approve it. To generate diverse editing instructions, we establish an instruction generation platform based on Gemini 3 Pro~\cite{gemini3pro} and GPT-5.1~\cite{openai2025gpt51}. For each image and each task category, the two models generate three candidate editing instructions. Human experts then select the final instruction to ensure feasibility, clarity, and alignment with the intended editing task.

\subsection{Dynamic Manipulation, World Knowledge Reasoning, and Multi-Image Tasks}
For Dynamic Manipulation, World Knowledge Reasoning, and Multi-Image tasks, experts in image editing first conduct in-depth discussions to define each task and construct a coarse-grained description of the source image together with the corresponding editing instruction. The coarse-grained source-image description is then refined by Gemini 3 Pro~\cite{gemini3pro} and used to generate the source image. After generation, multiple human experts assess whether each image--instruction pair is feasible and valid for evaluation. A case is included in the benchmark only if it is approved by all human experts.

\subsection{Algorithmic Visual Reasoning tasks.}
\label{Algorithm Visual Reasoning tasks}
For Algorithmic Visual Reasoning tasks, multiple human experts first discuss and define the target editing problems. We then render the source images using Python programs, which also provide the corresponding ground-truth annotations. Based on the rendered images and annotations, human experts design editing instruction templates for each task category. These templates are then applied to the corresponding cases, ensuring that each instance has a well-defined visual structure and an unambiguous expected outcome.  The detailed design of each task category is described below.

\subsubsection{Longest Word Discovery}
\label{supp:Longest Word Discovery}

\textbf{Task Definition.} Given a letter grid \( G \in \Sigma^{n \times n} \) and a fixed starting cell
\( s = (r_0, c_0) \), we seek the longest valid English word that can be formed
by a monotone traversal starting at \( s \). At each step, the traversal may
move only downward or rightward, i.e.,
\[
(r,c) \rightarrow (r+1,c)
\quad \text{or} \quad
(r,c) \rightarrow (r,c+1).
\]

Let \( \mathcal{P}(s) \) be the set of all such paths starting from \( s \).
For each \( p \in \mathcal{P}(s) \), let \( \mathrm{str}(p) \) denote the string
obtained by concatenating the letters visited along \( p \). We define
\[
\mathcal{W}(G,s) = \{\, \mathrm{str}(p) \mid p \in \mathcal{P}(s),\ \mathrm{str}(p)
\text{ is a valid English word} \,\}.
\]
The target word is then
\begin{equation}
w^* = \arg\max_{w \in \mathcal{W}(G,s)} |w|.
\end{equation}

To mitigate character-level inaccuracies introduced by generative models, we
develop a Python-based image reconstruction pipeline. Compared with directly
rendering text through generative models, this pipeline provides precise
control over textual content, spatial layout, and visual attributes, while
also avoiding the ambiguity commonly associated with text appearing in natural
scene images. Overall, the pipeline consists of four components:

\textbf{Word source and selection strategy.}
The target words are sampled from a publicly available large-scale English lexicon containing several hundred thousand entries. 
We use this resource as the base vocabulary to ensure broad lexical coverage and standardized word forms. To make the task more challenging, we focus on words with relatively long character spans and then perform manual curation,
including cleaning, deduplication, and final verification of the candidate
list. We do not explicitly control for word frequency or semantic domain;
instead, preference is given to words exhibiting greater morphological or
semantic complexity, such as those containing multiple roots, prefixes,
suffixes, or compound-like structures. This procedure yields a final target
set with high diversity and increased difficulty.

\textbf{Data Generation.}
Given the curated target vocabulary, we generate each sample by embedding one
target word into a letter grid. Specifically, for each word, we randomly
select a valid starting position and place the word character by character
along a monotone path that permits only downward or rightward moves. 
The remaining unfilled cells are populated with randomly sampled uppercase letters. 
In addition, to suppress the accidental formation of valid words that
extend beyond the intended target, we apply a local blocking mechanism near
the terminal position of the embedded word, preferentially assigning
low-frequency letters such as \texttt{X}, \texttt{Z}, \texttt{Q}, and
\texttt{J}. As a result, each sample is paired with both an input grid image
that marks the starting position and a result image that annotates the full
ground-truth path, together with structured metadata containing the grid
configuration, the start location, the target word, and its corresponding
path.

\textbf{Verification.}
To ensure correctness and uniqueness, we perform an explicit verification step
after generating each instance. Using the definition of $\mathcal{W}(G,s)$ from Appendix ~\ref{supp:Longest Word Discovery}, the verifier exhaustively searches all valid words reachable from
the designated start cell $s$ under the Down/Right constraint, while using a
trie to prune invalid prefixes early. Let $\hat{w}$ denote the longest word
returned by Algorithm~\ref{alg:verify_longest_word}. We accept the generated
sample if and only if
\begin{equation}
    \hat{w} = w_{\text{target}}.
\end{equation}
This guarantees that the target word is reachable from the specified start cell
and that no longer valid word exists under the same movement constraint.

\begin{algorithm}[t]
\caption{Trie-based verification from a fixed start cell}
\label{alg:verify_longest_word}
\begin{algorithmic}[1]
\Require Letter grid $G \in \Sigma^{n \times n}$, start cell $s=(r_0,c_0)$, dictionary trie $\mathcal{T}$
\Ensure Longest valid word $w^*$ reachable from $s$

\State $\textit{bestWord} \gets \emptyset$
\State $\textit{bestLen} \gets 0$

\If{$G[r_0,c_0] \notin \mathrm{Children}(\mathcal{T}.\mathrm{root})$}
    \State \Return $\emptyset$
\EndIf

\Function{DFS}{$r,c,v,w$}
    \If{$v.\mathrm{isEnd} = \mathrm{True}$ \textbf{and} $|w| > \textit{bestLen}$}
        \State $\textit{bestWord} \gets w$
        \State $\textit{bestLen} \gets |w|$
    \EndIf

    \ForAll{$(r',c') \in \{(r+1,c), (r,c+1)\}$}
        \If{$(r',c')$ is inside the grid}
            \State $ch \gets G[r',c']$
            \If{$ch \in \mathrm{Children}(v)$}
                \State \Call{DFS}{$r',c',\mathrm{Child}(v,ch),w \circ ch$}
            \EndIf
        \EndIf
    \EndFor
\EndFunction

\State $ch_0 \gets G[r_0,c_0]$
\State \Call{DFS}{$r_0,c_0,\mathrm{Child}(\mathcal{T}.\mathrm{root},ch_0),ch_0$}
\State \Return $\textit{bestWord}$
\end{algorithmic}
\end{algorithm}

\subsubsection{Global Longest Word Discovery}

We introduce Global Longest Word Discovery as a more challenging variant of the standard word discovery task. In this task, the target is required to be the longest valid English word that can be formed from any position in the grid under the same down/right movement constraint. Formally, given a constructed grid, we perform a dictionary-based search over all admissible paths to identify the longest valid word. A sample is retained only when the identified word exactly matches the embedded target. This verification procedure ensures that the ground-truth solution is unique and well-defined.

\subsubsection{Knapsack Selection}

\textbf{Task Define.} We introduce a visual editing task inspired by the 0--1 knapsack problem. Given
an image containing a set of visual objects, each associated with a value and
a cost, together with a budget constraint, the objective is to select a subset
of objects that maximizes the total value while ensuring that the total cost
does not exceed the budget.

Formally, let \(\{o_i\}_{i=1}^n\) denote the set of objects in the image,
where each object \(o_i\) is associated with a value \(v_i \in \mathbb{R}_+\)
and a cost \(c_i \in \mathbb{R}_+\), and let \(B \in \mathbb{R}_+\) be the
budget. The goal is to solve
\begin{equation}
S^* = \arg\max_{S \subseteq \{1,\dots,n\}} \sum_{i \in S} v_i
\quad \text{s.t.} \quad
\sum_{i \in S} c_i \le B.
\end{equation}

The model must infer \(S^*\) from the visual input and perform consistent edits.

\textbf{Data Generation.}
We generate data in a programmatic manner. For each sample, we first sample the
knapsack capacity \(C \sim \mathcal{U}(10,20)\) and the number of items
\(N \sim \mathcal{U}(6,8)\). For each item \(i\), we then independently sample
its weight and value as \(w_i \sim \mathcal{U}(2,8)\) and
\(v_i \sim \mathcal{U}(10,50)\), respectively. We use dynamic programming to compute the optimal subset of items for each
instance. Specifically, we define a two-dimensional state table \(dp[i][w]\),
which denotes the maximum achievable value when considering the first \(i\)
items under capacity \(w\). The state transition is given by
\begin{equation}
dp[i][w] =
\begin{cases}
\max\bigl(dp[i-1][w],\; dp[i-1][w-w_i] + v_i\bigr), & \text{if } w_i \le w,\\
dp[i-1][w], & \text{otherwise}.
\end{cases}
\end{equation}
After filling the table, we recover the corresponding optimal item subset by
standard backtracking.

\subsubsection{Optimal Path Identification}

\textbf{Task Definition.}
\textit{Optimal Path Identification} defines a shortest-path image editing task on a two-dimensional grid environment $\mathcal{G} \in \mathbb{Z}^{H \times W}$. Each instance contains a start point $s$, an end point $t$, and a grid-based scene where each cell $(i,j)$ is associated with a terrain type and a corresponding traversal cost. The task is to identify a valid path $\pi$ from $s$ to $t$ with minimum total cost under 4-neighbor connectivity, and to express the solution as a structured edit by overlaying directional arrows on the image. Formally, the optimal path is defined as:
\[
\pi^\star = \arg\min_{\pi \in \Pi(s,t)} \sum_{(i,j) \in \pi} c(i,j),
\]
where $\Pi(s,t)$ denotes the set of all valid paths from $s$ to $t$ under 4-neighbor connectivity, and $c(i,j)$ denotes the traversal cost of cell $(i,j)$.

\begin{algorithm}[t]
\caption{Verification via Dijkstra Search}
\label{alg:dijkstra_verify}
\begin{algorithmic}[1]
\Require Terrain grid \(G \in \mathbb{Z}^{n \times m}\), start cell \(s\), end cell \(t\), terrain cost map \(c(\cdot)\)
\Ensure Optimal path \(P^*\) and minimum cost \(d^*\)

\State Initialize distance array \(D[r,c] \leftarrow +\infty\) for all cells
\State Set \(D[s] \leftarrow 0\)
\State Initialize parent map \(\mathrm{par}[\cdot] \leftarrow \varnothing\)
\State Initialize priority queue \(Q \leftarrow \{(0, s)\}\)

\While{\(Q\) is not empty}
    \State Pop the state \((d, u)\) with the smallest distance from \(Q\)
    \If{\(u = t\)}
        \State \textbf{break}
    \EndIf
    \If{\(d > D[u]\)}
        \State \textbf{continue}
    \EndIf
    \For{each 4-neighbor \(v\) of \(u\)}
        \If{\(v\) is outside the grid or blocked}
            \State \textbf{continue}
        \EndIf
        \State \(\hat{d} \leftarrow D[u] + c(v)\)
        \If{\(\hat{d} < D[v]\)}
            \State \(D[v] \leftarrow \hat{d}\)
            \State \(\mathrm{par}[v] \leftarrow u\)
            \State Push \((\hat{d}, v)\) into \(Q\)
        \EndIf
    \EndFor
\EndWhile

\If{\(D[t] = +\infty\)}
    \State \Return \(\varnothing, \varnothing\)
\EndIf

\State Recover the path \(P^*\) by backtracking from \(t\) using \(\mathrm{par}[\cdot]\)
\State Reverse \(P^*\)
\State \Return \(P^*, D[t]\)
\end{algorithmic}
\end{algorithm}

\textbf{Data Generation.}
Each instance is generated procedurally. We first sample a square grid with
side length uniformly drawn from 6 to 10. Each cell is assigned a terrain type
from \{road, grass, water, wall\} according to a fixed categorical
distribution, with traversal costs 1, 3, 8, and \(+\infty\), respectively. We
then sample two non-wall cells as the start and end points, subject to a
minimum Manhattan-distance constraint. Given the resulting grid, we compute the
minimum-cost path on the 4-neighbor graph using Dijkstra's algorithm and
discard instances with no feasible path. For each valid instance, we render an
input image with the terrain map and endpoint markers, together with a target
image annotated by the optimal path.

\subsubsection{Convex Hull Identification}
\textbf{Task Definition.}
Convex Hull Identification is a visual reasoning task. Given a set of points in
a 2D plane, the goal is to identify the subset of points that forms the convex
hull and recover the polygon enclosing all points. Formally, let
\(P = \{p_i\}_{i=1}^n \subset \mathbb{R}^2\) denote a set of points. The convex
hull of \(P\), denoted by \(\mathrm{conv}(P)\), is defined as
\begin{equation}
\mathrm{conv}(P) =
\left\{
\sum_{i=1}^n \lambda_i p_i \;\middle|\;
\lambda_i \ge 0,\;
\sum_{i=1}^n \lambda_i = 1
\right\}.
\end{equation}
Equivalently, \(\mathrm{conv}(P)\) is the smallest convex set containing all
points in \(P\). 

\textbf{Data Generation.}
Each instance is generated procedurally. We first sample the number of points
as \(n \sim \mathcal{U}\{8,15\}\). We then independently sample \(n\) point
coordinates on a \(10 \times 10\) canvas, with each coordinate drawn uniformly
from \([1,9]\) to avoid points lying too close to the boundary.

\subsubsection{Maximum Submatrix Sum Identification}
\textbf{Task Definition.}
Given a 2D grid of integers and a fixed kernel size, the task is to identify
the rectangular submatrix with the maximum sum of values.
Formally, let \(G \in \mathbb{Z}^{n \times n}\) denote the input grid and let
\((k_h, k_w)\) denote the kernel size. The objective is to find the top-left
coordinate \((r, c)\) that maximizes
\begin{equation}
(r^*, c^*) = \arg\max_{r, c}
\sum_{i=0}^{k_h-1} \sum_{j=0}^{k_w-1} G[r+i,\, c+j],
\end{equation}
subject to \(0 \le r \le n-k_h\) and \(0 \le c \le n-k_w\).

\textbf{Data generation.}
Each input image is generated procedurally. We first sample the grid size
\(n \sim \mathcal{U}\{6,10\}\) and construct a square grid
\(G \in \mathbb{Z}^{n \times n}\), where each entry is independently sampled
from a discrete uniform distribution over \([-9,9]\). We then sample the kernel
dimensions \(k_h, k_w \sim \mathcal{U}\{2, \min(4, n-1)\}\), ensuring that the
kernel fits within the grid. The grid is rendered as integers (positive in black, negative in red), with the
kernel size specified in the instruction and no additional annotation.

\subsubsection{Maximum Bonus Identification }
\textbf{Task Definition.}
Maximum Bonus Identification is the task of identifying a path with
maximum total reward in a 2D integer grid, given a designated start and end
cell. Formally, let \(G \in \mathbb{Z}^{n \times n}\) denote the grid, and let
\(s = (r_1, c_1)\) and \(t = (r_2, c_2)\) denote the start and end cells,
with \(r_2 \ge r_1\) and \(c_2 \ge c_1\). Let \(\mathcal{P}(s,t)\) denote the
set of all monotone paths from \(s\) to \(t\). The objective is to find
\begin{equation}
P^* = \arg\max_{P \in \mathcal{P}(s,t)} \sum_{(r,c)\in P} G[r,c].
\end{equation}

\textbf{Data generation.}
Each input image is generated procedurally. We first sample the grid size
\(n \sim \mathcal{U}\{5,12\}\) and construct a grid
\(G \in \mathbb{Z}^{n \times n}\), where each entry is independently sampled
from \([-5,9]\). We then sample a start cell \(s\) and an end cell \(t\) such
that \(t\) lies in the bottom-right region of \(s\) and their Manhattan
distance is at least 3. The grid is rendered as a table of integers, with positive values in black and
negative values in red. The start and end cells are highlighted and provided
as part of the visual input, with no additional annotation.

\textbf{Ground-truth Construction.}
The ground-truth solution is computed by dynamic programming on the subgrid
induced by \(s=(r_1,c_1)\) and \(t=(r_2,c_2)\). Let
\(F(i,j)\) denote the maximum achievable reward from \(s\) to cell \((i,j)\),
where \(r_1 \le i \le r_2\) and \(c_1 \le j \le c_2\). The recurrence is
\begin{equation}
F(i,j)=
\begin{cases}
G[r_1,c_1], & (i,j)=(r_1,c_1),\\[4pt]
F(i-1,j)+G[i,j], & j=c_1,\ i>r_1,\\[4pt]
F(i,j-1)+G[i,j], & i=r_1,\ j>c_1,\\[4pt]
\max\bigl(F(i-1,j),\,F(i,j-1)\bigr)+G[i,j], & \text{otherwise}.
\end{cases}
\end{equation}
The optimal reward is given by \(F(r_2,c_2)\), and the corresponding path is
recovered by standard backtracking from \(t\).

\subsubsection{Numberlink Path Identification}
\textbf{Task Definition.}
Numberlink Path Identification is defined on a 2D grid with multiple pairs of
colored endpoints. The goal is to construct one path for each pair of
same-colored endpoints, such that all paths are pairwise non-intersecting.

\textbf{Data generation.} Each instance is generated procedurally on an \(n \times n\) grid, with
\(n \sim \mathcal{U}\{6,9\}\). We first sample the number of path pairs
\(m \sim \mathcal{U}\{3,5\}\). Starting from an empty grid, we then construct
\(m\) non-overlapping paths sequentially. For each path, an unoccupied cell is
randomly chosen as the starting point, and the path is extended by a
self-avoiding random walk over unoccupied 4-neighbor cells. A candidate path
is accepted only if its length exceeds a predefined minimum; once accepted, all
cells on that path are marked as occupied and removed from subsequent path
generation. After all paths are constructed, only the two endpoints of each path are
retained in the input image and rendered as color-matched circles, while the
intermediate cells are hidden. 

\subsubsection{Word Path Recovery}
\textbf{Task Definition.}
Word Path Recovery is defined as the task of locating a given
target word in a 2D letter grid and recovering its corresponding path, under
the constraint that each step proceeds either downward or rightward from the
first character. Formally, let \(G \in \Sigma^{n \times n}\) denote a grid of uppercase letters,
and let \(w = (w_1, \dots, w_\ell)\) denote a target word of length \(\ell\).
A valid path for \(w\) is a sequence of grid cells
\[
P = \bigl((r_1,c_1), \dots, (r_\ell,c_\ell)\bigr)
\]
such that
\[
G[r_i,c_i] = w_i, \qquad i=1,\dots,\ell,
\]
and for each \(i=1,\dots,\ell-1\),
\[
(r_{i+1},c_{i+1}) \in \{(r_i+1,c_i),\, (r_i,c_i+1)\}.
\]
The objective is to recover a valid path \(P^*\) corresponding to the given
target word.

\textbf{Data generation.}
Each instance is generated procedurally from a predefined English vocabulary.
To control task difficulty, we sample target words with varying lengths, where
shorter words generally yield easier instances and longer words induce larger
search spaces and more challenging path recovery. For a target word \(w\), we
first sample the grid size as a function of its length and then randomly choose
a valid starting position such that a monotone down/right path of length
\(\ell\) can be embedded in the grid. The characters of \(w\) are then placed
sequentially along a randomly sampled monotone path.
All remaining cells are filled with randomly sampled uppercase letters. To
reduce the probability of unintended continuations beyond the target word, we
apply a local blocking strategy near the terminal position by preferentially
placing low-frequency letters such as \texttt{X}, \texttt{Z}, \texttt{Q}, and
\texttt{J}. The resulting input image is rendered as a letter grid, and the
ground-truth output highlights the cells along the target-word path.

\subsubsection{Global Word Path Recovery}
We introduce  Global Word Path Recovery, a more challenging variant of the standard Word Path Recovery task, where the initial letter of the target word is no longer provided. Compared with the original setting, this variant requires the model to infer the complete word path without explicit starting-letter guidance. We follow the same generation strategy as in the previous task to ensure data validity while maintaining a balanced level of difficulty.

\section{{\rmbench} Data Construction}
\label{supp: EditReward-Compass Data Construction}
\begin{table}[t]
\centering
\caption{Reward Model Benchmark Sampling Configuration.}
\label{tab:reward_model_benchmark_sampling_config}
\resizebox{\textwidth}{!}{
\begin{tabular}{l l c c c c c}
\toprule
\textbf{Model} & \textbf{Sampling Method} & \textbf{Group} & \textbf{Noise Level} & \textbf{Infer Timesteps} & \textbf{Sample Images} & \textbf{Supp. Multi} \\
\midrule
Bagel-Think~\cite{deng2025emerging}     & Same Model & 12 & 0.7 & 30 & 17,736 & \icoyes \\
Bagel~\cite{deng2025emerging}           & Same Model & 12 & 0.7 & 30 & 17,736 & \icoyes \\
Flux-Kontext~\cite{labs2025flux}    & Same Model & 16 & 0.8 & 21 & 20,656 & \icono  \\
Qwen-Image-Edit~\cite{wu2025qwen} & Same Model & 16 & 0.3 & 20 & 23,648 & \icoyes \\
OmniGen2~\cite{wu2025omnigen2}        & Same Model & 16 & --  & 36 & 23,648 & \icoyes \\
Uniworld-V2~\cite{li2025uniworld}     & Same Model & 16 & 0.7 & 15 & 23,648 & \icoyes \\
\bottomrule
\end{tabular}
}
\end{table}

We construct the raw data of {\rmbench} using a stratified sampling strategy. For tasks that require world-knowledge reasoning and algorithmic visual reasoning, current open-source image editing models often produce relatively weak outputs, making it difficult to form reliable and informative preference pairs from a single model. Therefore, we sample candidate edited images from a diverse pool of image editing models, including Bagel~\cite{deng2025emerging}, EMU3.5~\cite{cui2025emu3}, Flux2-dev~\cite{flux-2-2025}, LongCat-Image-Edit~\cite{team2025longcat}, NextStep-1-HF~\cite{han2025nextstep}, OmniGen2~\cite{wu2025omnigen2}, Qwen-Image-Edit~\cite{wu2025qwen}, Uniworld-V1~\cite{lin2025uniworld}, Nano Banana Pro~\cite{nanobananapro}, and Nano Banana 2~\cite{google2026gemini31propreview}. Human experts then manually inspect and filter these candidates to construct valid preference pairs with clear quality differences.
In addition to model-diverse sampling, we aim to evaluate reward models under conditions closer to their actual use in RL-based image editing. To this end, we simulate the candidate distributions encountered by reward models during RL optimization using FlowGRPO~\cite{liu2025flow}. As illustrated in Table~\ref{tab:reward_model_benchmark_sampling_config}, we explore different sampling configurations for different models, so as to obtain clear and high-quality candidate images while preserving the diversity of model behaviors. This RL-like sampling process allows {\rmbench} to better reflect the practical scenario where reward models must compare multiple candidate edits generated during policy optimization.

\section{Detailed Design of {\bench} Categories}
{\bench} consists of six distinct categories of image editing tasks. Figure~\ref{fig:gallery} provides representative examples for each category. We provide detailed definitions and descriptions of each task in the following sections.

\textbf{General Tasks}
\begin{itemize}[leftmargin=2.2em, labelsep=0.4em, itemsep=2pt]
    \item \textbf{Subject Addition:}  This task aims to insert a subject into a source image under specified spatial and semantic constraints. We consider three variants: adding an object at a specified location without explicit attribute requirements; adding an object with specified attributes at a designated location; and copy-based subject addition, where the inserted subject is copied from the source image and placed at a new location.
    \item \textbf{Subject Remove: } This task aims to remove a subject from a source image under specified semantic, attribute, and spatial constraints. We consider three variants: single-subject removal, where the source image contains only one corresponding object to be removed; attribute-guided removal, where multiple objects satisfying a specified attribute must be removed simultaneously; and spatially guided removal, where multiple identical or similar objects appear at different locations and the target object is removed according to the specified spatial constraint.
    \item \textbf{Subject Replace:} This task replaces a specified subject in a source image with another subject while preserving the surrounding context. We consider two variants: common-subject replacement, where the replacement subject is a common object or concept; and knowledge-guided replacement, where the replacement subject requires world knowledge or fine-grained semantic understanding, such as Kobe Bryant’s jersey.
    \item \textbf{Subject Extract: } This task aims to extract a specified subject from a source image while preserving its original visual and geometric properties. The extracted subject is isolated on a white background, with its position, orientation, and size kept unchanged.
    \item \textbf{Change Color \& Size: }  This task consists of two sub-tasks that modify the visual attributes of a specified subject, including changing its color and adjusting its size.
    \item \textbf{Change Material:} This task changes the material of a specified subject in a source image. The edited subject should reflect the target material properties while preserving its original shape, structure, and the surrounding context.
    \item \textbf{Style Transfer: }  This task aims to change the style of a specified subject in a source image while preserving its identity and structural properties. The edited subject should exhibit the target style consistently, and the task covers more than 60 distinct styles.
    \item \textbf{Change Background: } This task changes the background of a source image while keeping the foreground subject unchanged. The edited image should preserve the subject’s identity and appearance while reflecting the new background consistently.
    \item \textbf{Visual Text Editing(cn):}  This task aims to edit Chinese text in an image through a variety of operations, including text insertion, text removal, font style modification, color modification, and character replacement.
    \item \textbf{Visual Text Editing(en): } This task aims to edit English text in an image through a variety of operations, including text insertion, text removal, font style modification, color modification, and character replacement.
\end{itemize}

\textbf{Dynamic Manipulation Tasks}
\begin{itemize}[leftmargin=2.2em, labelsep=0.4em, itemsep=2pt]
    \item \textbf{Object Movement:}  This task aims to change the spatial position of a specified subject in a source image while preserving its identity, appearance, and structural properties.
    \item \textbf{Object Swap: } This task swaps two specified subjects in a source image, including their spatial positions and relevant attributes.
    \item \textbf{Object Interaction: } This task aims to edit a source image by introducing interactions among multiple specified subjects.
    \item \textbf{Change Emotion: } This task aims to modify the emotion of a specified subject in a source image while preserving its identity, appearance, and structural properties.  
    \item \textbf{Action: } This task changes the action of a specified object or person in a source image.
\end{itemize}

\textbf{World Knowledge Reasoning Tasks}
\begin{itemize}[leftmargin=2.2em, labelsep=0.4em, itemsep=2pt]
    \item \textbf{Temporal Reasoning:} This task requires the model to understand how temporal changes would affect a subject in the source image. It includes two variants: predicting how the subject would change over time, and inferring what the subject may have looked like in the past from its current appearance. 
    \item \textbf{Causal Reasoning: } This task requires the model to reason about how a subject in the source image would change under given conditions or external factors.
    \item \textbf{Math Reasoning: } This task requires applying math-domain knowledge to edit an image.
    \item \textbf{Chemical Reasoning: }  This task requires applying chemical-domain knowledge to edit an image.
    \item \textbf{Game Reasoning: } This task requires applying game-domain knowledge to edit an image.
\end{itemize}

\textbf{Algorithm Visual Reasoning tasks}
A detailed definition of this task type is provided in Section.~\ref{Algorithm Visual Reasoning tasks}.

\textbf{Multi-Image Tasks}
\begin{itemize}[leftmargin=2.2em, labelsep=0.4em, itemsep=2pt]
    \item \textbf{Multi-Image Awareness:}  This task involves reasoning over multiple input images and transferring various attributes of a subject from the reference image(s) to the source image. These attributes may include action, color, function, and other relevant properties. 
    \item \textbf{Multi-Image Composition: } This task aims to compose a coherent image from multiple input images. Different input images may provide different visual elements, such as subjects, background scenes, or human figures.
    \item \textbf{Virtual Try-On:} This task transfers the garment from a reference image onto the person in a source image while preserving the person's identity and pose.
\end{itemize}

\textbf{Complex Tasks}
\begin{itemize}[leftmargin=2.2em, labelsep=0.4em, itemsep=2pt]
    \item \textbf{Complex Instruction:}  This task involves editing multiple objects in a source image based on a composite instruction, which integrates multiple tasks from General, Dynamic Manipulation, and World Knowledge Reasoning categories.
    \item \textbf{Complex Paint(en):} This task requires the model to understand multimodal signals embedded in the source image, including English text instructions and visual annotations such as arrows, circles, and cross (“X”) marks, and to perform the intended edits accordingly.
    \item \textbf{Complex Paint(cn):} This task requires the model to understand multimodal signals embedded in the source image, including Chinese text instructions and visual annotations such as arrows, circles, and cross (“X”) marks, and to perform the intended edits accordingly.
\end{itemize}

\section{Image Editing Model Evaluation}
\label{supp:Evaluation Details}
% \subsection{Image Editing Model Evaluation}
\textbf{Evaluation Models.}  We evaluate 29 mainstream image editing models, covering both open-source and closed-source models, as well as Chinese and English variants.

\hspace*{1em} \noindent\textbf{(1) Diffusion Models:} InstructPix2Pix~\cite{brooks2023instructpix2pix}, MagicBrush~\cite{zhang2023magicbrush}, AnyEdit~\cite{yu2025anyedit}, UltraEdit~\cite{zhao2024ultraedit}, FLUX.2~\cite{flux-2-2025}. Specifically, InstructPix2Pix~\cite{brooks2023instructpix2pix}, MagicBrush~\cite{zhang2023magicbrush}, and AnyEdit adopt UNet-based architectures built upon Stable Diffusion. In contrast, UltraEdit~\cite{zhao2024ultraedit} and FLUX.2~\cite{flux-2-2025} employ MM-DiT-style architectures. UltraEdit is a 12B DiT model, whereas FLUX.2 is a 32B DiT model equipped with a 24B language model.

\hspace*{1em} \noindent\textbf{(2) Unified Multimodal Models:} OneCAT~\cite{li2025onecat}, 
Nextstep-V1~\cite{han2025nextstep}, EMU3.5~\cite{cui2025emu3}, Lumina-DiMOO~\cite{xin2025lumina}, UniWorld-V1~\cite{lin2025uniworld}, 
UniWorld-V2~\cite{li2025uniworld},
OmniGen2~\cite{wu2025omnigen2}, Step1X-Edit-v1p2~\cite{liu2025step1x}, Unipic3.0~\cite{wei2026skywork}, Qwen-Image-Edit-2511~\cite{wu2025qwen}, Longcat-Image-Edit~\cite{team2025longcat}, InternVL-U~\cite{tian2026internvl}, DeepGen 1.0~\cite{wang2026deepgen}, ChronoEdit~\cite{wu2025chronoedit},  JoyAI-Image-Edit~\cite{joyaiimage2026}, Bagel\cite{deng2025emerging}, and UniReason1.0~\cite{wang2026unireason}. These models integrate the visual and semantic understanding capabilities of vision-language models with the generative capacity of diffusion models. They can be broadly grouped into four representative architectural paradigms: 
\textbf{(a). Perception-to-Editing Architecture: } Some models perform editing by using the hidden states produced by vision-language models as conditioning inputs to diffusion models. Representative examples include UniWorld-V1~\cite{lin2025uniworld}, OmniGen2~\cite{wu2025omnigen2}, Step1X-Edit-v1p2~\cite{liu2025step1x}, Unipic3.0~\cite{wei2026skywork}, Qwen-Image-Edit-2511~\cite{wu2025qwen}, and Longcat-Image-Edit~\cite{team2025longcat}, among others. These works further explore the design of visual encoders: some adopt semantic encoders, some use VAEs, and others combine both to provide complementary visual representations for diffusion-based editing. \textbf{(b). Autoregressive Architecture:} This paradigm performs visual generation with autoregressive models. Representative works include OneCAT~\cite{li2025onecat}, NextStep-V1~\cite{han2025nextstep}, and EMU3.5~\cite{cui2025emu3}. These methods typically represent images as sequences and progressively predict visual content through autoregressive modeling. Unlike earlier approaches that rely on discrete visual tokenizers, NextStep-V1 moves beyond discrete token representations by introducing a flow-matching head to predict continuous visual representations, enabling image generation in a continuous visual space.
\textbf{(c). Omni Discrete Diffusion Architecture:} Combining diffusion large language models and diffusion models to achieve unified understanding and generation, including Lumina-DiMOO~\cite{xin2025lumina}.
\textbf{(d). Hybrid Architecture:} Other models, such as BAGEL~\cite{deng2025emerging} and UniReason1.0~\cite{wang2026unireason}, introduce a mixture-of-transformer architecture within a single integrated model. This design enables the model to jointly handle visual understanding and visual generation.

\hspace*{1em} \noindent\textbf{(3) Closed-source Models:} Nano Banana Pro~\cite{nanobananapro}, Nano Banana 2~\cite{google2026gemini31flashimagepreview},  Seedream4.5~\cite{seedream2025seedream}, and Wan2.7-Image~\cite{wan}. Since the model weights are not publicly available, we evaluate these models through their official API services. 

\textbf{Evaluation Metrics.}
\label{supp: Evaluation pipeline}
The evaluation of {\bench} is based on a dynamic and comprehensive set of five metrics: Instruction Following (IF), World Knowledge Awareness (WA), Unedited Region Consistency (URC), Identity Consistency (IC), and Visual Quality (VQ). These metrics are further organized into higher-level evaluation dimensions. Specifically, Instruction Awareness is composed of Instruction Following and World Knowledge Awareness. While Instruction Following is evaluated for all tasks, World Knowledge Awareness is only assessed for tasks that require world knowledge or deep reasoning. Formally, Instruction Awareness is computed as:
\[
\mathrm{IA} =
\frac{1}{|\mathcal{M}_{\mathrm{IA}}|}
\sum_{m \in \mathcal{M}_{\mathrm{IA}}} m,
\quad
\mathcal{M}_{\mathrm{IA}} \subseteq \{\mathrm{IF}, \mathrm{WA}\},
\]

Visual Consistency consists of Unedited Region Consistency (URC) and Identity Consistency (IC). URC is evaluated for all tasks except style transfer, as preserving the consistency of unedited regions is crucial for image editing. IC is evaluated for tasks that require identity preservation, such as Object Movement and Object Swap. Formally, Visual Consistency is computed as:
\[
\mathrm{VC} =
\frac{1}{|\mathcal{M}_{\mathrm{VC}}|}
\sum_{m \in \mathcal{M}_{\mathrm{VC}}} m,
\quad
\mathcal{M}_{\mathrm{VC}} \subseteq \{\mathrm{URC}, \mathrm{IC}\},
\]
Visual Quality is evaluated for all tasks, since preserving the perceptual quality of edited images is crucial for reliable image editing. For the assessment of all metrics, we employ Gemini-3.1-Pro~\cite{gemini3pro} as the automatic evaluator, which rates each result on a 1--5 scale using carefully designed, dimension-specific prompts.

Considering that different task categories emphasize different evaluation aspects, we adopt category-specific aggregation strategies to compute the overall score. Let
\[
\mathcal{M}=\{\mathrm{IA},\mathrm{VC},\mathrm{VQ}\},
\qquad
s_{\min}=\min_{m\in\mathcal{M}} m.
\]
We first apply a conservative failure check. If \(s_{\min}\leq 1\), the overall score is set to \(s_{\min}\). Otherwise, it is computed as a weighted geometric mean:
\[
\mathrm{Overall} =
\begin{cases}
s_{\min}, & s_{\min}\leq 1, \\[3pt]
\prod_{m\in\mathcal{M}} m^{w_m}, & s_{\min}>1.
\end{cases}
\]

Let \(\mathcal{T}_{\mathrm{base}}\) denote General, Dynamic Manipulation, Multi-Image, and Complex Tasks; \(\mathcal{T}_{\mathrm{wkr}}\) denote World Knowledge Reasoning tasks; and \(\mathcal{T}_{\mathrm{avr}}\) denote Algorithm Visual Reasoning. The weights are defined as:
\[
(w_{\mathrm{IA}},w_{\mathrm{VC}},w_{\mathrm{VQ}})=
\begin{cases}
(0.4,\,0.4,\,0.2), & t\in\mathcal{T}_{\mathrm{base}},\\
(0.5,\,0.3,\,0.2), & t\in\mathcal{T}_{\mathrm{wkr}},\\
(0.6,\,0.2,\,0.2), & t\in\mathcal{T}_{\mathrm{avr}}.
\end{cases}
\]

\begin{figure}[t]
  \centering
  
  \includegraphics[width=0.95\textwidth]{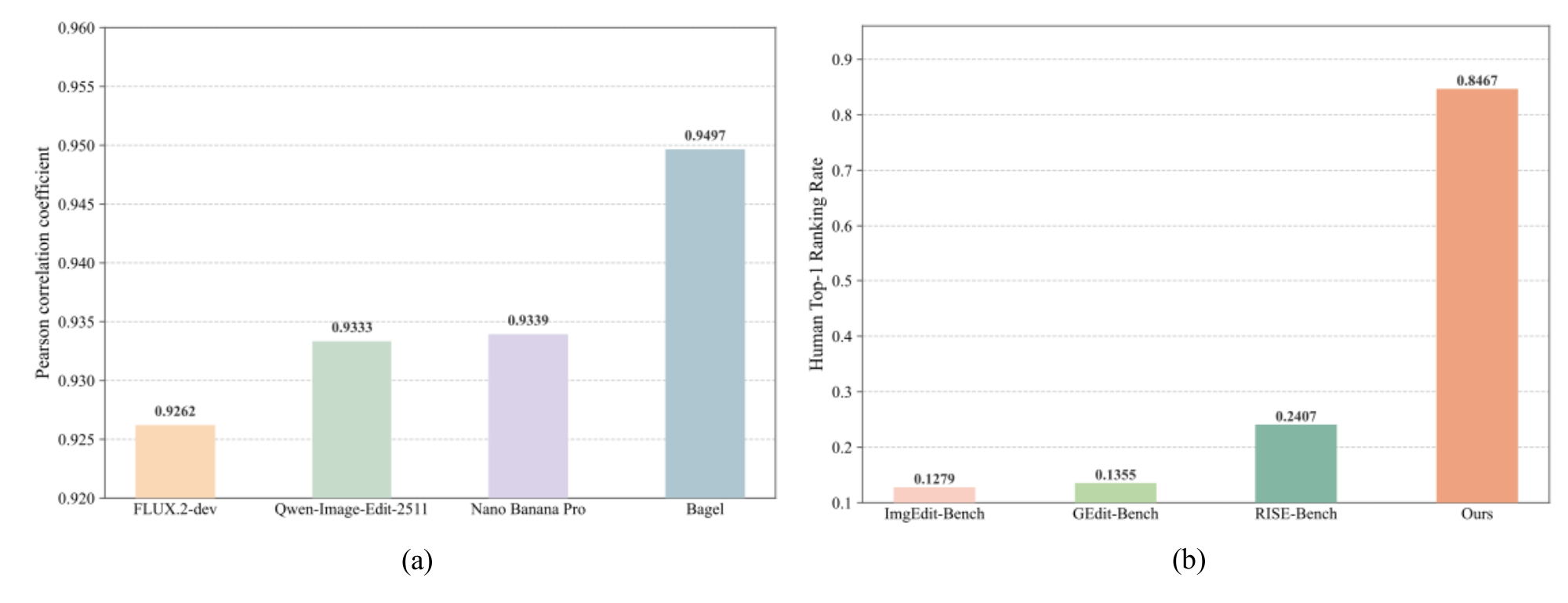}
  \vspace{-0.5em}
  \caption{
  (a) Pearson correlation between human ratings and MLLM scores. (b) Human Top-1 ranking rate of different evaluation protocols.
  }
  \label{User_Study}
  \vspace{-1em}
\end{figure}

Given the exceptional task coverage of {\bench}, we categorize the tasks into different groups and conduct small-scale pilot studies to ensure that the prompts accurately assess each task type. Specifically, we consider four groups: Complex Instruction, Complex Paint, Multi-Image, and Other Tasks. For Instruction Following, the corresponding prompt templates are provided in Templates~\ref{System Prompts: IF_Complex_Instruction}, \ref{System Prompts: IF_Complex Paint}, \ref{System Prompts: IF_Multi-Image}, and \ref{System Prompts: IF_other tasks}, respectively. For World Knowledge Awareness, we use a unified system prompt, as this metric is only evaluated on tasks involving world knowledge or deep reasoning; the corresponding template is provided in Template~\ref{System Prompts: WA}. For Unedited Region Consistency, we use separate prompt templates for the Complex Paint task and other tasks, as provided in Templates~\ref{System Prompts: URC_Complex_paint} and \ref{System Prompts: URC_Other_Tasks}, respectively. Similarly, for Identity Consistency, we use separate prompt templates for Multi-Image tasks and other tasks, as provided in Templates~\ref{System Prompts: IC_Multi-Image} and \ref{System Prompts: IC_Other_Tasks}, respectively. Finally, the prompt template for Visual Quality is provided in Template~\ref{System Prompts: Visual Quality}.

\begin{figure}[t]
  \centering
  \includegraphics[width=0.95\textwidth]{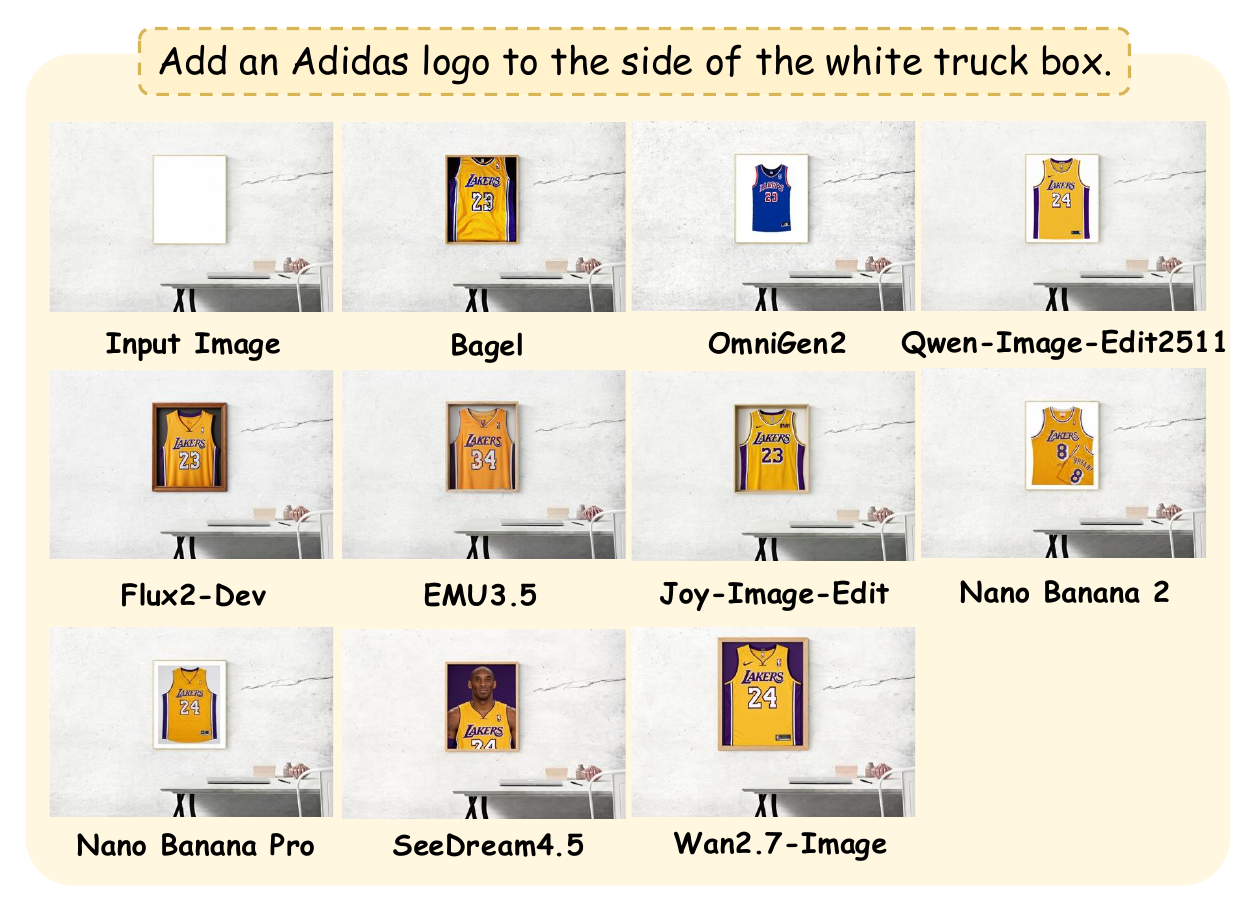}
  \vspace{-0.5em}
  \caption{Qualitative comparisons on the Subject Addition task.}
  \label{Fig:ADD}
  \vspace{-1em}
\end{figure}

\begin{figure}[t]
  \centering
  
  \includegraphics[width=0.95\textwidth]{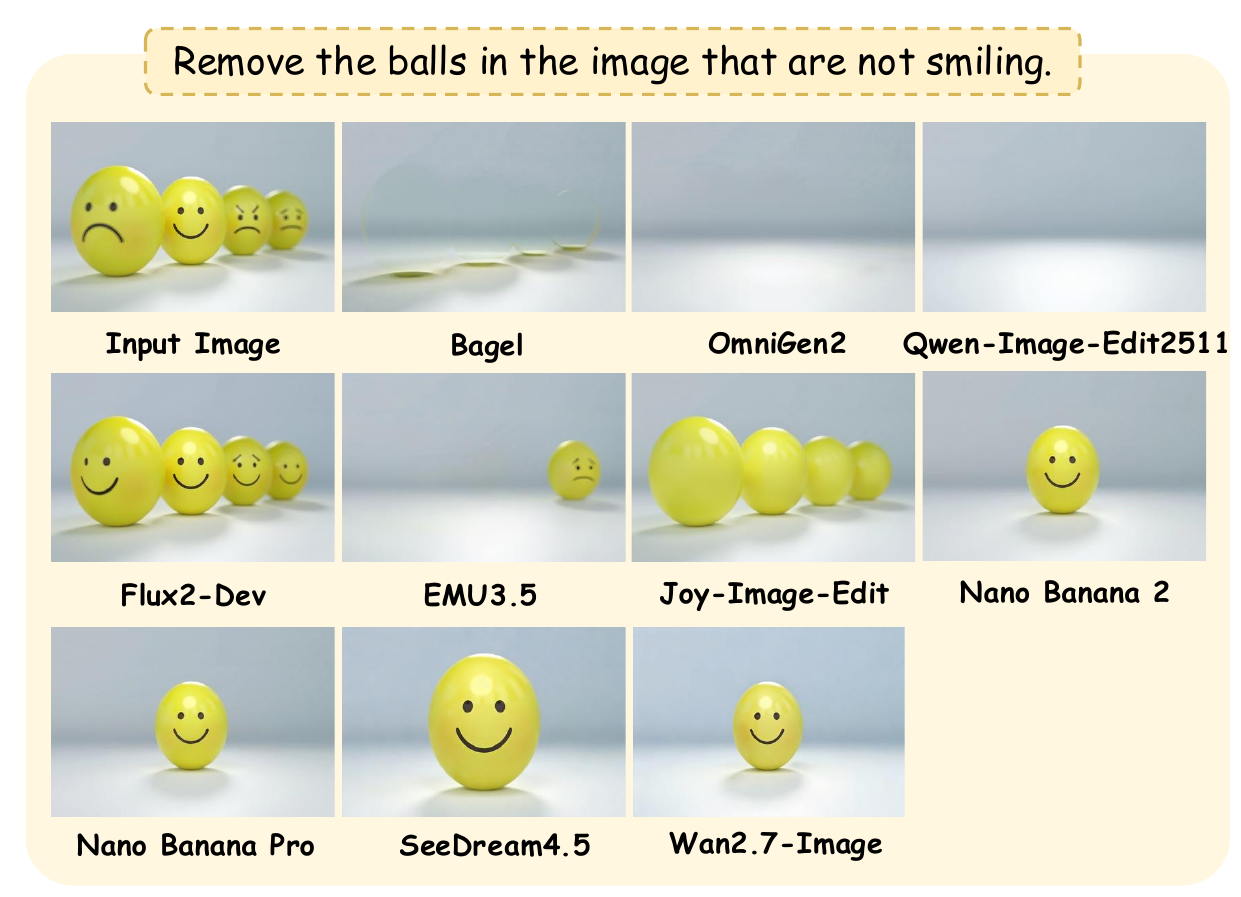}
  \vspace{-0.5em}
  \caption{Qualitative comparisons on the Subject Remove task.}
  \label{Fig: Remove}
  \vspace{-1em}
\end{figure}

\begin{figure}[t]
  \centering
  \includegraphics[width=0.95\textwidth]{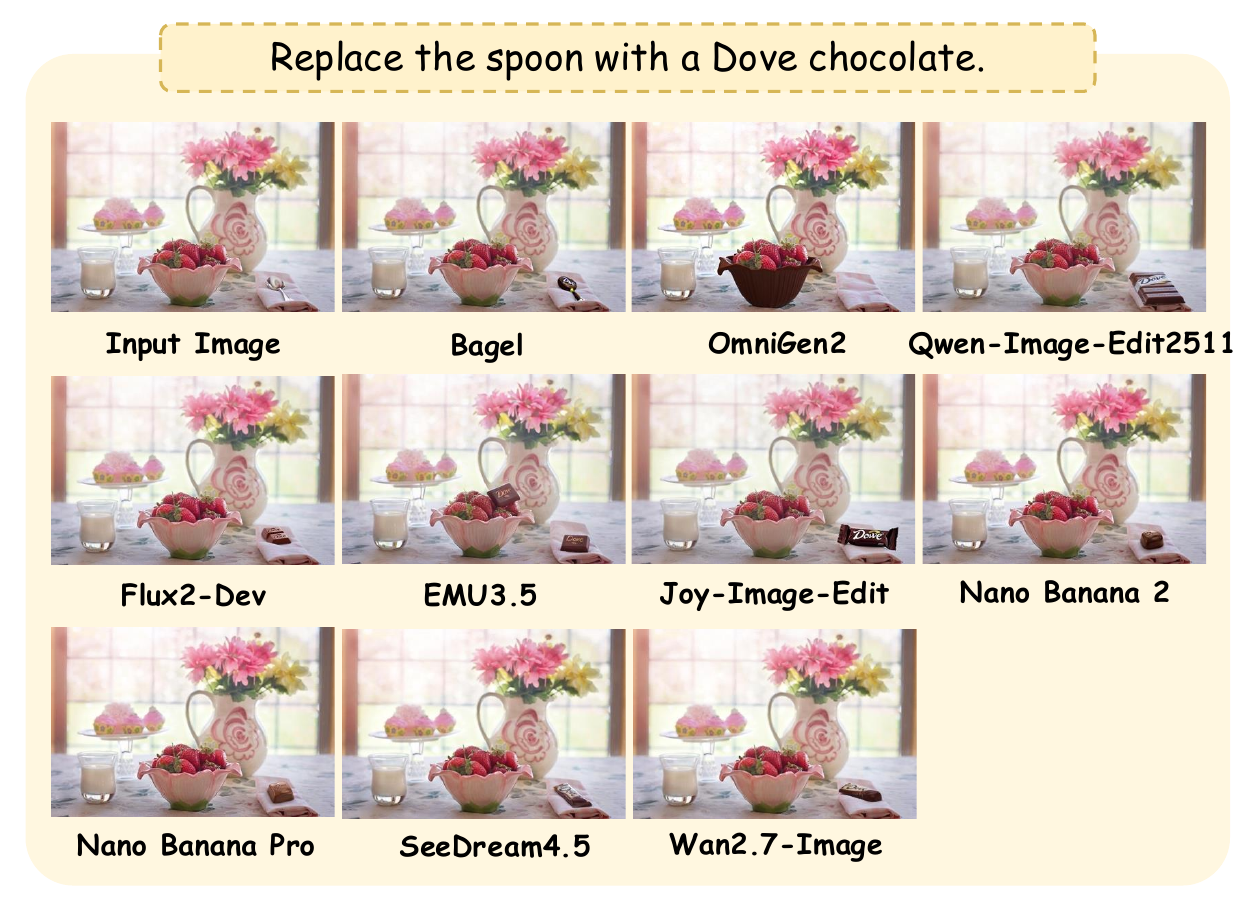}
  \vspace{-0.5em}
  \caption{Qualitative comparisons on the Subject Replace task.}
   \label{Fig: Replace}
  \vspace{-1em}
\end{figure}

\begin{figure}[t]
  \centering
  
  \includegraphics[width=0.95\textwidth]{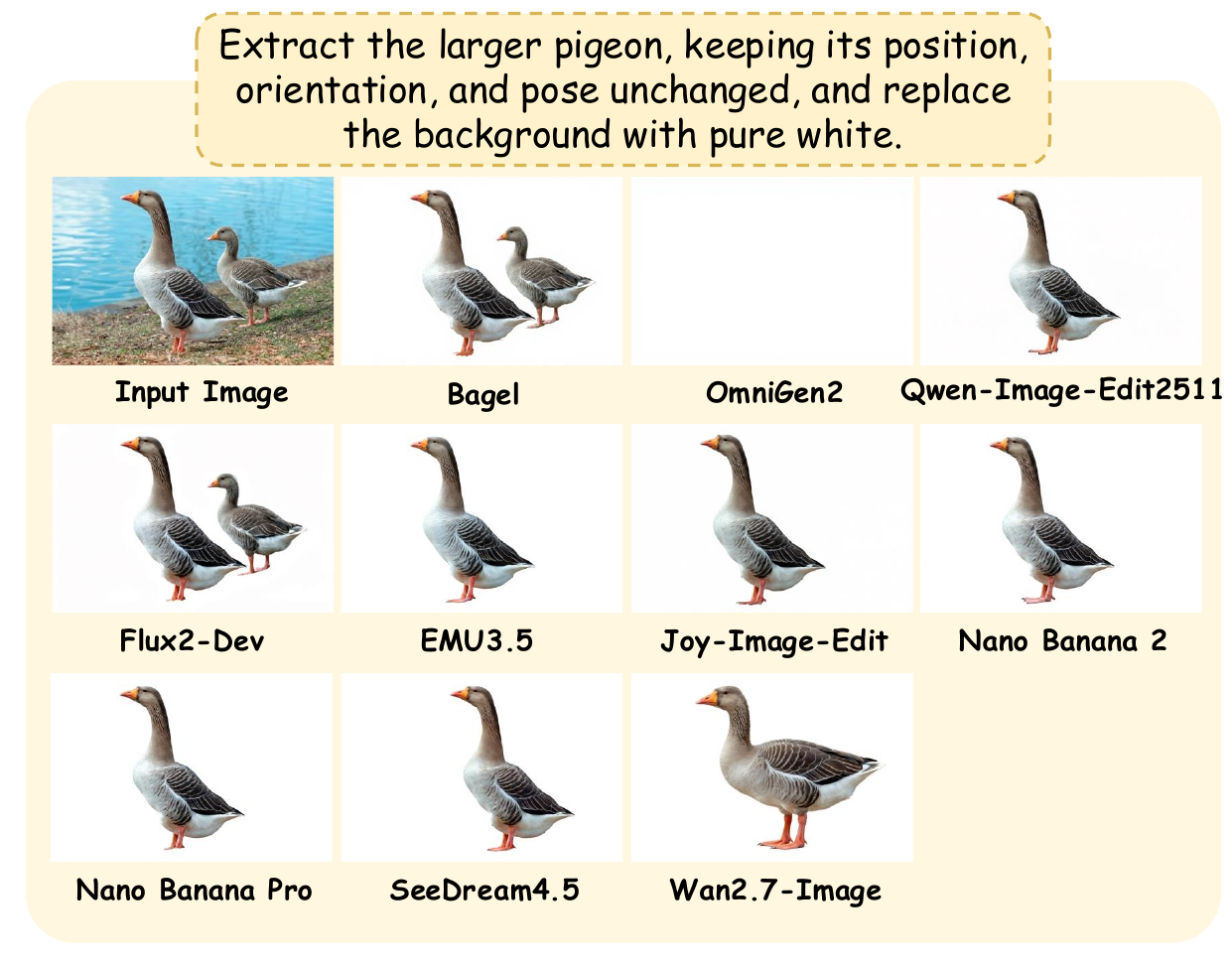}
  \vspace{-0.5em}
  \caption{Qualitative comparisons on the Subject Extract task.}
  \label{Fig: Extract}
  \vspace{-1em}
\end{figure}

\begin{figure}[t]
  \centering
  
  \includegraphics[width=0.95\textwidth]{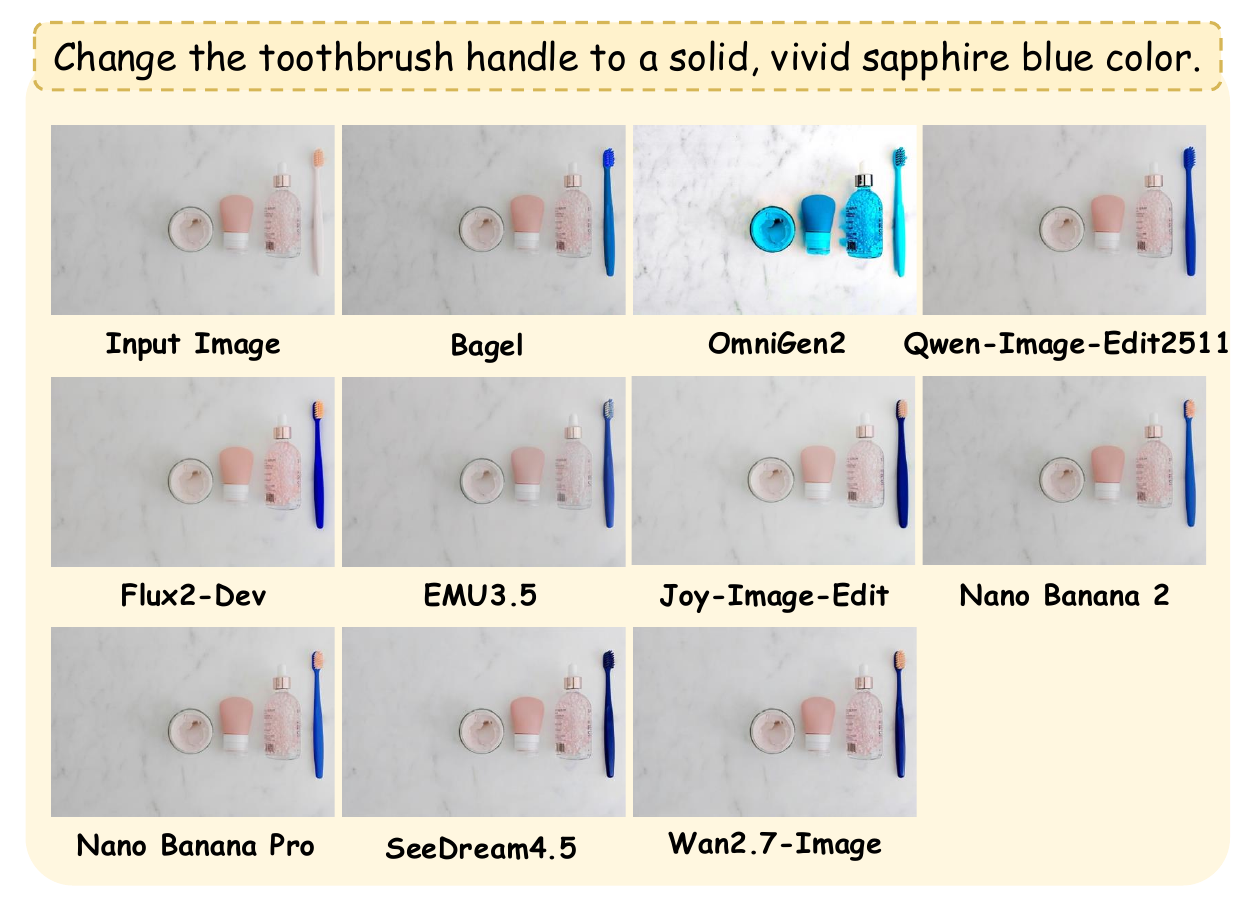}
  \vspace{-0.5em}
  \caption{Qualitative comparisons on the  Change Color task.}
  \label{Fig: Change_color}
  \vspace{-1em}
\end{figure}

\begin{figure}[t]
  \centering
  
  \includegraphics[width=0.95\textwidth]{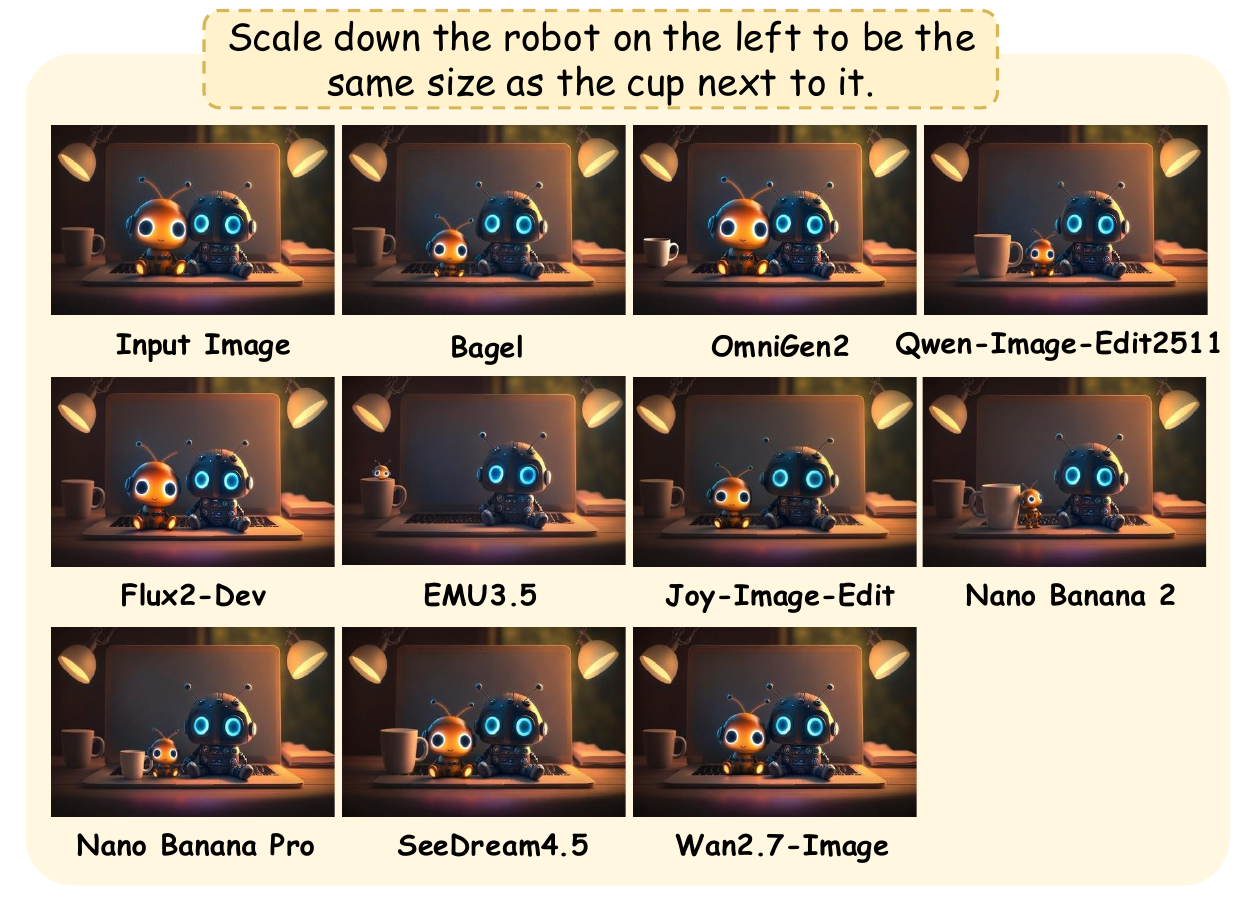}
  \vspace{-0.5em}
  \caption{Qualitative comparisons on the  Change Size task.}
  \label{Fig: Change_Size}
  \vspace{-1em}
\end{figure}

\begin{figure}[t]
  \centering
 
  \includegraphics[width=0.95\textwidth]{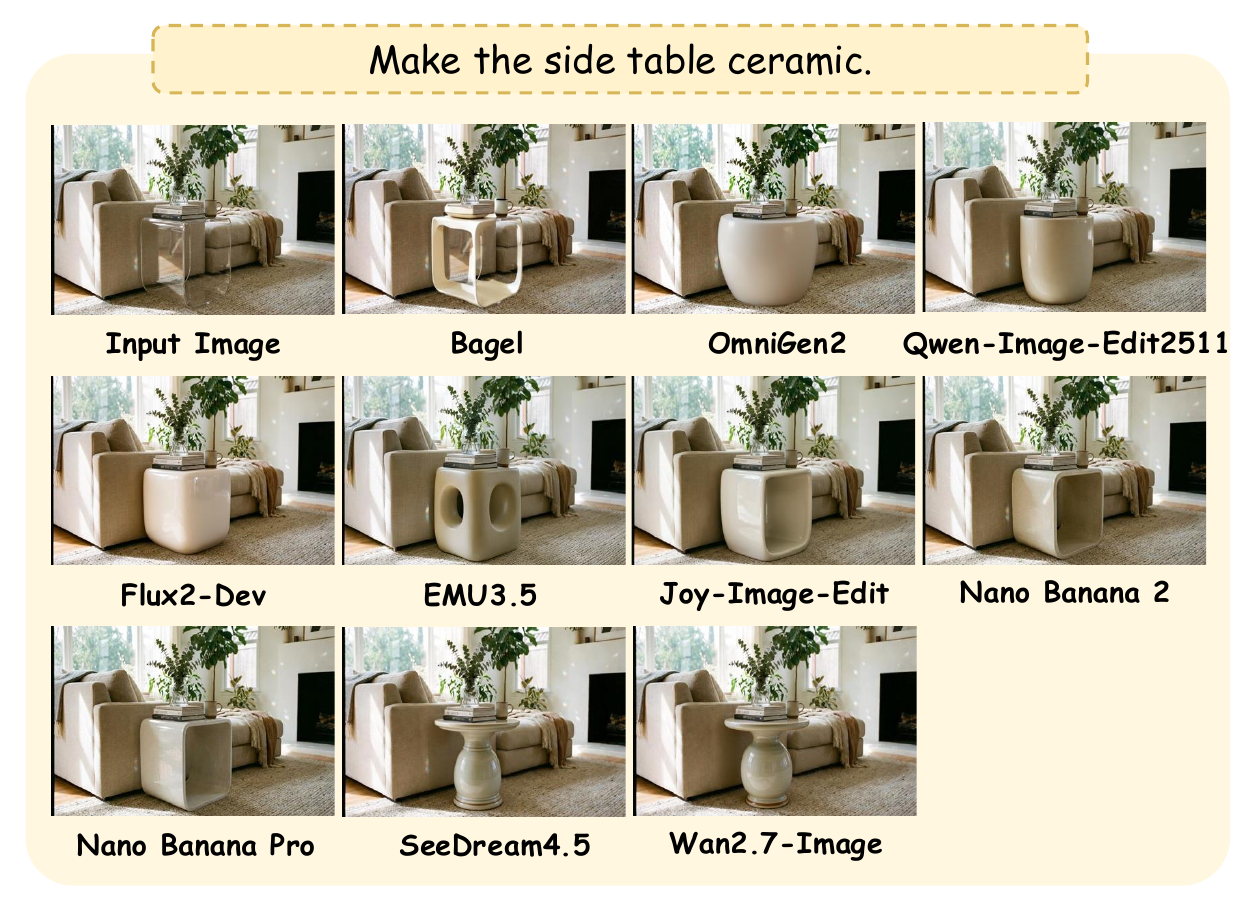}
  \vspace{-0.5em}
  \caption{Qualitative comparisons on the Change Material task.}
   \label{Fig: Change_Material}
  \vspace{-1em}
\end{figure}

\begin{figure}[t]
  \centering
  
  \includegraphics[width=0.95\textwidth]{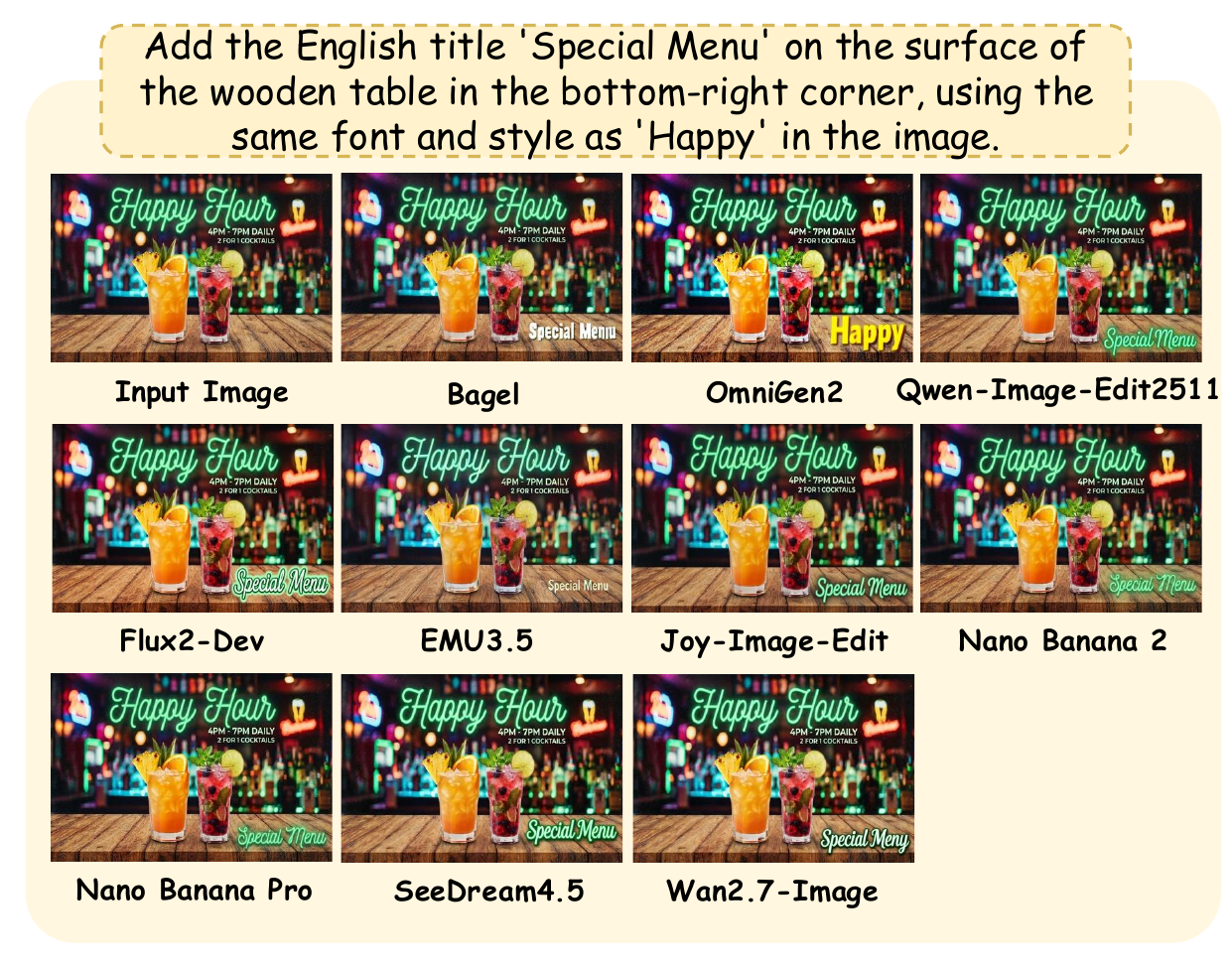}
  \vspace{-0.5em}
  \caption{Qualitative comparisons on the Visual Text Editing (EN) task.}
  \label{Fig: Visual_Text_Editing_en}
  \vspace{-1em}
\end{figure}

\begin{figure}[t]
  \centering
  
  \includegraphics[width=0.95\textwidth]{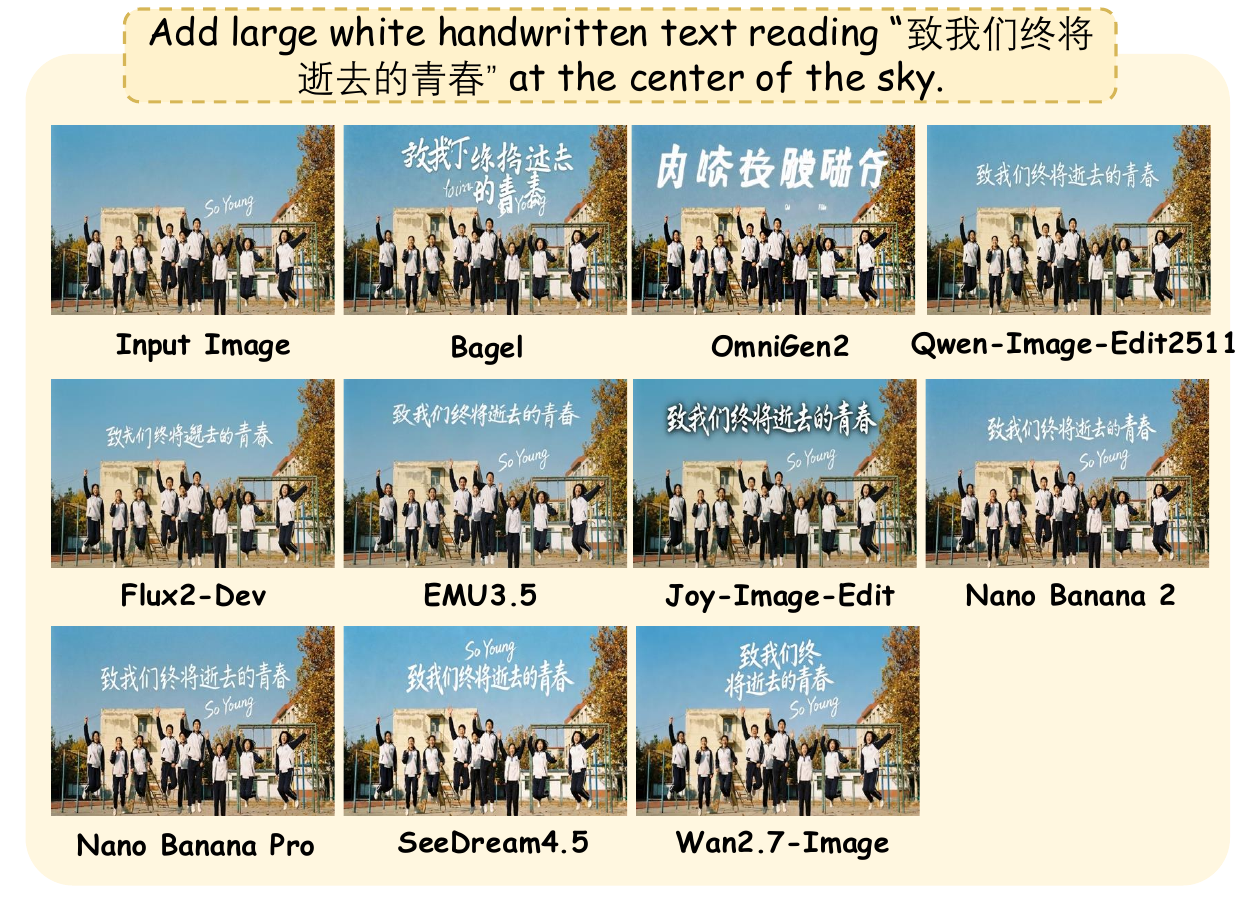}
  \vspace{-0.5em}
  \caption{Qualitative comparisons on the Visual Text Editing (CN) task.}
  \label{Fig: Visual_Text_Editing_cn}
  \vspace{-1em}
\end{figure}

\begin{figure}[t]
  \centering
  
  \includegraphics[width=0.95\textwidth]{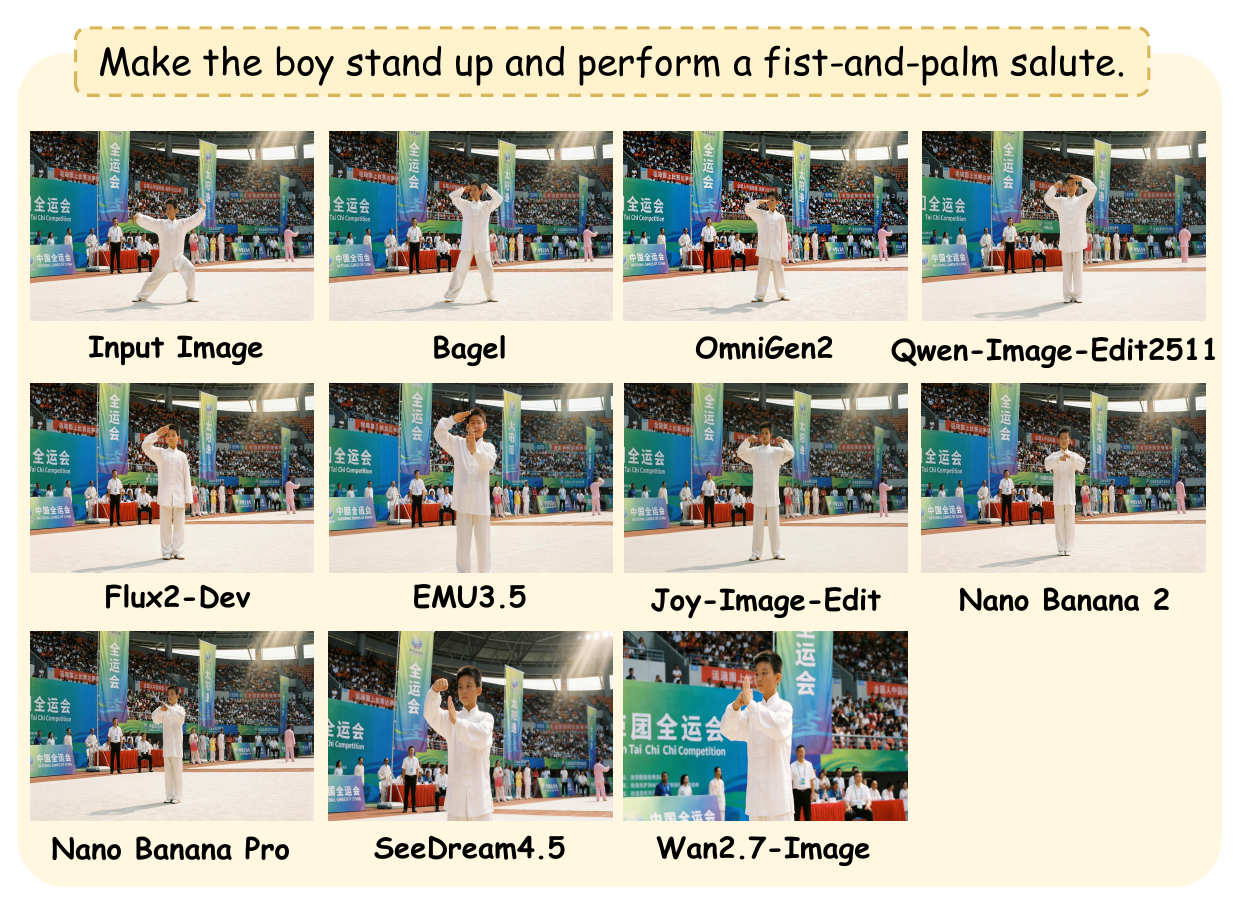}
  \vspace{-0.5em}
  \caption{Qualitative comparisons on the Action task.}
  \label{Fig: Action}
  \vspace{-1em}
\end{figure}

\begin{figure}[t]
  \centering
  
  \includegraphics[width=0.95\textwidth]{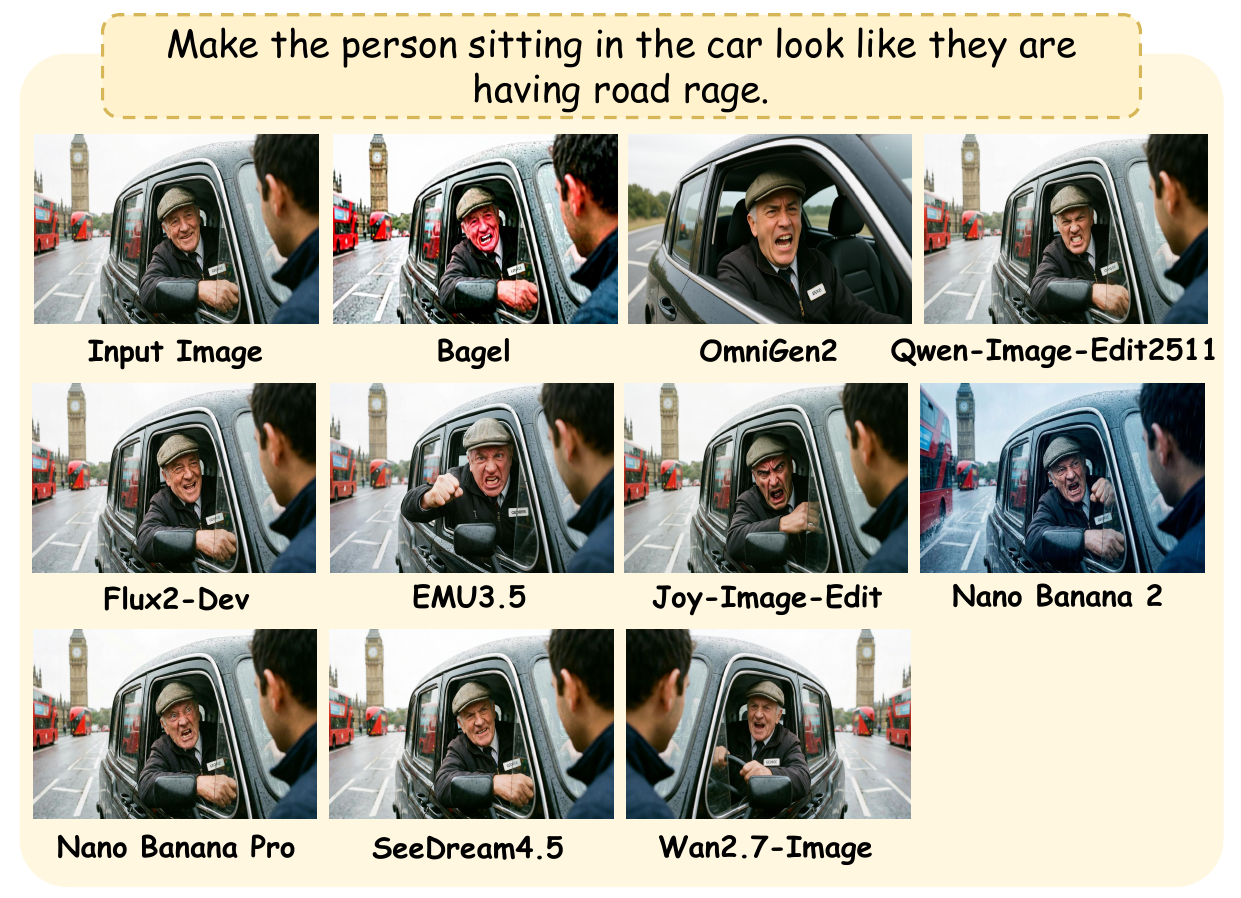}
  \vspace{-0.5em}
  \caption{Qualitative comparisons on the Change Emotion task.}
  \label{Fig: Emotion}
  \vspace{-1em}
\end{figure}

\begin{figure}[t]
  \centering
  
  \includegraphics[width=0.95\textwidth]{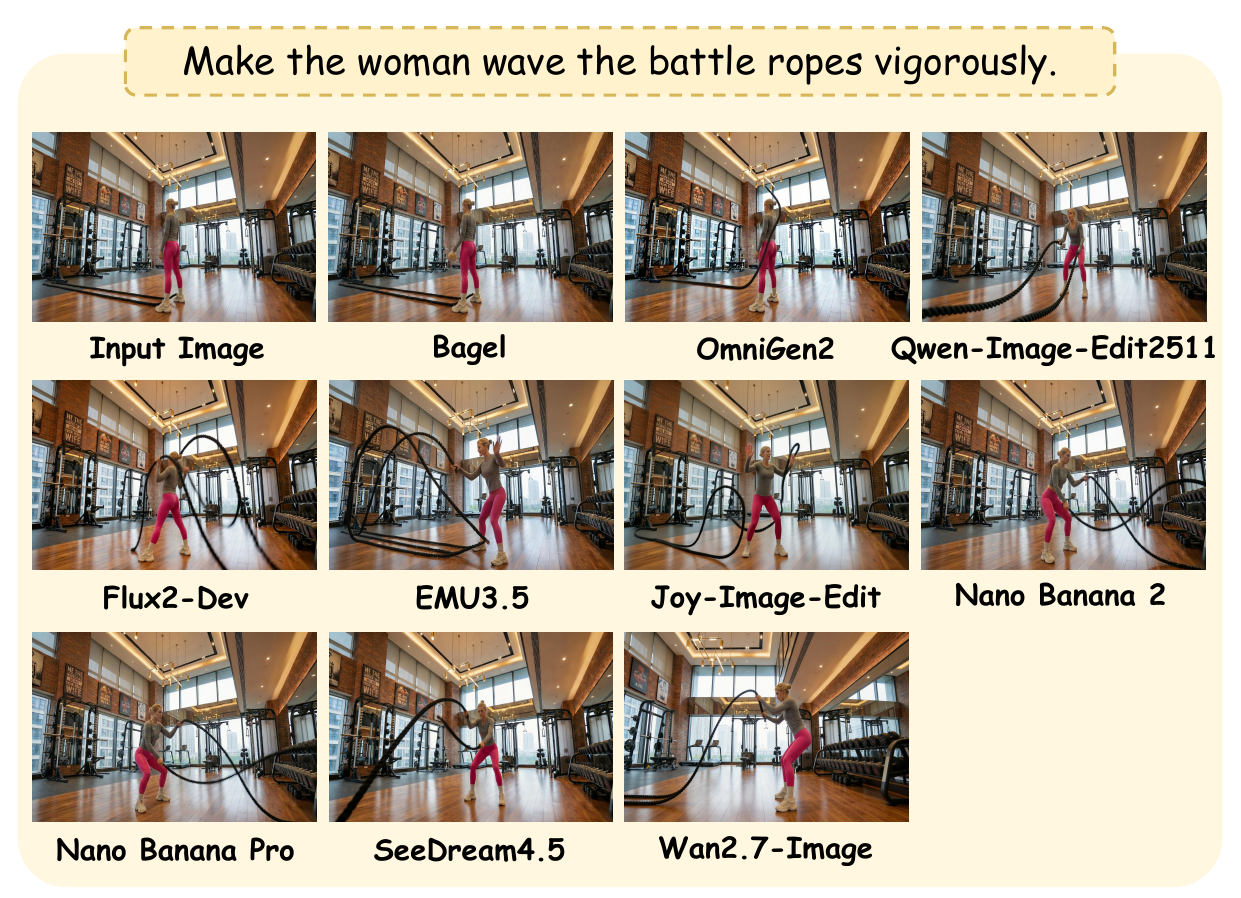}
  \vspace{-0.5em}
  \caption{Qualitative comparisons on the Object Interaction task.}
  \label{Fig: Object_Interaction}
  \vspace{-1em}
\end{figure}

\begin{figure}[t]
  \centering
  
  \includegraphics[width=0.95\textwidth]{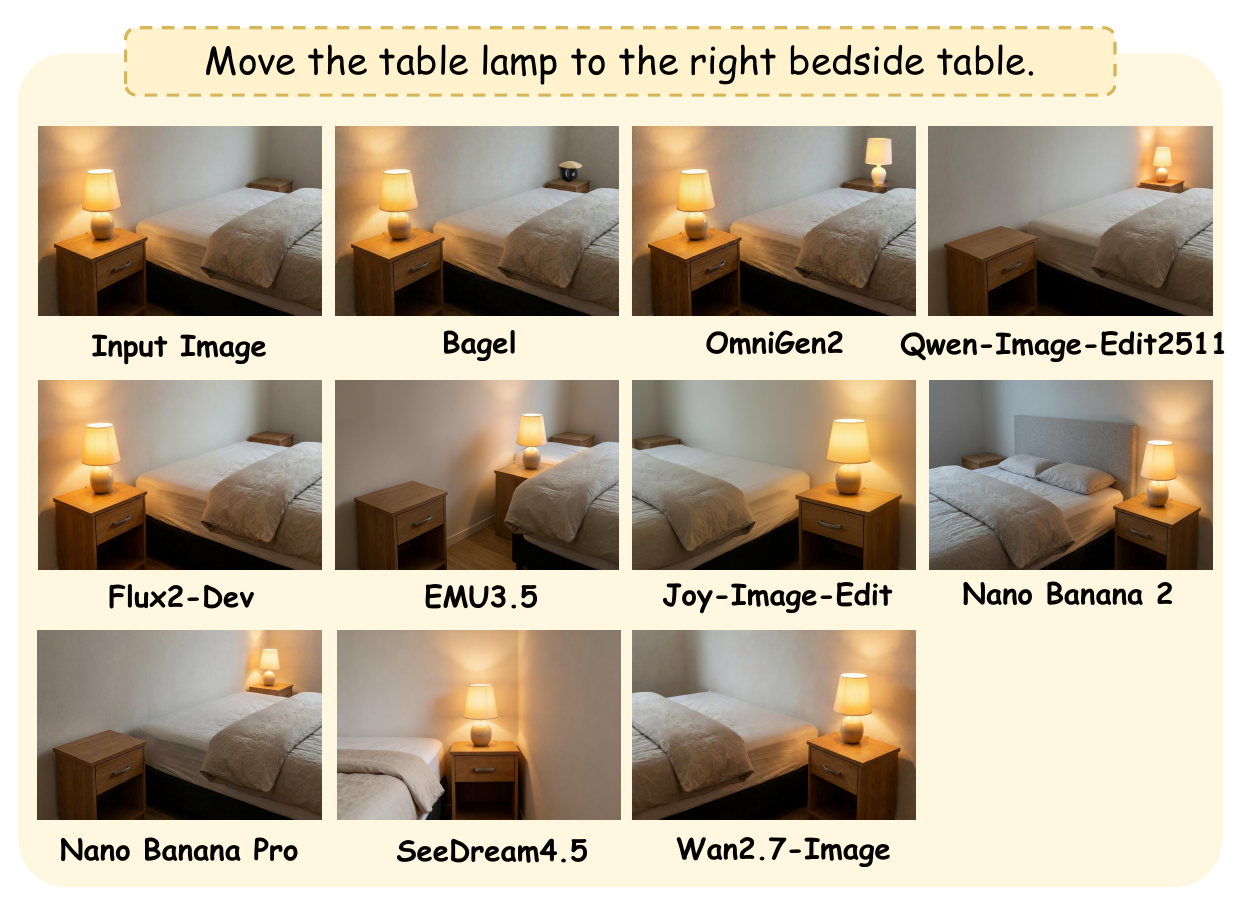}
  \vspace{-0.5em}
  \caption{Qualitative comparisons on the Object Movement task.}
  \label{Fig: Movement}
  \vspace{-1em}
\end{figure}

\begin{figure}[t]
  \centering
  
  \includegraphics[width=0.95\textwidth]{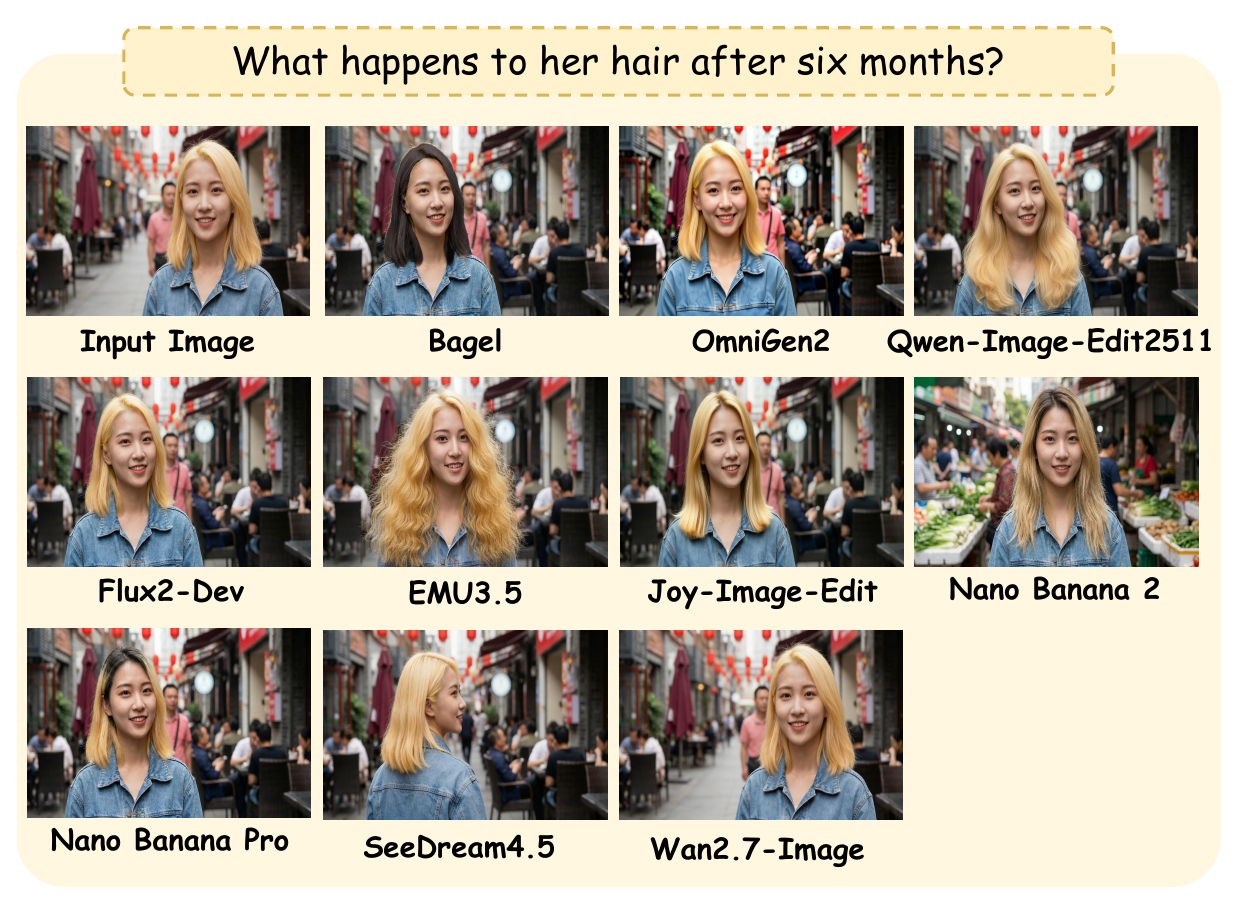}
  \vspace{-0.5em}
  \caption{Qualitative comparisons on the Temporal Reasoning task.}
  \label{Fig: Temporal}
  \vspace{-1em}
\end{figure}

\begin{figure}[t]
  \centering
  
  \includegraphics[width=0.95\textwidth]{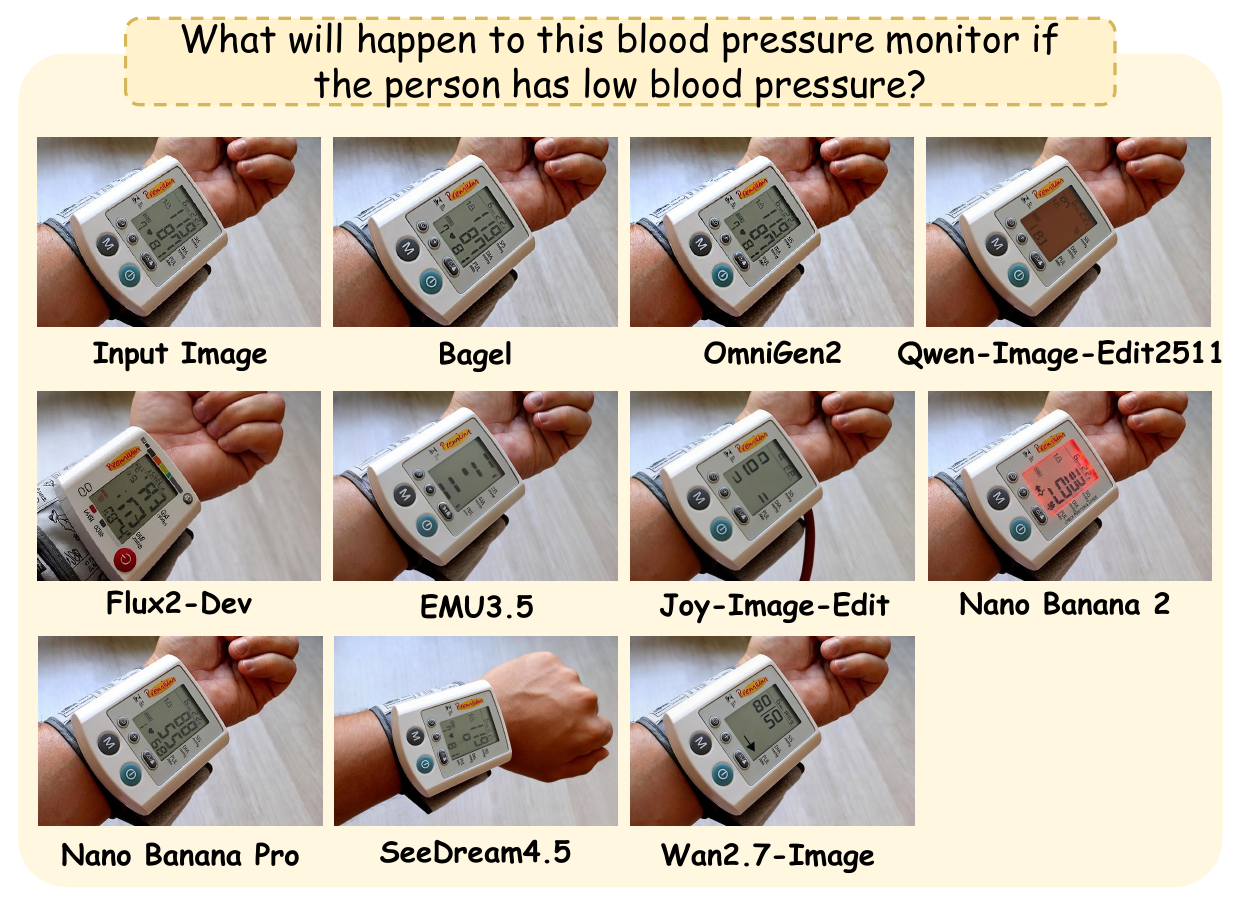}
  \vspace{-0.5em}
  \caption{Qualitative comparisons on the Casual Reasoning task.}
  \label{Fig: Casual}
  \vspace{-1em}
\end{figure}

\begin{figure}[t]
  \centering
  
  \includegraphics[width=0.95\textwidth]{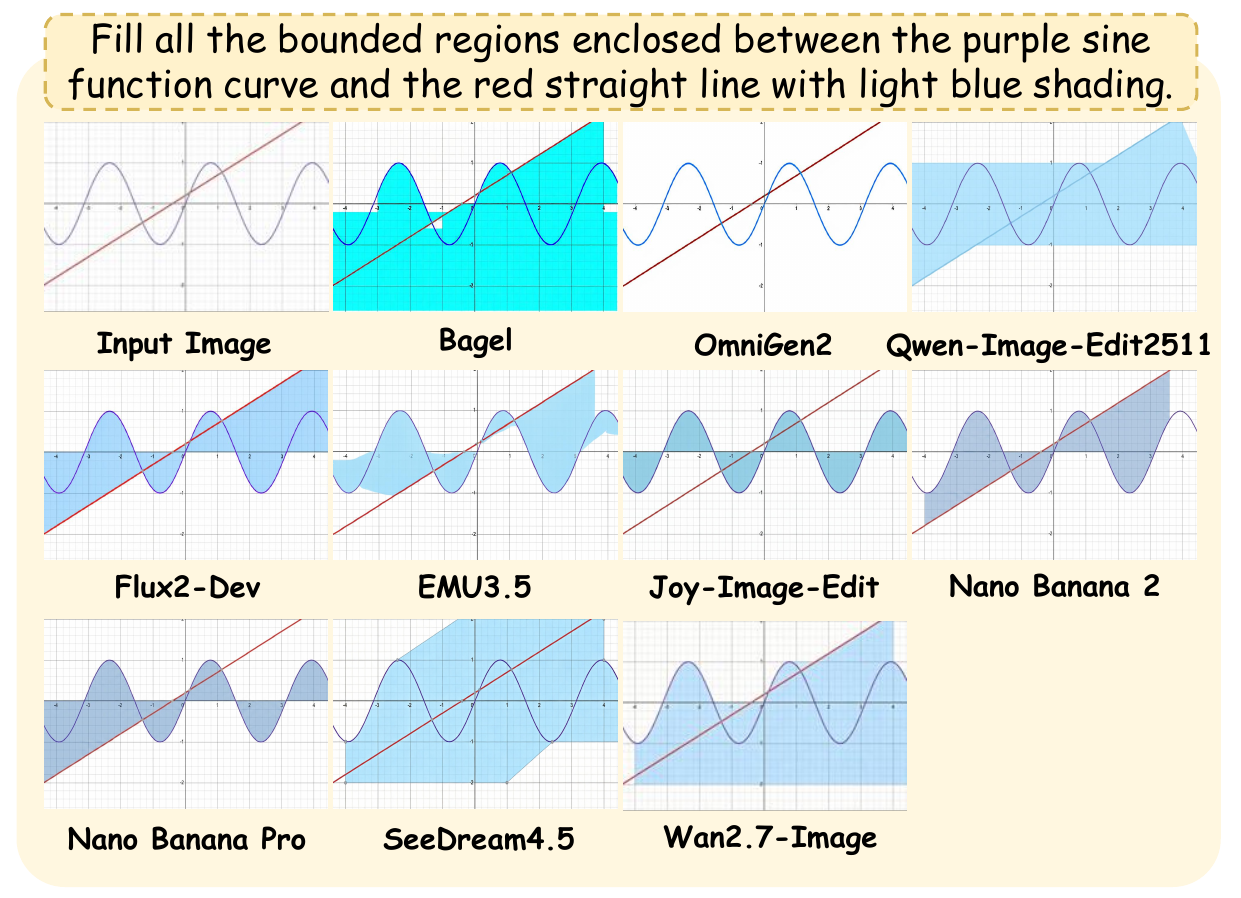}
  \vspace{-0.5em}
  \caption{Qualitative comparisons on the Math Reasoning task.}
  \label{Fig: Math}
  \vspace{-1em}
\end{figure}

\begin{figure}[t]
  \centering
  
  \includegraphics[width=0.95\textwidth]{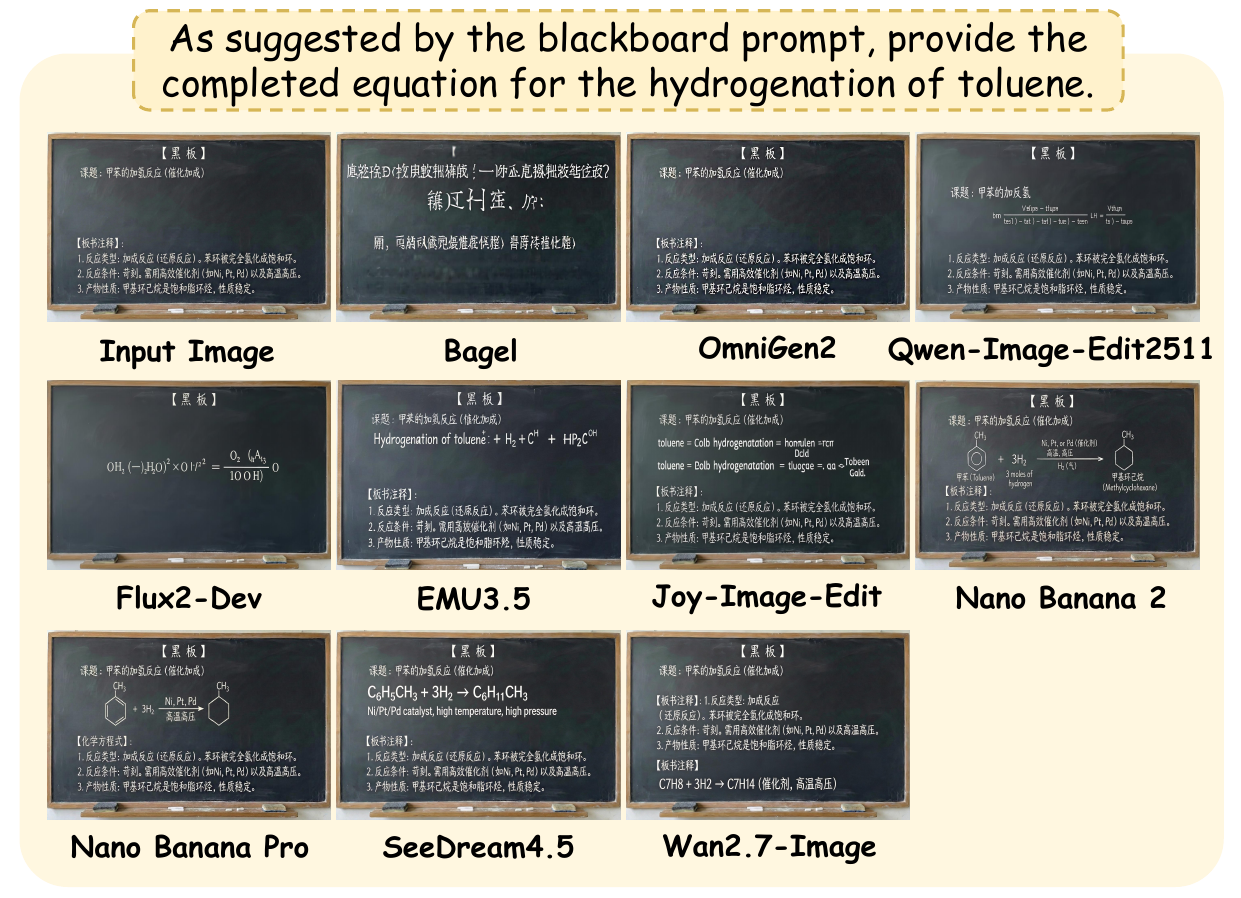}
  \vspace{-0.5em}
  \caption{Qualitative comparisons on the Chemical Reasoning task.}
  \label{Fig: Chemical}
  \vspace{-1em}
\end{figure}

\begin{figure}[t]
  \centering
 
  \includegraphics[width=0.95\textwidth]{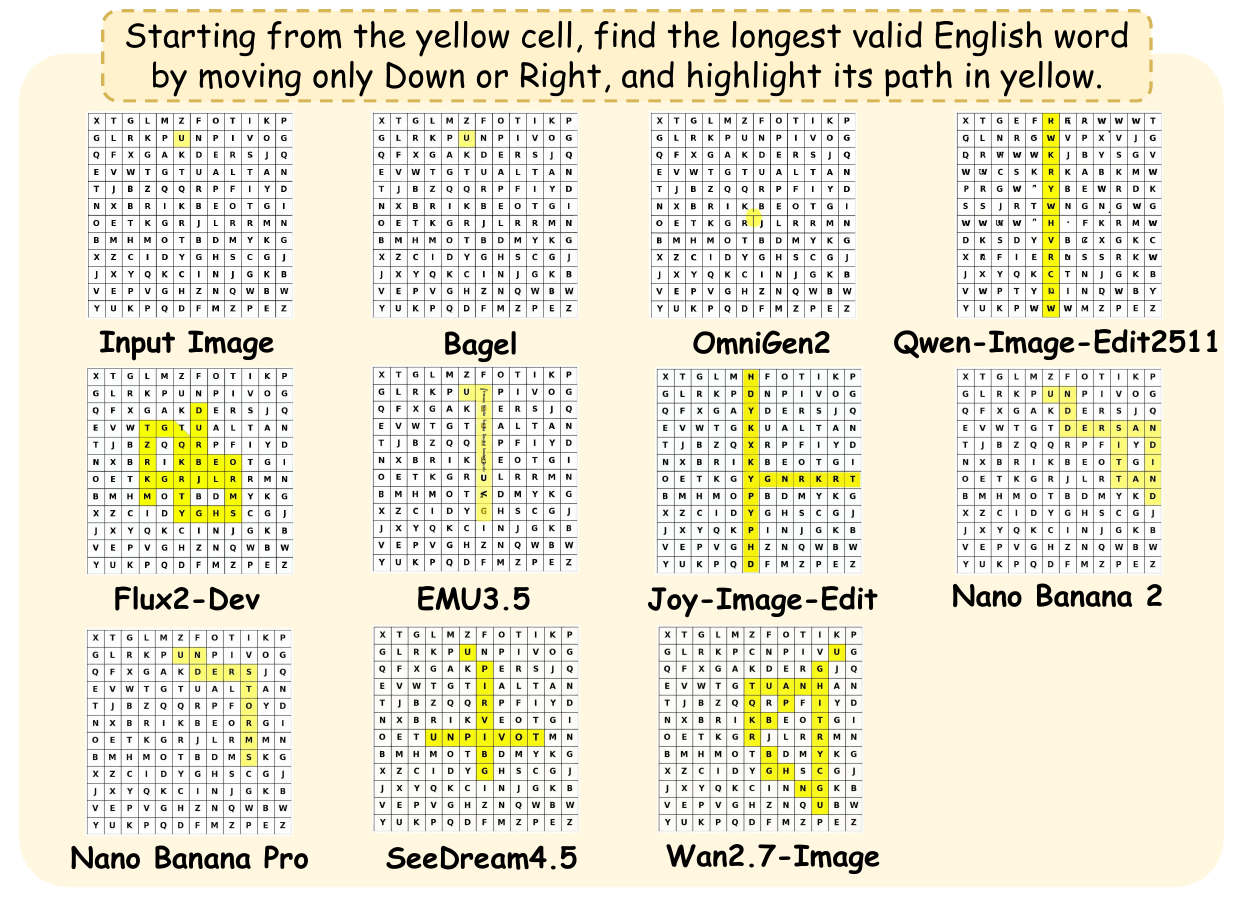}
  \vspace{-0.5em}
  \caption{Qualitative comparisons on the Global Longest Word Discovery task.}
   \label{Fig: Longest_word_with_start}
  \vspace{-1em}
\end{figure}

\begin{figure}[t]
  \centering
  
  \includegraphics[width=0.95\textwidth]{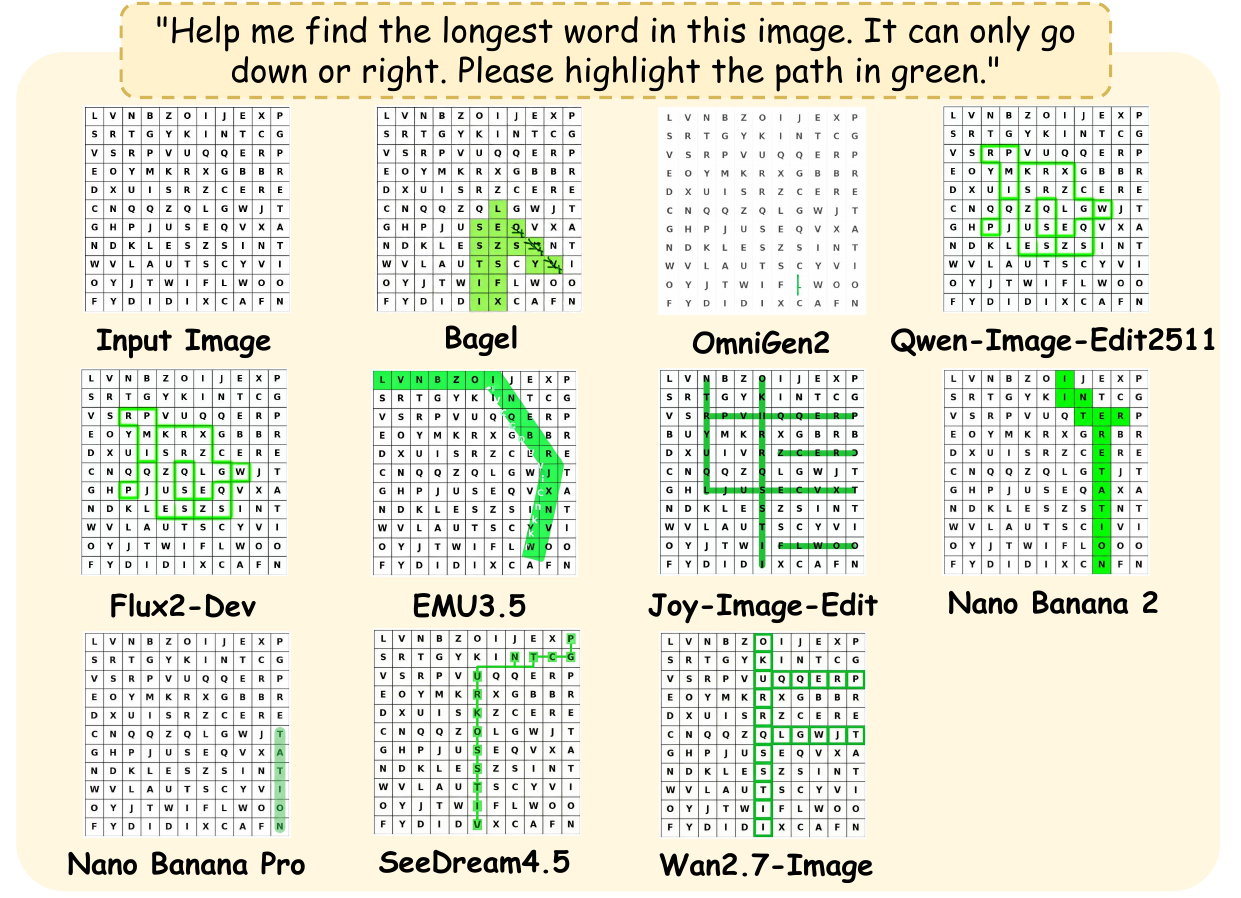}
  \vspace{-0.5em}
  \caption{Qualitative comparisons on the Longest Word Discovery task.}
  \label{Fig: Longest_word_no_start}
  \vspace{-1em}
\end{figure}

\begin{figure}[t]
  \centering
  
  \includegraphics[width=0.95\textwidth]{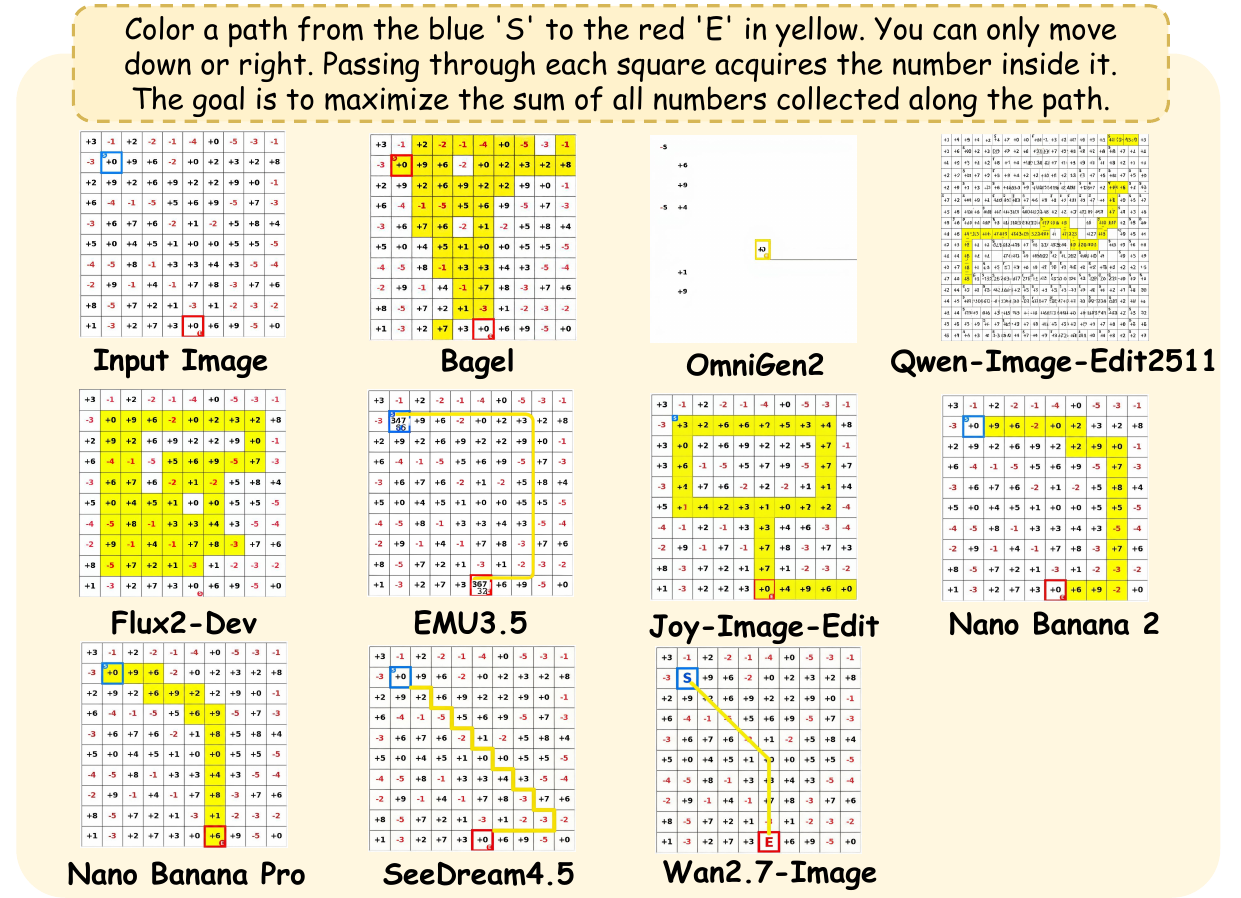}
  \vspace{-0.5em}
  \caption{Qualitative comparisons on the Maximum Bonus task.}
  \label{Fig: Max_Bonus}
  \vspace{-1em}
\end{figure}

\begin{figure}[t]
  \centering
  
  \includegraphics[width=0.95\textwidth]{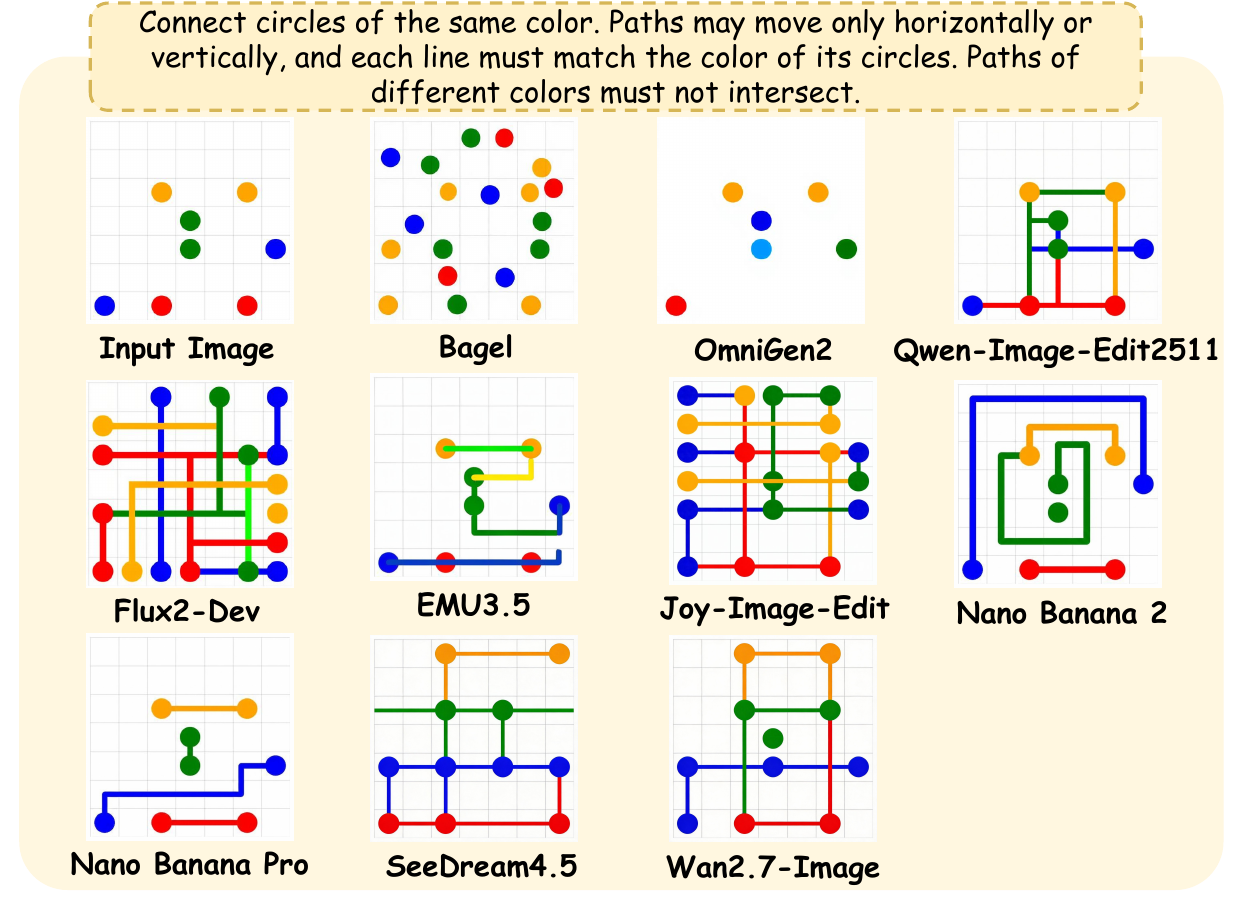}
  \vspace{-0.5em}
  \caption{Qualitative comparisons on the Number Link task.}
  \label{Fig: Number_Link}
  \vspace{-1em}
\end{figure}

\begin{figure}[t]
  \centering
  
  \includegraphics[width=0.95\textwidth]{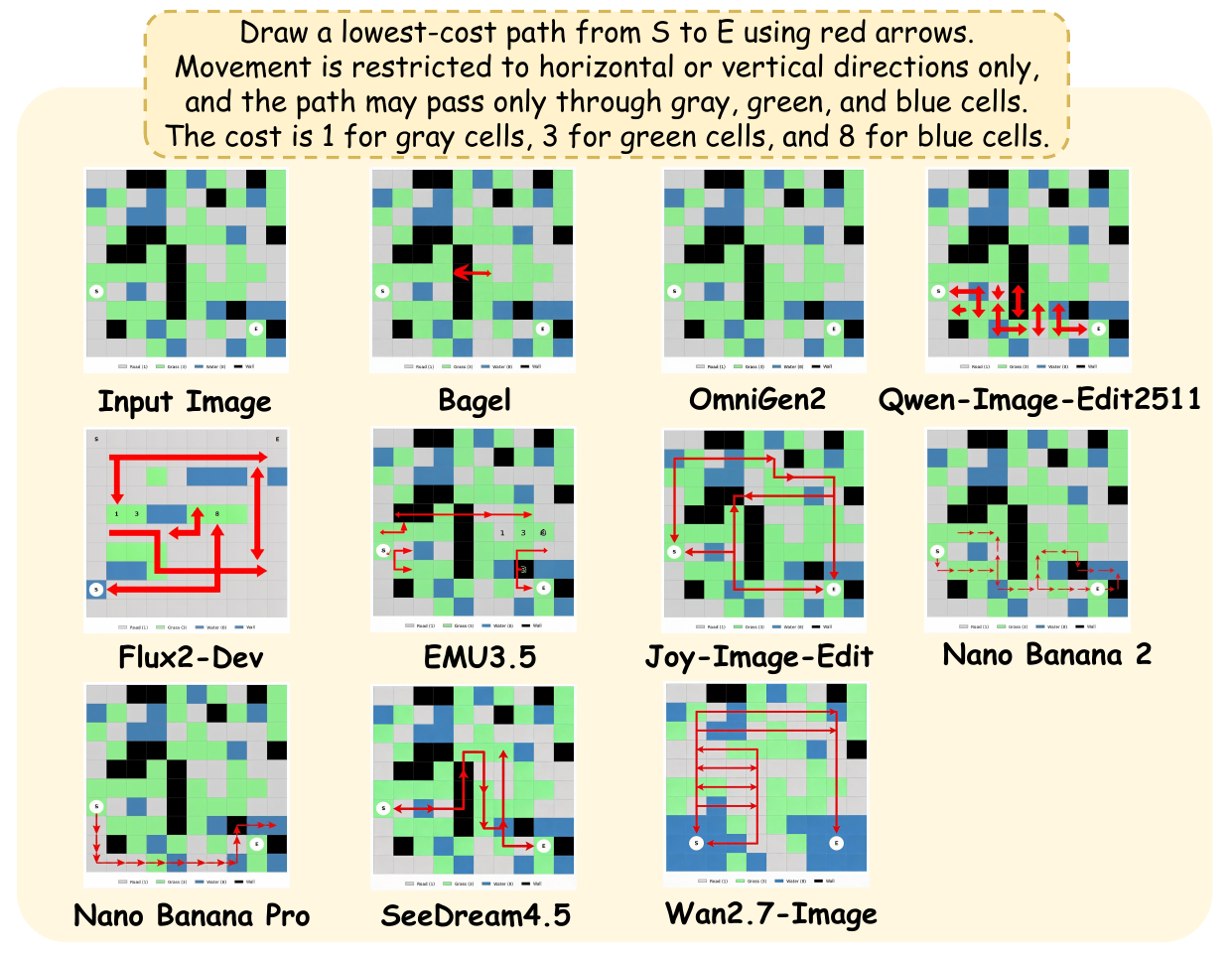}
  \vspace{-0.5em}
  \caption{Qualitative comparisons on the Optimal Path Identification task.}
  \label{Fig: Shortest_Path}
  \vspace{-1em}
\end{figure}

\begin{figure}[t]
  \centering
  
  \includegraphics[width=0.95\textwidth]{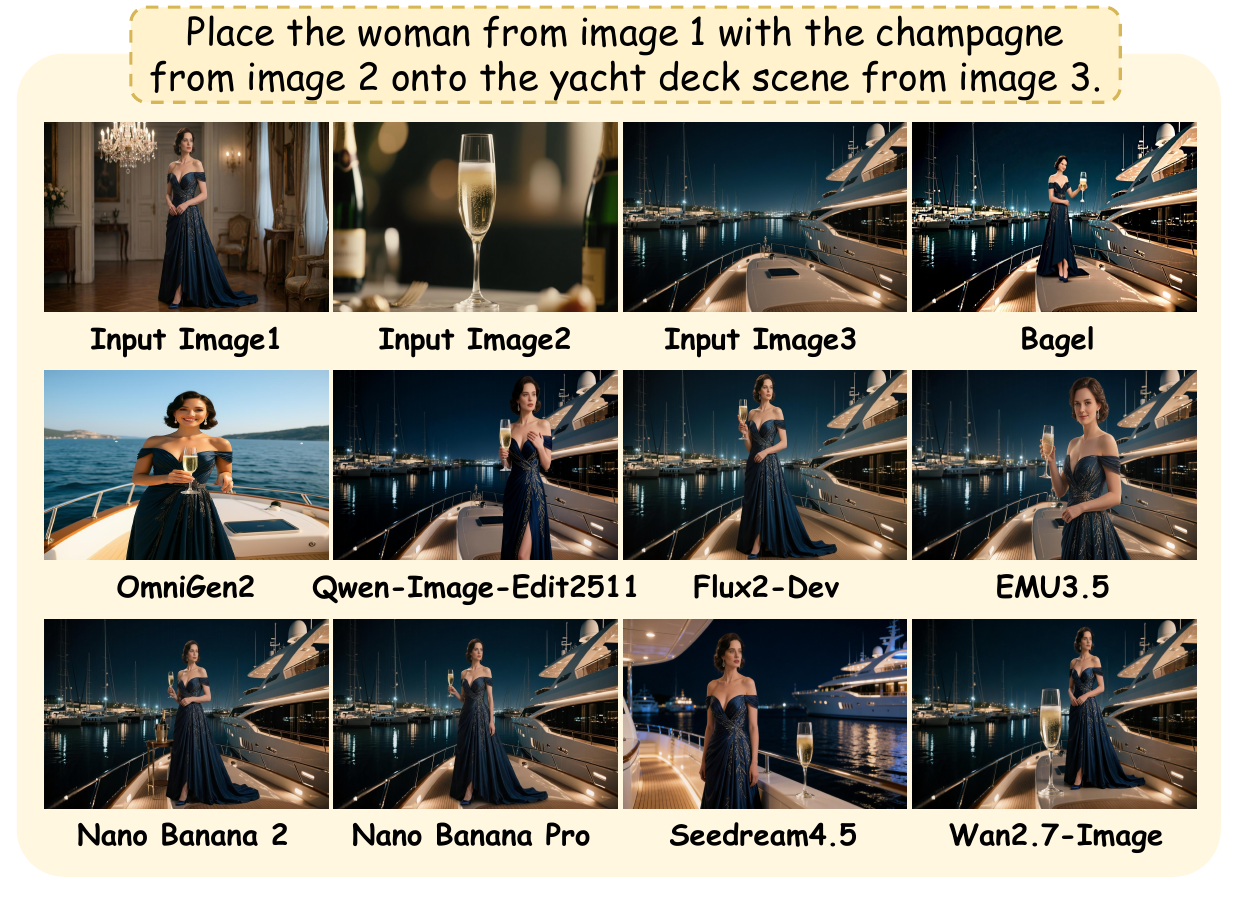}
  \vspace{-0.5em}
  \caption{Qualitative comparisons on the Multi-Image Composition task.}
  \label{Fig: Multi-Image Composition}
  \vspace{-1em}
\end{figure}

\begin{figure}[t]
  \centering
  
  \includegraphics[width=0.95\textwidth]{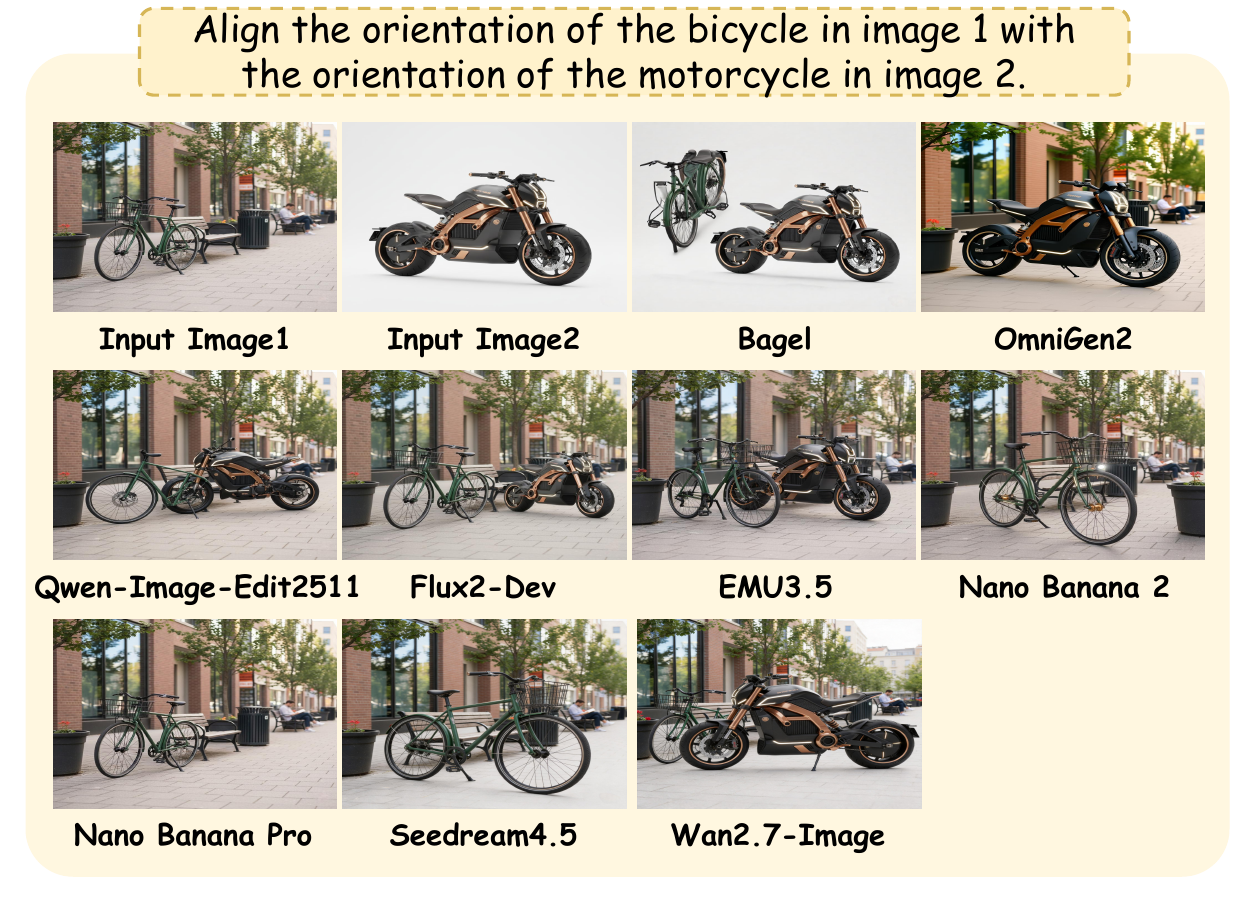}
  \vspace{-0.5em}
  \caption{Qualitative comparisons on the Multi-Image Awareness task.}
  \label{Fig: Multi-Image Awareness}
  \vspace{-1em}
\end{figure}

\begin{figure}[t]
  \centering
  
  \includegraphics[width=0.95\textwidth]{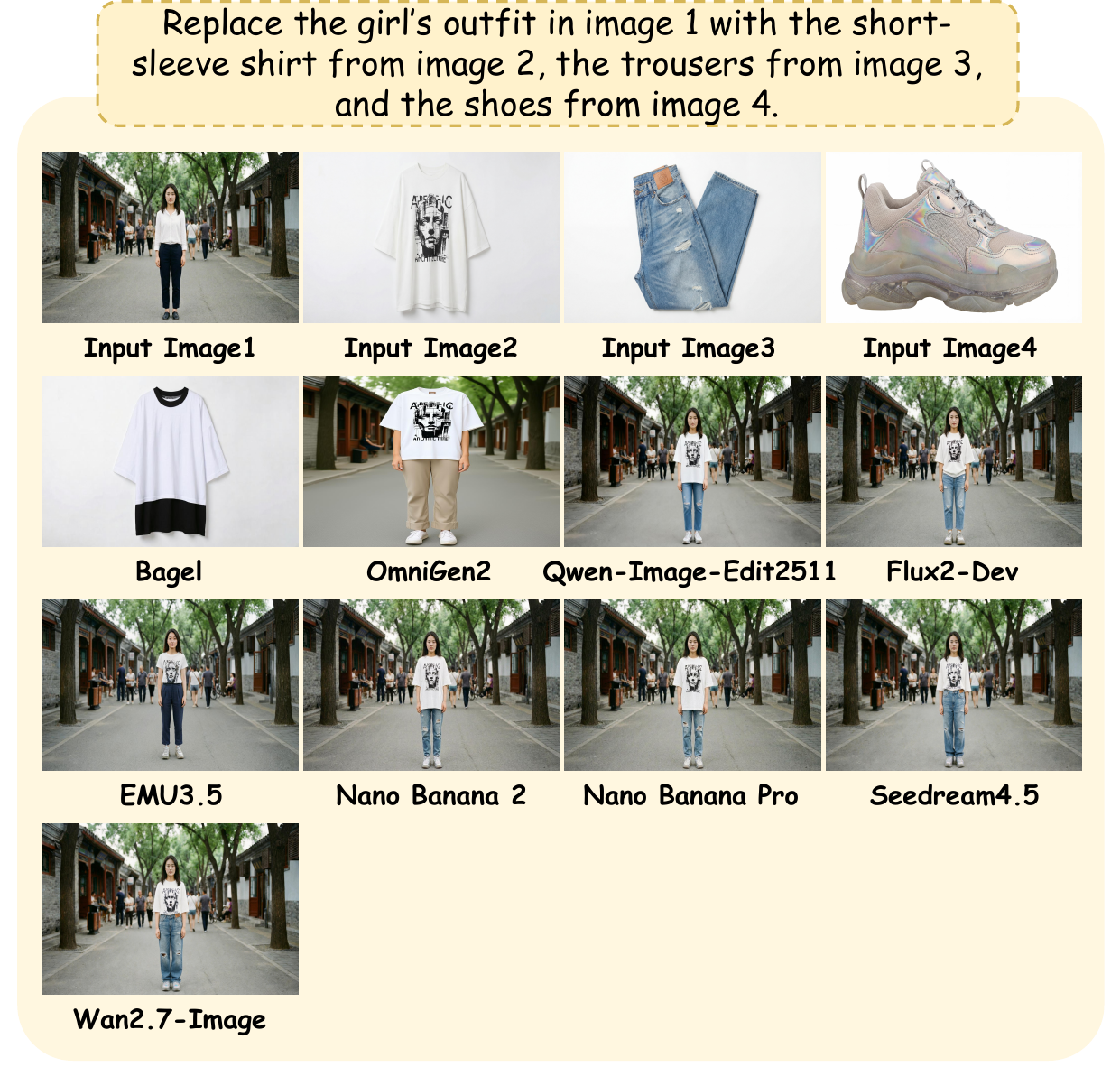}
  \vspace{-0.5em}
  \caption{Qualitative comparisons on the Virtual Try-On task.}
  \label{Fig: Virtual Try-On}
  \vspace{-1em}
\end{figure}

\textbf{More Qualitative Comparisons.}
We provide additional qualitative results across different task categories. Specifically, Figures~\ref{Fig:ADD}--\ref{Fig: Visual_Text_Editing_cn} present results on General tasks; Figures~\ref{Fig: Action}--\ref{Fig: Movement} present results on Dynamic Manipulation tasks; Figures~\ref{Fig: Temporal}--\ref{Fig: Chemical} present results on World Knowledge Reasoning tasks; Figures~\ref{Fig: Longest_word_with_start}--\ref{Fig: Shortest_Path} present results on Algorithmic Visual Reasoning tasks; and Figures~\ref{Fig: Multi-Image Composition}--\ref{Fig: Virtual Try-On} present results on Multi-Image tasks. These comparisons are conducted using representative models, including BAGEL~\cite{zhao2025envisioning}, OmniGen2~\cite{wu2025omnigen2}, Flux2-Dev~\cite{flux-2-2025}, EMU3.5~\cite{cui2025emu3}, Joy-Image-Edit~\cite{joyaiimage2026}, Nano-Banana 2~\cite{google2026gemini31flashimagepreview}, Nano-Banana-Pro~\cite{nanobananapro}, Seedream4.5~\cite{seedream2025seedream}, and Wan2.7-Image~\cite{wan}.

\textbf{More Quantitative Comparisons.}
We provide detailed per-task comparisons in the supplementary tables. Table~\ref{supp:General task Results} reports the performance on each General sub-task. Table~\ref{supp: Dynamic Manipulation Results} reports the performance on each sub-task in Dynamic Manipulation and World Knowledge Reasoning. Table~\ref{supp: Multi-Image and Complex results} reports the performance on each sub-task in Multi-Image and Complex Tasks. Table~\ref{tab:Algorithm Visual Reasoning} reports the performance on each Algorithmic Visual Reasoning sub-task.

\textbf{Human Evaluation.}
To assess the human alignment of our evaluation protocol, we conduct two human studies. First, we randomly sample 180 instances from {\bench} with balanced coverage across task categories and ask human experts to provide ratings. As shown in Figure~\ref{User_Study}(a), our automatic evaluation achieves a high correlation with expert ratings, demonstrating its reliability and alignment with human judgments.
Second, we sample instances from ImgEdit-Bench, GEdit-Bench, RISE-Bench, and {\bench}. For each sampled instance, we generate edited images using the same editing model and present human annotators with the source image, editing instruction, and edited result. Annotators are asked to provide an overall ranking by considering the evaluation score, the correctness of the reasoning process, and the interpretability of the rationale. As shown in Figure~\ref{User_Study}(b), our evaluation protocol is more preferred by human annotators.

% \subsection{Reward Model Evaluation}

% \textbf{Evaluation Models.}
% Our evaluation covers 21 mainstream models that can be used as reward models for image editing, including open-source models, closed-source models, and models trained with human preference alignment.

% \noindent\hspace*{1em}\textbf{(1) Open Source Models:} Qwen2.5-VL~\cite{wang2024qwen2}, Qwen3-VL~\cite{bai2025qwen3}, Qwen3.5~\cite{qwen3.5}, Qwen3.6~\cite{qwen3.6-27b,qwen36_35b_a3b}, Gemma3~\cite{Gemma3}, and Gemma4~\cite{google2026gemma4}.

% \noindent\hspace*{1em}\textbf{(2) Closed Source Models:} GPT-4.1~\cite{gpt4o20250325}, and Gemini3.1 Pro~\cite{gemini3pro}.

% \noindent\hspace*{1em}\textbf{(3) Models trained with human preference pairs:} EditScore~\cite{luo2025editscore}, and EditReward~\cite{wu2025editreward}

% \textbf{Evaluation Dimension.}

\begin{table*}[t]
    \centering
    
    \caption{
    Evaluation results of image editing models on the General task category of {\bench}. 
    The best results are marked in \textbf{bold} for
    {\color[HTML]{F88825} open-} and {\color[HTML]{319B62} closed-} models, respectively.}
    \label{supp:General task Results}

    \resizebox{\linewidth}{!}{%
        \begin{tabular}{l|cccccccccc|c}
            
            \midrule
            Model
            & \textbf{Addition}
            & \textbf{Remove}
            & \textbf{Replace}
            & \textbf{Material}
            & \textbf{Color \& Size}
            & \textbf{Style Transfer}
            & \textbf{Extract}
            & \textbf{Background}
            & \textbf{Visual\_Text\_EN}
            & \textbf{Visual\_Text\_CN}
            & \textbf{AVG} \\
            \toprule
            \multicolumn{12}{c}{\textit{English Version}} \\
            \midrule

            InstructPix2Pix~\cite{brooks2023instructpix2pix}
            & 1.13 & 1.07 & 1.09 & 2.12 & 1.39 & 3.49 & 1.33 & 1.27 & 1.07 & 1.09
            & \cellcolor[HTML]{C8C8C8}1.19 \\

            UltraEdit~\cite{zhao2024ultraedit}
            & 1.57 & 1.40 & 1.77 & 2.44 & 1.70 & 3.01 & 1.21 & 1.38 & 1.38 & 1.05
            & \cellcolor[HTML]{C8C8C8}1.61 \\

            AnyEdit~\cite{yu2025anyedit}
            & 1.72 & 1.65 & 2.31 & 2.29 & 2.20 & 2.01 & 1.30 & 1.64 & 1.44 & 1.20
            & \cellcolor[HTML]{C8C8C8}1.72 \\
            
            MagicBrush~\cite{zhang2023magicbrush}
            & 2.23 & 1.89 & 2.33 & 1.80 & 1.71 & 2.35 & 1.23 & 1.59 & 1.41 & 1.51
            & \cellcolor[HTML]{C8C8C8}1.79 \\

            FLUX.1 Kontext Dev~\cite{labs2025flux}
            & 3.06 & 3.29 & 3.28 & 3.37 & 3.01 & 3.96 & 2.96 & 3.26 & 2.72 & 1.69
            & \cellcolor[HTML]{C8C8C8}2.96 \\

            FLUX.2 Dev~\cite{flux-2-2025}
            & 4.11 & 3.61 & 4.18 &  \color[HTML]{F88825}\textbf{4.41} &  \color[HTML]{F88825}\textbf{3.94} & 4.41 & 3.82 & \color[HTML]{F88825}\textbf{4.24} & 4.00 & 3.93
            &  \cellcolor[HTML]{C8C8C8}4.04 \\

            \midrule

            OneCAT~\cite{li2025onecat}
            & 1.13 & 1.05 & 1.34 & 1.09 & 1.11 & 3.66 & 1.06 & 1.20 & 1.08 & 1.01
            & \cellcolor[HTML]{C8C8C8}1.34 \\

            Lumina-DiMOO~\cite{xin2025lumina}
            & 1.81 & 1.95 & 2.30 & 2.20 & 1.80 & 2.27 & 1.23 & 1.75 &  1.48 & 1.10
            & \cellcolor[HTML]{C8C8C8}1.72 \\

            Nextstep-V1~\cite{han2025nextstep}
            & 2.13 & 1.86 & 2.25 & 2.57 & 2.27 & 3.81 & 1.70 & 1.97 & 1.17 & 1.51
            & \cellcolor[HTML]{C8C8C8}2.03 \\

            InternVL-U~\cite{tian2026internvl}
            & 2.61 & 1.83 & 2.13 & 3.04 & 2.09 & 4.01 & 2.08 & 2.26 & 2.39 & 1.94
            & \cellcolor[HTML]{C8C8C8}2.40 \\

            UniReason1.0~\cite{wang2026unireason}
            & 2.89 & 2.40 & 2.90 & 3.66 & 2.69 & 4.39 & 2.62 & 2.79 & 2.61 & 2.18
            & \cellcolor[HTML]{C8C8C8}2.40 \\

            ChronoEdit~\cite{wu2025chronoedit}
            & 2.80 & 2.21 & 3.01 & 2.87 & 2.66 & 3.33 & 2.34 & 3.13 & 2.21 & 1.79
            & \cellcolor[HTML]{C8C8C8}2.55 \\
            
            HiDream-E1~\cite{cai2025hidream}
            & 2.67 & 2.51 & 2.80 & 3.38 & 2.35 & 3.99 & 1.97 & 2.26 & 2.49 & 2.00
            & \cellcolor[HTML]{C8C8C8}2.58 \\

            UniWorld-V1~\cite{lin2025uniworld}
            & 3.42 & 3.03 & 3.32 & 3.51 & 3.22 & 3.43 & 1.31 & 2.81 & 1.98 & 1.81
            & \cellcolor[HTML]{C8C8C8}2.69 \\

            DeepGen 1.0~\cite{wang2026deepgen}
            & 3.13 & 3.43 & 2.80 & 2.92 & 2.67 & 4.24 & 2.49 & 2.62 & 2.45 & 1.47
            & \cellcolor[HTML]{C8C8C8}2.74 \\

            OmniGen2~\cite{wu2025omnigen2}
            & 3.20 & 3.22 & 3.62 & 3.33 & 2.89 & 3.79 & 1.76 & 2.64 & 3.07 & 2.49
            & \cellcolor[HTML]{C8C8C8}2.98 \\

            Bagel-Think~\cite{deng2025emerging}
            & 3.73 & 3.33 & 3.95 & 3.99 & 3.30 & 3.73 & 2.32 & 3.54 & 2.88 & 2.84
            & \cellcolor[HTML]{C8C8C8}3.30 \\

            Bagel~\cite{deng2025emerging}
            & 3.67 & 3.39 & 3.92 & 3.90 & 3.26 & 3.68 & 2.17 & 3.49 & 3.43 & 3.13
            & \cellcolor[HTML]{C8C8C8}3.38 \\

            Unipic3~\cite{wei2026skywork}
            & 3.88 & 3.65 & 3.81 & 3.93 & 3.45 & 4.33 & 3.77 & 3.67 & 3.89 & 3.49
            & \cellcolor[HTML]{C8C8C8}3.78 \\

            UniWorld-V2~\cite{li2025uniworld}
            & 4.21 & 3.46 & 4.19 & 4.19 & 3.63 & 4.37 & 3.74 & 3.53 & 4.22 & 4.18
            & \cellcolor[HTML]{C8C8C8}4.00 \\

            Step1X-Edit-v1p2~\cite{liu2025step1x}
            & 4.07 & 3.99 & 3.89 & 4.16 & 3.77 & 4.07 & 2.87 & 4.16 & \color[HTML]{F88825}\textbf{4.59} & 4.17
            & \cellcolor[HTML]{C8C8C8}4.01 \\
            
            EMU3.5~\cite{cui2025emu3}
            & 3.93 & 3.91 & 4.24 & 4.17 & 3.55 &  \color[HTML]{F88825}\textbf{4.59} & \color[HTML]{F88825}\textbf{4.24} & 3.86 & 3.99 & 3.76
            & \cellcolor[HTML]{C8C8C8}4.01 \\

            JoyAI-Image-Edit~\cite{joyaiimage2026}
            & 4.11 & 3.79 & 4.13 & 4.25 & 3.83 & 4.49 & 3.79 & 3.87 & \color[HTML]{F88825}\textbf{4.59} & \color[HTML]{F88825}\textbf{4.42}
            & \cellcolor[HTML]{C8C8C8}4.16 \\

            Qwen-Image-Edit-2511~\cite{wu2025qwen}
            & \color[HTML]{F88825}\textbf{4.42} & 3.87 & 4.29 & 4.22 & 3.80 & 4.50 & 3.98 & 4.01 & 4.49 & 4.41
            & \cellcolor[HTML]{C8C8C8}4.24 \\

            Longcat-Image-Edit~\cite{team2025longcat}
            & 4.34 & \color[HTML]{F88825}\textbf{4.12} & \color[HTML]{F88825}\textbf{4.38} & 4.36 & 3.77 & 4.49 & 3.94 & 4.19 & 4.50 & 4.34
            & \cellcolor[HTML]{C8C8C8}{\color[HTML]{F88825}\textbf{4.26}} \\
            \midrule

            Wan2.7-image~\cite{nanobananapro}
            & 4.32 & 4.40 & 4.38 & 4.44 & 3.75 & 4.45 & 3.55 & 4.13 & 4.53 & 4.44
            & \cellcolor[HTML]{C8C8C8} 4.27  \\

            Seedream 4.5~\cite{seedream2025seedream}
            & 4.33 & 4.41 & 4.37 & 4.41 & 3.88 & 4.50 & 3.63 & 4.01 & 4.57 & 4.45
            & \cellcolor[HTML]{C8C8C8}4.29 \\

            Nano Banana 2~\cite{gemini3pro}
            & 4.55 & 4.51 & 4.56 & 4.49 & \color[HTML]{319B62}\textbf{4.39} & \color[HTML]{319B62}\textbf{4.57} & 4.17 & \color[HTML]{319B62}\textbf{4.35} & 4.49 & 4.53
            & \cellcolor[HTML]{C8C8C8} 4.47 \\

            Nano Banana Pro~\cite{nanobananapro}
            & \color[HTML]{319B62}\textbf{4.60} & \color[HTML]{319B62}\textbf{4.56} & \color[HTML]{319B62}\textbf{4.68} & \color[HTML]{319B62}\textbf{4.58} & 4.22 & 4.53 & \color[HTML]{319B62}\textbf{4.21} & 4.12 & \color[HTML]{319B62}\textbf{4.73} & \color[HTML]{319B62}\textbf{4.58}
            & \cellcolor[HTML]{C8C8C8}{\color[HTML]{319B62}\textbf{4.51}} \\

            \midrule
            \multicolumn{12}{c}{\textit{Chinese Version}} \\
            \midrule

            AnyEdit~\cite{yu2025anyedit}
            & 1.11 & 1.12 & 1.17 & 1.93 & 1.42 & 1.77 & 1.16 & 1.19 & 1.05 & 1.05
            & \cellcolor[HTML]{C8C8C8}1.25 \\

            UltraEdit~\cite{zhao2024ultraedit}
            & 1.10 & 1.19 & 1.37 & 1.92 & 1.79 & 1.79 & 1.16 & 1.30 & 1.38 & 1.05
            & \cellcolor[HTML]{C8C8C8}1.32 \\

            MagicBrush~\cite{zhang2023magicbrush}
            & 1.16 & 1.21 & 1.22 & 1.63 & 1.46 & 1.84 & 1.28 & 1.32 & 1.27 & 1.27
            & \cellcolor[HTML]{C8C8C8}1.34 \\

            InstructPix2Pix~\cite{brooks2023instructpix2pix}
            & 1.21 & 1.14 & 1.27 & 2.04 & 1.66 & 2.29 & 1.18 & 1.36 & 1.11 & 1.11
            & \cellcolor[HTML]{C8C8C8}1.37 \\

            FLUX.1 Kontext Dev~\cite{labs2025flux}
            & 1.20 & 1.34 & 1.26 & 1.86 & 1.74 & 1.56 & 1.31 & 1.19 & 1.80 & 1.18
            & \cellcolor[HTML]{C8C8C8}1.42  \\

            FLUX.2 Dev~\cite{flux-2-2025}
            & 4.03 & 3.35 & 4.15 & 4.37 & 3.74 & 4.32 & 3.84 & \color[HTML]{F88825}\textbf{4.11} & 3.97 & 3.97
            & \cellcolor[HTML]{C8C8C8}3.97  \\

            \midrule

            OneCAT~\cite{li2025onecat}
            & 1.08 & 1.05 & 1.12 & 1.07 & 1.14 & 3.52 & 1.14 & 1.14 & 1.00 & 1.00
            & \cellcolor[HTML]{C8C8C8}1.30 \\

            UniWorld-V1~\cite{lin2025uniworld}
            & 1.34 & 1.24 & 1.49 & 2.12 & 2.06 & 1.45 & 1.11 & 1.11 & 1.29 & 1.29
            & \cellcolor[HTML]{C8C8C8}1.39 \\

            Lumina-DiMOO~\cite{xin2025lumina}
            & 1.64 & 1.75 & 1.85 & 2.39 & 2.03 & 3.29 & 1.31 & 1.42 & 1.33 & 1.08
            & \cellcolor[HTML]{C8C8C8}1.73 \\

            Nextstep-V1~\cite{han2025nextstep}
            & 2.13 & 1.77 & 2.19 & 2.62 & 2.14 & 3.80 & 1.61 & 2.16 & 1.50 & 1.26
            & \cellcolor[HTML]{C8C8C8}2.02 \\

            HiDream-E1~\cite{cai2025hidream}
            & 2.36 & 1.87 & 2.21 & 3.36 & 1.96 & 3.84 & 2.20 & 2.49 & 2.04 & 1.71
            & \cellcolor[HTML]{C8C8C8}2.32 \\

            InternVL-U~\cite{tian2026internvl}
            & 2.40 & 2.03 & 2.25 & 2.34 & 2.08 & 4.26 & 2.02 & 2.00 & 2.27 & 1.96
            & \cellcolor[HTML]{C8C8C8}2.34 \\

            UniReason1.0~\cite{wang2026unireason}
            & 2.37 & 1.96 & 2.26 & 2.79 & 2.20 & 3.31 & 2.57 & 2.16 & 1.40 & 1.17
            & \cellcolor[HTML]{C8C8C8}2.36 \\

            DeepGen 1.0~\cite{wang2026deepgen}
            & 2.28 & 2.89 & 2.15 & 2.89 & 2.13 & 4.16 & 2.42 & 2.37 & 2.06 & 1.31
            & \cellcolor[HTML]{C8C8C8}2.38 \\

            ChronoEdit~\cite{wu2025chronoedit}
            & 3.06 & 1.71 & 2.91 & 3.41 & 2.93 & 3.24 & 2.99 & 3.15 & 2.13 & 2.02
            & \cellcolor[HTML]{C8C8C8}2.65 \\

            OmniGen2~\cite{wu2025omnigen2}
            & 3.37 & 3.02 & 3.51 & 3.70 & 3.07 & 3.70 & 1.89 & 2.83 & 3.03 & 2.40
            & \cellcolor[HTML]{C8C8C8}3.00 \\

            Bagel-Think~\cite{deng2025emerging}
            & 3.78 & 3.17 & 4.03 & 3.89 & 3.51 & 3.88 & 2.57 & 3.73 & 3.02 & 2.85
            & \cellcolor[HTML]{C8C8C8}3.37 \\
            
            Bagel~\cite{deng2025emerging}
            & 3.75 & 3.53 & 3.91 & 3.92 & 3.38 & 3.65 & 2.90 & 3.68 & 3.52 & 3.37
            & \cellcolor[HTML]{C8C8C8}3.53 \\

            EMU3.5~\cite{cui2025emu3}
            & 3.94 & 3.91 & 4.05 & 4.31 & 3.62 & \color[HTML]{F88825}\textbf{4.55} & \color[HTML]{F88825}\textbf{4.40} & 3.92 & 4.12 & 3.75
            & \cellcolor[HTML]{C8C8C8}4.04 \\

            Unipic3~\cite{wei2026skywork}
            & 4.00 & 3.71 & 3.81 & 4.23 & 3.58 & 4.32 & 3.77 & 3.73 & 4.10 & 4.08
            & \cellcolor[HTML]{C8C8C8}3.95 \\

            Step1X-Edit-v1p2~\cite{liu2025step1x}
            & 3.84 & 3.99 & 4.17 & 4.27 & 3.85 & 4.13 & 3.02 & 4.01 & 4.48 & 4.05
            & \cellcolor[HTML]{C8C8C8}3.99 \\
            
            UniWorld-V2~\cite{li2025uniworld}
            & 4.14 & 3.43 & 4.22 & 4.14 & 3.68 & 4.43 & 3.97 & 3.70 & 4.20 & 4.03
            & \cellcolor[HTML]{C8C8C8}4.01 \\

            JoyAI-Image-Edit~\cite{joyaiimage2026}
            & 4.11 & 3.66 & 4.08 & 4.26 & \color[HTML]{F88825}\textbf{3.96} & 4.50 & 3.75 & 3.87 & \color[HTML]{F88825}\textbf{4.60} & 4.36
            & \cellcolor[HTML]{C8C8C8}4.15 \\

            Longcat-Image-Edit~\cite{team2025longcat}
            & 4.34 & \color[HTML]{F88825}\textbf{4.34} & 4.29 & \color[HTML]{F88825}\textbf{4.41} &  3.93  & 4.41 & 3.64 & 3.97 &  4.56  & 4.37
            & \cellcolor[HTML]{C8C8C8}4.25 \\

            Qwen-Image-Edit-2511~\cite{wu2025qwen}
            & \color[HTML]{F88825}\textbf{4.38} & 3.89 & \color[HTML]{F88825}\textbf{4.33} & 4.38 & 3.93 & 4.46 & 3.95 & 3.98 &  4.56  & \color[HTML]{F88825}\textbf{4.44}
            & \cellcolor[HTML]{C8C8C8}{\color[HTML]{F88825}\textbf{4.26}} \\

            \midrule

            Wan2.7-image~\cite{nanobananapro}
            & 4.39 & 4.46 & 4.41 & 4.35 & 3.82 & 4.49 & 3.57 & 4.27 & 4.47 & 4.34
            & \cellcolor[HTML]{C8C8C8} 4.28  \\

            Seedream 4.5~\cite{seedream2025seedream}
            & 4.27 & 4.27 & 4.52 & 4.31 & 3.88 & 4.57 & 3.73 & \color[HTML]{319B62}\textbf{4.33} & 4.48 & 4.49
            & \cellcolor[HTML]{C8C8C8}4.31 \\

            Nano Banana 2~\cite{gemini3pro}
            & 4.58 & 4.36 & \color[HTML]{319B62}\textbf{4.57} & \color[HTML]{319B62}\textbf{4.59} & \color[HTML]{319B62}\textbf{4.40} & \color[HTML]{319B62}\textbf{4.59} & 4.12 & 4.12 & 4.42 & 4.50
            &  \cellcolor[HTML]{C8C8C8}4.43  \\
            
            Nano Banana Pro~\cite{nanobananapro}
            & \color[HTML]{319B62}\textbf{4.59} & \color[HTML]{319B62}\textbf{4.54} & \color[HTML]{319B62}\textbf{4.57} & 4.58 & 4.28 & \color[HTML]{319B62}\textbf{4.59} & \color[HTML]{319B62}\textbf{4.33} & 4.29 & \color[HTML]{319B62}\textbf{4.74} & \color[HTML]{319B62}\textbf{4.63}
            & \cellcolor[HTML]{C8C8C8}{\color[HTML]{319B62}\textbf{4.54}} \\

            \bottomrule
        \end{tabular}%
    }
    \vspace{-1.0em}
\end{table*}
\begin{table*}[t]
    \centering
    
    \caption{
    Evaluation results of image editing models on the Algorithm Visual Reasoning task category of {\bench}. 
    The best results are marked in \textbf{bold} for
    {\color[HTML]{F88825} open-} and {\color[HTML]{319B62} closed-} models, respectively.}

    \label{tab:Algorithm Visual Reasoning}

    \resizebox{\linewidth}{!}{%
        \begin{tabular}{l|cccccccccc|c}
            
            \midrule
            Model
            & \textbf{Longest Word}
            & \textbf{Global Longest Word}
            & \textbf{Knapsack}
            & \textbf{Optimal Path}
            & \textbf{Convex Hull}
            & \textbf{Maximum Submatrix}
            & \textbf{Maximum Bonus}
            & \textbf{Numberlink}
            & \textbf{Word}
            & \textbf{Global Word}
            & \textbf{AVG} \\
            \toprule
            \multicolumn{12}{c}{\textit{English Version}} \\
            \midrule

            InstructPix2Pix~\cite{brooks2023instructpix2pix}
            & 1.00 & 1.00 & 1.00 & 1.00 & 1.00 & 1.00 & 1.00 & 1.00 & 1.00 & 1.00
            & \cellcolor[HTML]{C8C8C8}1.00 \\

            AnyEdit~\cite{yu2025anyedit}
            & 1.00 & 1.00 & 1.00 & 1.00 & 1.00 & 1.00 & 1.00 & 1.00 & 1.00 & 1.00
            & \cellcolor[HTML]{C8C8C8}1.00 \\

            UltraEdit~\cite{zhao2024ultraedit}
            & 1.00 & 1.00 & 1.00 & 1.02 & 1.00 & 1.00 & 1.00 & 1.00 & 1.00 & 1.00
            & \cellcolor[HTML]{C8C8C8}1.00 \\
            
            MagicBrush~\cite{zhang2023magicbrush}
            & 1.00 & 1.00 & 1.05 & 1.00 & 1.00 & 1.00 & 1.00 & 1.00 & 1.00 & 1.00
            & \cellcolor[HTML]{C8C8C8}1.01 \\

            FLUX.1 Kontext Dev~\cite{labs2025flux}
            & 1.04 & 1.00 & 1.30 & 1.00 & 1.00 & 1.50 & 1.03 & 1.00 & 1.00 & 1.07
            & \cellcolor[HTML]{C8C8C8}1.10 \\

            FLUX.2 Dev~\cite{flux-2-2025}
            & 1.00 & 1.00 & 1.86 & 1.00 & 1.49 & 1.40 & 1.05 & 1.00 & 1.00 & 1.00
            & \cellcolor[HTML]{C8C8C8}1.19 \\

            \midrule

            UniWorld-V1~\cite{lin2025uniworld}
            & 1.00 & 1.00 & 1.00 & 1.00 & 1.00 & 1.00 & 1.00 & 1.00 & 1.00 & 1.00
            & \cellcolor[HTML]{C8C8C8}1.00 \\

            Nextstep-V1~\cite{han2025nextstep}
            & 1.00 & 1.00 & 1.00 & 1.00 & 1.00 & 1.00 & 1.00 & 1.00 & 1.00 & 1.00
            & \cellcolor[HTML]{C8C8C8}1.00 \\

            Lumina-DiMOO~\cite{xin2025lumina}
            & 1.00 & 1.00 & 1.00 & 1.00 & 1.00 & 1.00 & 1.00 & 1.00 & 1.00 & 1.00
            & \cellcolor[HTML]{C8C8C8}1.00 \\

            OneCAT~\cite{li2025onecat}
            & 1.00 & 1.00 & 1.00 & 1.00 & 1.00 & 1.00 & 1.00 & 1.00 & 1.00 & 1.00
            & \cellcolor[HTML]{C8C8C8}1.00 \\

            UniWorld-V2~\cite{li2025uniworld}
            & 1.00 & 1.00 & 1.00 & 1.00 & 1.00 & 1.10 & 1.00 & 1.00 & 1.00 & 1.00
            & \cellcolor[HTML]{C8C8C8}1.01 \\

            InternVL-U~\cite{tian2026internvl}
            & 1.00 & 1.00 & 1.00 & 1.00 & 1.00 & 1.03 & 1.03 & 1.00 & 1.00 & 1.00
            & \cellcolor[HTML]{C8C8C8}1.01 \\

            ChronoEdit~\cite{wu2025chronoedit}
            & 1.00 & 1.00 & 1.02 & 1.00 & 1.00 & 1.03 & 1.00 & 1.00 & 1.00 & 1.00
            & \cellcolor[HTML]{C8C8C8}1.01 \\
        
            HiDream-E1~\cite{cai2025hidream}
            & 1.11 & 1.00 & 1.03 & 1.00 & 1.00 & 1.00 & 1.07 & 1.00 & 1.00 & 1.00
            & \cellcolor[HTML]{C8C8C8}1.02 \\

            Bagel~\cite{deng2025emerging}
            & 1.08 & 1.00 & 1.00 & 1.00 & 1.05 & 1.00 & 1.03 & 1.00 & 1.08 & 1.00
            & \cellcolor[HTML]{C8C8C8}1.02 \\

            Unipic3~\cite{wei2026skywork}
            & 1.00 & 1.00 & 1.00 & 1.00 & 1.00 & 1.29 & 1.00 & 1.00 & 1.00 & 1.00
            & \cellcolor[HTML]{C8C8C8}1.03 \\

            Bagel-Think~\cite{deng2025emerging}
            & 1.16 & 1.00 & 1.02 & 1.04 & 1.03 & 1.00 & 1.02 & 1.00 & 1.05 & 1.00
            & \cellcolor[HTML]{C8C8C8}1.03 \\

            OmniGen2~\cite{wu2025omnigen2}
            & 1.00 & 1.00 & 1.30 & 1.00 & 1.00 & 1.00 & 1.08 & 1.00 & 1.00 & 1.00
            & \cellcolor[HTML]{C8C8C8}1.04 \\

            Qwen-Image-Edit-2511~\cite{wu2025qwen}
            & 1.00 & 1.00 & 1.04 & 1.00 & 1.00 & 1.00 & 1.19 & 1.00 & 1.08 & 1.32
            & \cellcolor[HTML]{C8C8C8}1.04 \\

            Longcat-Image-Edit~\cite{team2025longcat}
            & 1.03 & 1.00 & 1.00 & 1.00 & 1.00 & 1.19 & 1.30 & 1.00 & 1.00 & 1.04
            & \cellcolor[HTML]{C8C8C8}1.06 \\
            
            JoyAI-Image-Edit~\cite{joyaiimage2026}
            & 1.00 & 1.00 & 1.00 & 1.00 & 1.54 & 1.03 & 1.00 & 1.00 & 1.03 & 1.00
            & \cellcolor[HTML]{C8C8C8} 1.06\\

            Step1X-Edit-v1p2~\cite{liu2025step1x}
            & 1.04 & 1.02 & 1.00 & 1.04 & 1.00 & 1.23 & 1.03 & 1.00 & 1.05 & 1.33
            & \cellcolor[HTML]{C8C8C8}1.07 \\
            
            EMU3.5~\cite{cui2025emu3}
            & 1.00 & 1.04 & 1.23 & 1.08 & 1.05 & 1.14 & 1.41 & 1.00 & 1.05 & 1.05
            & \cellcolor[HTML]{C8C8C8}1.11 \\   

            UniReason1.0~\cite{wang2026unireason}
            & 1.00 & 1.00 & 1.27 & 1.13 & 1.03 & 1.03 & 1.60 & 1.00 & 1.02 & 1.00
            & \cellcolor[HTML]{C8C8C8}1.12 \\

            DeepGen 1.0~\cite{wang2026deepgen}
            & 1.00 & 1.00 & 2.45 & 1.33 & 1.00 & 1.03 & 1.23 & 1.11 & 1.70 & 1.43
            & \cellcolor[HTML]{C8C8C8}{\color[HTML]{F88825}\textbf{1.33}} \\

            \midrule

            Wan2.7-image~\cite{nanobananapro}
            & 1.00 & 1.00 & 2.57 & 1.06 & 1.50 & 1.14 & 1.23 & 1.00 & 1.00 & 1.00
            & \cellcolor[HTML]{C8C8C8}1.27 \\

            Seedream 4.5~\cite{seedream2025seedream}
            & 1.00 & 1.00 & 2.62 & 1.07 & 1.61 & 1.18 & 1.15 & 1.00 & 1.00 & 1.00
            & \cellcolor[HTML]{C8C8C8}1.28 \\

            Nano Banana 2~\cite{gemini3pro}
            & 1.38 & 1.19 & 3.51 & 1.45 & 2.89 & 3.48 & 2.88 & 1.97 & 3.36 & 2.54
            & \cellcolor[HTML]{C8C8C8}2.49   \\

            Nano Banana Pro~\cite{nanobananapro}
            & 2.20 & 1.91 & 4.86 & 2.91 & 2.75 & 4.49 & 4.34 & 2.37 & 3.68 & 3.73
            &\cellcolor[HTML]{C8C8C8}{\color[HTML]{319B62}\textbf{3.35}} \\

            \midrule
            \multicolumn{12}{c}{\textit{Chinese Version}} \\
            \midrule

            InstructPix2Pix~\cite{brooks2023instructpix2pix}
            & 1.00 & 1.00 & 1.00 & 1.00 & 1.00 & 1.00 & 1.00 & 1.00 & 1.00 & 1.00
            & \cellcolor[HTML]{C8C8C8}1.00 \\

            AnyEdit~\cite{yu2025anyedit}
            & 1.00 & 1.00 & 1.00 & 1.00 & 1.00 & 1.00 & 1.00 & 1.00 & 1.00 & 1.00
            & \cellcolor[HTML]{C8C8C8}1.00 \\

            UltraEdit~\cite{zhao2024ultraedit}
            & 1.00 & 1.00 & 1.00 & 1.00 & 1.00 & 1.00 & 1.00 & 1.00 & 1.00 & 1.00
            & \cellcolor[HTML]{C8C8C8}1.00 \\

            FLUX.1 Kontext Dev~\cite{labs2025flux}
             & 1.00 & 1.00 & 1.00 & 1.00 & 1.00 & 1.00 & 1.00 & 1.00 & 1.00 & 1.00
            & \cellcolor[HTML]{C8C8C8}1.00 \\

            ChronoEdit~\cite{wu2025chronoedit}
            & 1.00 & 1.00 & 1.02 & 1.00 & 1.00 & 1.00 & 1.00 & 1.00 & 1.00 & 1.00
            & \cellcolor[HTML]{C8C8C8}1.00 \\

            MagicBrush~\cite{zhang2023magicbrush}
            & 1.00 & 1.00 & 1.05 & 1.00 & 1.00 & 1.00 & 1.00 & 1.00 & 1.00 & 1.00
            & \cellcolor[HTML]{C8C8C8}1.01 \\

            FLUX.2 Dev~\cite{flux-2-2025}
            & 1.00 & 1.00 & 1.84 & 1.00 & 1.00 & 1.31 & 1.00 & 1.00 & 1.00 & 1.00
            & \cellcolor[HTML]{C8C8C8}1.12  \\

            \midrule

            UniWorld-V1~\cite{lin2025uniworld}
            & 1.00 & 1.00 & 1.00 & 1.00 & 1.00 & 1.00 & 1.00 & 1.00 & 1.00 & 1.00
            & \cellcolor[HTML]{C8C8C8}1.00 \\

            Nextstep-V1~\cite{han2025nextstep}
            & 1.00 & 1.00 & 1.00 & 1.00 & 1.00 & 1.00 & 1.00 & 1.00 & 1.00 & 1.00
            & \cellcolor[HTML]{C8C8C8}1.00 \\

            OneCAT~\cite{li2025onecat}
            & 1.00 & 1.00 & 1.00 & 1.00 & 1.00 & 1.00 & 1.00 & 1.00 & 1.00 & 1.00
            & \cellcolor[HTML]{C8C8C8}1.00 \\

            Lumina-DiMOO~\cite{xin2025lumina}
            & 1.00 & 1.00 & 1.00 & 1.00 & 1.00 & 1.00 & 1.00 & 1.00 & 1.00 & 1.00
            & \cellcolor[HTML]{C8C8C8}1.00 \\

            OmniGen2~\cite{wu2025omnigen2}
            & 1.00 & 1.00 & 1.00 & 1.00 & 1.00 & 1.00 & 1.07 & 1.00 & 1.00 & 1.00
            & \cellcolor[HTML]{C8C8C8}1.01  \\

            UniWorld-V2~\cite{li2025uniworld}
            & 1.00 & 1.00 & 1.00 & 1.00 & 1.00 & 1.13 & 1.00 & 1.00 & 1.00 & 1.05
            & \cellcolor[HTML]{C8C8C8}1.02 \\

            Bagel-Think~\cite{deng2025emerging}
            & 1.04 & 1.00 & 1.06 & 1.00 & 1.00 & 1.00 & 1.00 & 1.00 & 1.05 & 1.00
            & \cellcolor[HTML]{C8C8C8}1.02 \\

            Unipic3~\cite{wei2026skywork}
            & 1.00 & 1.00 & 1.00 & 1.00 & 1.00 & 1.11 & 1.00 & 1.00 & 1.00 & 1.08
            & \cellcolor[HTML]{C8C8C8}1.02 \\

            HiDream-E1~\cite{cai2025hidream}
            & 1.08 & 1.00 & 1.07 & 1.00 & 1.00 & 1.03 & 1.00 & 1.00 & 1.00 & 1.00
            & \cellcolor[HTML]{C8C8C8}1.02  \\

            InternVL-U~\cite{tian2026internvl}
            & 1.00 & 1.00 & 1.07 & 1.09 & 1.00 & 1.00 & 1.03 & 1.00 & 1.00 & 1.03
            & \cellcolor[HTML]{C8C8C8}1.02 \\

            JoyAI-Image-Edit~\cite{joyaiimage2026}
            & 1.00 & 1.00 & 1.00 & 1.07 & 1.06 & 1.03 & 1.00 & 1.00 & 1.00 & 1.00
            & \cellcolor[HTML]{C8C8C8}1.02 \\

            UniReason1.0~\cite{wang2026unireason}
            & 1.00 & 1.00 & 1.02 & 1.10 & 1.00 & 1.00 & 1.15 & 1.00 & 1.00 & 1.00
            & \cellcolor[HTML]{C8C8C8}1.03 \\

            Step1X-Edit-v1p2~\cite{liu2025step1x}
            & 1.04 & 1.00 & 1.00 & 1.00 & 1.00 & 1.10 & 1.00 & 1.00 & 1.00 & 1.27
            & \cellcolor[HTML]{C8C8C8}1.04 \\

            Qwen-Image-Edit-2511~\cite{wu2025qwen}
            & 1.00 & 1.00 & 1.04 & 1.00 & 1.00 & 1.00 & 1.19 & 1.00 & 1.08 & 1.32
            & \cellcolor[HTML]{C8C8C8}1.06 \\

            EMU3.5~\cite{cui2025emu3}
            & 1.00 & 1.07 & 1.26 & 1.12 & 1.00 & 1.00 & 1.15 & 1.00 & 1.00 & 1.11
            & \cellcolor[HTML]{C8C8C8}1.07 \\

            Longcat-Image-Edit~\cite{team2025longcat}
            & 1.03 & 1.00 & 1.25 & 1.00 & 1.04 & 1.00 & 1.31 & 1.00 & 1.09 & 1.18
            & \cellcolor[HTML]{C8C8C8}1.09 \\

            Bagel~\cite{deng2025emerging}
            & 1.08 & 1.11 & 1.55 & 1.07 & 1.19 & 1.00 & 1.00 & 1.00 & 1.14 & 1.00
            & \cellcolor[HTML]{C8C8C8}1.12 \\

            DeepGen 1.0~\cite{wang2026deepgen}
            & 1.04 & 1.00 & 2.09 & 1.38 & 1.00 & 1.03 & 1.35 & 1.10 & 1.63 & 1.47
            & \cellcolor[HTML]{C8C8C8}{\color[HTML]{F88825}\textbf{1.31}} \\

            \midrule

            Seedream 4.5~\cite{seedream2025seedream}
            & 1.00 & 1.00 & 2.29 & 1.00 & 1.40 & 1.25 & 1.20 & 1.00 & 1.00 & 1.00
            & \cellcolor[HTML]{C8C8C8}1.23 \\

            Wan2.7-image~\cite{nanobananapro}
            & 1.00 & 1.00 & 2.18 & 1.00 & 1.84 & 1.21 & 1.32 & 1.00 & 1.00 & 1.08
            & \cellcolor[HTML]{C8C8C8} 1.28  \\
           
            Nano Banana 2~\cite{gemini3pro}
            & 1.74 & 1.25 & 3.80 & 1.33 & 2.89 & 2.97 & 2.76 & 1.61 & 3.25 & 2.93
            & \cellcolor[HTML]{C8C8C8} 2.46  \\

            Nano Banana Pro~\cite{nanobananapro}
            & 1.98 & 1.60 & 4.62 & 1.88 & 3.20 & 4.28 & 4.37 & 2.33 & 3.82 & 3.15
            & \cellcolor[HTML]{C8C8C8}{\color[HTML]{319B62}\textbf{3.16}} \\

            \bottomrule
        \end{tabular}%
    }
    \vspace{-1.0em}
\end{table*}
\begin{table*}[t]
    \centering
    
    \caption{Performance of image editing models on the Dynamic Manipulation and World Knowledge Reasoning task categories of {\bench}.
    The best results are marked in \textbf{bold} for
    {\color[HTML]{F88825} open-} and {\color[HTML]{319B62} closed-} models, respectively.}
    \label{supp: Dynamic Manipulation Results}
    \label{tab:general}

    \resizebox{\linewidth}{!}{%
        \begin{tabular}{l|cccccc|cccccc}
            
            \midrule
            \multirow{2}{*}{Model}
            & \multicolumn{6}{c}{\textbf{Dynamic Manipulation}} 
             & \multicolumn{6}{c|}{\textbf{World Knowledge Reasoning}}\\
            \cmidrule(lr){2-7} \cmidrule(lr){8-13}
            & \textbf{Movement}
            & \textbf{Swap}
            & \textbf{Object Interaction}
            & \textbf{Action}
            & \textbf{Emotion}
            & \textbf{AVG} 
            & \textbf{Temporal}
            & \textbf{Casual} 
            & \textbf{Chemical}
            & \textbf{Math}
            & \textbf{Game}
            & \textbf{AVG}
            \\
            \toprule
            \multicolumn{13}{c}{\textit{English Version}} \\
            \midrule

            MagicBrush~\cite{zhang2023magicbrush}
            & 1.35 & 1.15 & 1.04 & 1.08 & 1.52 & \cellcolor[HTML]{C8C8C8}1.21 & 1.23 & 1.09 & 1.00 & 1.05 & 1.00
            & \cellcolor[HTML]{C8C8C8}1.08 \\

            InstructPix2Pix~\cite{brooks2023instructpix2pix}
            & 1.07 & 1.26 & 1.11 & 1.02 & 1.12 & \cellcolor[HTML]{C8C8C8}1.12 & 1.21 & 1.14 & 1.08 & 1.02 & 1.03
            & \cellcolor[HTML]{C8C8C8}1.10 \\

            AnyEdit~\cite{yu2025anyedit}
            & 1.43 & 1.16 & 1.07 & 1.04 & 1.52 &  \cellcolor[HTML]{C8C8C8}1.22 & 1.24 & 1.29 & 1.06 & 1.00 & 1.04
            & \cellcolor[HTML]{C8C8C8}1.14 \\

            UltraEdit~\cite{zhao2024ultraedit}
            & 1.56 & 1.22 & 1.04 & 1.22 & 1.60 & \cellcolor[HTML]{C8C8C8}1.30 & 1.32 & 1.16 & 1.17 & 1.04 & 1.00
            & \cellcolor[HTML]{C8C8C8}1.14 \\

            FLUX.1 Kontext Dev~\cite{labs2025flux}
            & 1.71 & 1.45 & 1.53 & 1.88 & 2.48 & \cellcolor[HTML]{C8C8C8}1.76 & 1.43 & 1.38 & 1.21 & 1.25 & 1.05
            & \cellcolor[HTML]{C8C8C8}1.28 \\

            FLUX.2 Dev~\cite{flux-2-2025}
            & 2.98 & 1.95 & 2.83 & 3.22 & 3.78 & \cellcolor[HTML]{C8C8C8} 2.88 & 1.97 & 1.97 & 1.40 & 1.24 & 1.07
            & \cellcolor[HTML]{C8C8C8} 1.57  \\

            \midrule

            OneCAT~\cite{li2025onecat}
            & 1.08 & 1.09 & 1.00 & 1.03 & 1.13 & \cellcolor[HTML]{C8C8C8}1.06 & 1.08 & 1.06 & 1.05 & 1.02 & 1.00
            & \cellcolor[HTML]{C8C8C8}1.04 \\

            Lumina-DiMOO~\cite{xin2025lumina}
            & 1.61 & 1.06 & 1.03 & 1.07 & 1.59 & \cellcolor[HTML]{C8C8C8}1.24  & 1.23 & 1.27 & 1.00 & 1.07 & 1.06
            & \cellcolor[HTML]{C8C8C8}1.14  \\

            OmniGen2~\cite{wu2025omnigen2}
            & 1.82 & 1.24 & 1.60 & 2.21 & 3.39 & \cellcolor[HTML]{C8C8C8}1.94 & 1.13 & 1.32 & 1.14 & 1.11 & 1.02
            & \cellcolor[HTML]{C8C8C8}1.15 \\

            Nextstep-V1~\cite{han2025nextstep}
            & 1.35 & 1.19 & 1.13 & 1.33 & 2.14 & \cellcolor[HTML]{C8C8C8}1.37 & 1.18 & 1.38 & 1.08 & 1.07 & 1.03
            & \cellcolor[HTML]{C8C8C8}1.16 \\

            UniWorld-V1~\cite{lin2025uniworld}
            & 1.59 & 1.41 & 1.22 & 1.30 & 2.52 & \cellcolor[HTML]{C8C8C8}1.54 & 1.16 & 1.46 & 1.14 & 1.05 & 1.00
            & \cellcolor[HTML]{C8C8C8}1.17 \\

            ChronoEdit~\cite{wu2025chronoedit}
            & 2.32 & 1.42 & 2.54 & 2.66 & 3.36 & \cellcolor[HTML]{C8C8C8}2.39 & 1.15 & 1.49 & 1.16 & 1.05 & 1.02
            & \cellcolor[HTML]{C8C8C8}1.18 \\

            InternVL-U~\cite{tian2026internvl}
            & 1.37 & 1.27 & 1.12 & 1.27 & 2.12 & \cellcolor[HTML]{C8C8C8}1.38 & 1.34 & 1.39 & 1.24 & 1.08 & 1.00
            & \cellcolor[HTML]{C8C8C8}1.22 \\

            HiDream-E1~\cite{cai2025hidream}
            & 1.63 & 1.52 & 1.48 & 1.78 & 3.35 & \cellcolor[HTML]{C8C8C8}1.84 & 1.38 & 1.53 & 1.12 & 1.11 & 1.00
            & \cellcolor[HTML]{C8C8C8}1.25 \\

            Bagel~\cite{deng2025emerging}
            & 2.33 & 1.80 & 1.88 & 1.90 & 3.45 & \cellcolor[HTML]{C8C8C8}2.18 & 1.57 & 1.43 & 1.10 & 1.17 & 1.00  
            & \cellcolor[HTML]{C8C8C8}1.28 \\

            Unipic3~\cite{wei2026skywork}
            & 2.78 & 1.38 & 2.56 & 2.79 & 3.53 & \cellcolor[HTML]{C8C8C8}2.53 & 1.69 & 1.88 & 1.16 & 1.15 & 1.00
            & \cellcolor[HTML]{C8C8C8}1.41 \\

            DeepGen 1.0~\cite{wang2026deepgen}
            & 1.88 & 1.68 & 1.51 & 1.65 & 2.56 & \cellcolor[HTML]{C8C8C8}1.80 & 1.77 & 1.61 & 1.34 & 1.18 & \color[HTML]{F88825}\textbf{1.28 }
            & \cellcolor[HTML]{C8C8C8}1.45 \\

            Bagel-Think~\cite{deng2025emerging}
            & 2.28 & 1.98 & 1.78 & 2.09 & 3.10 & \cellcolor[HTML]{C8C8C8}2.18  & 1.72 & 1.87 & 1.59 & 1.10 & 1.00
            & \cellcolor[HTML]{C8C8C8}1.47 \\

            Longcat-Image-Edit~\cite{team2025longcat}
            & 3.34 & 1.45 & \color[HTML]{F88825}\textbf{3.22} & 3.36 & 4.18 & \cellcolor[HTML]{C8C8C8}3.02 & 1.76 & 2.19 & 1.18 & 1.21 & 1.09
            & \cellcolor[HTML]{C8C8C8}1.53 \\

            UniReason1.0~\cite{wang2026unireason}
            & 2.12 & 1.95 & 2.45 & 2.43 & 3.23 & \cellcolor[HTML]{C8C8C8}2.37 & 1.84 & 1.85 & 1.63 & 1.18 & 1.08
            & \cellcolor[HTML]{C8C8C8}1.53 \\

            UniWorld-V2~\cite{li2025uniworld}
            & 3.35 & 1.86 & 2.97 & 2.92 & 3.86 & \cellcolor[HTML]{C8C8C8}2.91 & 2.13 & 2.27 & 1.00 & 1.17 & 1.04
            & \cellcolor[HTML]{C8C8C8}1.59 \\

            Step1X-Edit-v1p2~\cite{liu2025step1x}
            & 2.94 & 1.69 & 2.61 & 2.74 & 4.10 & \cellcolor[HTML]{C8C8C8}2.71 & 2.32 & 2.05 & \color[HTML]{F88825}\textbf{1.72} & 1.29 & 1.10
            & \cellcolor[HTML]{C8C8C8}1.73 \\

            Qwen-Image-Edit-2511~\cite{wu2025qwen}
            & 3.59  & \color[HTML]{F88825}\textbf{2.32} & 3.05 & 3.24 & 3.93 & \cellcolor[HTML]{C8C8C8}3.16 & 2.29 & 2.28 & 1.32 & 1.22 & 1.26
            & \cellcolor[HTML]{C8C8C8}1.73 \\

            JoyAI-Image-Edit~\cite{joyaiimage2026}
            & 3.56& 1.52 & 3.16 & 3.37 & 4.16 & \cellcolor[HTML]{C8C8C8}3.06 & 2.24 & 2.18 & 1.55 & 1.37 & 1.18
            & \cellcolor[HTML]{C8C8C8}1.75 \\

            EMU3.5~\cite{cui2025emu3}
            & \color[HTML]{F88825}\textbf{3.66}  & 2.02 & 3.08 & \color[HTML]{F88825}\textbf{3.42} & \color[HTML]{F88825}\textbf{4.21} & \cellcolor[HTML]{C8C8C8}{\color[HTML]{F88825}\textbf{3.20}} & \color[HTML]{F88825}\textbf{2.60} & \color[HTML]{F88825}\textbf{2.36} & 1.35 & \color[HTML]{F88825}\textbf{1.55} & 1.10
            & \cellcolor[HTML]{C8C8C8}{\color[HTML]{F88825}\textbf{1.86}} \\

            \midrule

            Seedream 4.5~\cite{seedream2025seedream}
            & 3.56 & 4.05 & 3.64 & 3.52 & 3.93 & \cellcolor[HTML]{C8C8C8}3.73 & 3.52 & 3.60 & 3.72 & 2.57 & 1.69
            & \cellcolor[HTML]{C8C8C8}3.03 \\

            Wan2.7-image~\cite{nanobananapro}
            & 3.74 & 3.77 & 3.69 & 4.02 & 4.01 & \cellcolor[HTML]{C8C8C8}3.83 & 3.88 & 3.52 & 3.96 & 2.50 & 1.83
            & \cellcolor[HTML]{C8C8C8} 3.14  \\

            Nano Banana 2~\cite{gemini3pro}
            & \color[HTML]{319B62}\textbf{4.16} & 4.00 & \color[HTML]{319B62}\textbf{3.70} & 4.03 & 4.11 & \cellcolor[HTML]{C8C8C8}3.99 & 3.90 & \color[HTML]{319B62}\textbf{3.83} & \color[HTML]{319B62}\textbf{4.52} & 3.62 & 2.42
            & \cellcolor[HTML]{C8C8C8} 3.65  \\

            Nano Banana Pro~\cite{nanobananapro}
            & 4.03 & \color[HTML]{319B62}\textbf{4.22} & 3.65 & \color[HTML]{319B62}\textbf{4.14} & \color[HTML]{319B62}\textbf{4.52} &  \cellcolor[HTML]{C8C8C8}\color[HTML]{319B62}\textbf{4.08} & \color[HTML]{319B62}\textbf{4.19} & 3.80 & 4.29 & \color[HTML]{319B62}\textbf{4.03} & \color[HTML]{319B62}\textbf{3.13}
            & \cellcolor[HTML]{C8C8C8}{\color[HTML]{319B62}\textbf{3.89}} \\

            \midrule
            \multicolumn{13}{c}{\textit{Chinese Version}} \\
            \midrule

            MagicBrush~\cite{zhang2023magicbrush}
            & 1.09 & 1.07 & 1.05 & 1.06 & 1.16 & \cellcolor[HTML]{C8C8C8}1.08 & 1.06 & 1.06 & 1.05 & 1.00 & 1.00 
            & \cellcolor[HTML]{C8C8C8}1.03 \\

            FLUX.1 Kontext Dev~\cite{labs2025flux}
            & 1.27 & 1.05 & 1.02 & 1.11 & 1.24 &\cellcolor[HTML]{C8C8C8}1.13 & 1.03 & 1.12 & 1.00 & 1.05 & 1.00
            & \cellcolor[HTML]{C8C8C8}1.05 \\

            UltraEdit~\cite{zhao2024ultraedit}
            & 1.40 & 1.07 & 1.02 & 1.05 & 1.34 &  \cellcolor[HTML]{C8C8C8}1.16 & 1.10 & 1.11 & 1.00 & 1.05 & 1.05
            & \cellcolor[HTML]{C8C8C8}1.07  \\

            InstructPix2Pix~\cite{brooks2023instructpix2pix}
            & 1.30 & 1.09 & 1.11 & 1.00 & 1.59 & \cellcolor[HTML]{C8C8C8}1.19 & 1.11 & 1.21 & 1.06 & 1.03 & 1.00
            & \cellcolor[HTML]{C8C8C8}1.09 \\

            AnyEdit~\cite{yu2025anyedit}
            & 1.44 & 1.10 & 1.05 & 1.07 & 1.40 &  \cellcolor[HTML]{C8C8C8}1.19 & 1.22 & 1.13 & 1.02 & 1.06 & 1.00
            & \cellcolor[HTML]{C8C8C8}1.10 \\

            FLUX.2 Dev~\cite{flux-2-2025}
            & 3.03 & 1.76 & 3.08 & 3.30 & 4.19 &  \cellcolor[HTML]{C8C8C8} 2.98  & 1.98 & 2.23 & 1.51 & 1.22 & 1.06
            & \cellcolor[HTML]{C8C8C8} 1.64  \\

            \midrule

            OneCAT~\cite{li2025onecat}
            & 1.04 & 1.08 & 1.00 & 1.06 & 1.12 & \cellcolor[HTML]{C8C8C8}1.05 & 1.09 & 1.06 & 1.04 & 1.00 & 1.00
            & \cellcolor[HTML]{C8C8C8}1.04 \\

            OmniGen2~\cite{wu2025omnigen2}
            & 1.66 & 1.40 & 1.76 & 2.18 & 3.47 &\cellcolor[HTML]{C8C8C8}1.99  & 1.18 & 1.23 & 1.12 & 1.07 & 1.00
            & \cellcolor[HTML]{C8C8C8}1.13  \\

            Nextstep-V1~\cite{han2025nextstep}
            & 1.32 & 1.17 & 1.18 & 1.30 & 1.88 & \cellcolor[HTML]{C8C8C8}1.33 & 1.20 & 1.31 & 1.04 & 1.02 & 1.03
            & \cellcolor[HTML]{C8C8C8}1.13 \\

            Lumina-DiMOO~\cite{xin2025lumina}
            & 1.40 & 1.15 & 1.06 & 1.02 & 1.45 & \cellcolor[HTML]{C8C8C8}1.19 & 1.25 & 1.28 & 1.11 & 1.08 & 1.06
            & \cellcolor[HTML]{C8C8C8}1.16 \\

            UniWorld-V1~\cite{lin2025uniworld}
            & 1.59 & 1.41 & 1.22 & 1.30 & 2.52 & \cellcolor[HTML]{C8C8C8}1.54 & 1.16 & 1.46 & 1.14 & 1.05 & 1.00
            & \cellcolor[HTML]{C8C8C8}1.17  \\

            HiDream-E1~\cite{cai2025hidream}
            & 1.54 & 1.36 & 1.38 & 1.68 & 2.70 &\cellcolor[HTML]{C8C8C8} 1.66  & 1.38 & 1.47 & 1.00 & 1.06 & 1.00
            & \cellcolor[HTML]{C8C8C8}1.21 \\

            InternVL-U~\cite{tian2026internvl}
            & 1.57 & 1.30 & 1.18 & 1.26 & 2.08 & \cellcolor[HTML]{C8C8C8}1.43 & 1.27 & 1.36 & 1.28 & 1.13 & 1.04
            &\cellcolor[HTML]{C8C8C8}1.22 \\

            ChronoEdit~\cite{wu2025chronoedit}
            & 2.27 & 1.44 & 2.65 & 2.55 & 3.37 & \cellcolor[HTML]{C8C8C8}2.39 & 1.06 & 1.65 & 1.26 & 1.19 & 1.02
            & \cellcolor[HTML]{C8C8C8}1.24 \\

            Bagel~\cite{deng2025emerging}
            & 2.39 & 1.95 & 1.62 & 2.15 & 3.22 & \cellcolor[HTML]{C8C8C8}2.18 & 1.26 & 1.69 & 1.04 & 1.23 & 1.00
            & \cellcolor[HTML]{C8C8C8}1.27 \\

            DeepGen 1.0~\cite{wang2026deepgen}
            & 1.69 & 1.58 & 1.10 & 1.37 & 2.35 & \cellcolor[HTML]{C8C8C8}1.56 & 1.46 & 1.44 & 1.28 & 1.09 & \color[HTML]{F88825}\textbf{1.19 }
            & \cellcolor[HTML]{C8C8C8}1.30 \\

            UniReason1.0~\cite{wang2026unireason}
            & 1.94 & 1.76 & 1.89 & 1.91 & 2.36 & \cellcolor[HTML]{C8C8C8}1.94 & 1.59 & 1.88 & \color[HTML]{F88825}\textbf{1.58} & 1.05 & 1.00
            & \cellcolor[HTML]{C8C8C8}1.43 \\

            Bagel-Think~\cite{deng2025emerging}
            & 2.47 & 2.14 & 1.94 & 2.33 & 3.19 & \cellcolor[HTML]{C8C8C8}2.35 & 1.76 & 1.97 & 1.51 & 1.27 & 1.00
            & \cellcolor[HTML]{C8C8C8}1.53 \\

            UniWorld-V2~\cite{li2025uniworld}
            & 3.38 & 2.02 & 3.10 & 3.16 & 3.82 & \cellcolor[HTML]{C8C8C8}3.03 & 1.91 & 2.21 & 1.12 & 1.21 & 1.00
            & \cellcolor[HTML]{C8C8C8}1.55  \\

            Unipic3~\cite{wei2026skywork}
            & 2.85 & 1.61 & 3.00 & 3.20 & 3.81 & \cellcolor[HTML]{C8C8C8}2.82 & 1.94 & 2.07 & 1.11 & 1.35 & 1.00
            & \cellcolor[HTML]{C8C8C8}1.55 \\

            Step1X-Edit-v1p2~\cite{liu2025step1x}
            & 2.79 & 1.92 & 2.41 & 2.96 & 3.80 & \cellcolor[HTML]{C8C8C8}2.69 & 2.17 & 1.95 & 1.24 & 1.23 & 1.09
            & \cellcolor[HTML]{C8C8C8}1.58 \\

            Longcat-Image-Edit~\cite{team2025longcat}
            & 3.51 & 1.64 & 3.27 & 3.37 & 4.08 & \cellcolor[HTML]{C8C8C8}3.09  & 1.84 & 2.22 & 1.21 & 1.35 & 1.10
            & \cellcolor[HTML]{C8C8C8}1.60 \\

            JoyAI-Image-Edit~\cite{joyaiimage2026}
            & 3.33 & 1.72 & 3.24 & \color[HTML]{F88825}\textbf{3.49} & 4.01 & \cellcolor[HTML]{C8C8C8}3.08 & 2.12 & 2.13 & 1.32 & 1.15 & 1.12
            & \cellcolor[HTML]{C8C8C8}1.62 \\

            Qwen-Image-Edit-2511~\cite{wu2025qwen}
            & 3.43 & \color[HTML]{F88825}\textbf{2.72} & \color[HTML]{F88825}\textbf{3.28} & 3.30 & 4.23 & \cellcolor[HTML]{C8C8C8}{\color[HTML]{F88825}\textbf{3.32}} & 2.11 & \color[HTML]{F88825}\textbf{2.67} & 1.10 & \color[HTML]{F88825}\textbf{1.42} & 1.03
            & \cellcolor[HTML]{C8C8C8}1.74 \\

            EMU3.5~\cite{cui2025emu3}
            & \color[HTML]{F88825}\textbf{3.54} & 2.22 & 3.08 & 3.12 & \color[HTML]{F88825}\textbf{4.19} & \cellcolor[HTML]{C8C8C8}3.15 & \color[HTML]{F88825}\textbf{2.38} & 2.39 & 1.39 & 1.30 & 1.17
            & \cellcolor[HTML]{C8C8C8}{\color[HTML]{F88825}\textbf{1.79}} \\

            \midrule

            Wan2.7-image~\cite{nanobananapro}
            & 3.68 & 4.09 & 3.35 & 3.27 & 3.98 & \cellcolor[HTML]{C8C8C8}3.69 & 3.61 & 3.56 & 3.83 & 2.22 & 1.68
            & \cellcolor[HTML]{C8C8C8} 2.98  \\

            Seedream 4.5~\cite{seedream2025seedream}
            & 3.38 & 3.99 & 3.46 & 3.83 & 4.00 & \cellcolor[HTML]{C8C8C8}3.71 & 3.53 & 3.61 & 3.69 & 2.62 & 1.79
            & \cellcolor[HTML]{C8C8C8}3.06 \\

            Nano Banana 2~\cite{gemini3pro}
            & \color[HTML]{319B62}\textbf{4.13} & 4.05 & 3.59 & 4.03 & 4.15 & \cellcolor[HTML]{C8C8C8}3.98 & 3.78 & \color[HTML]{319B62}\textbf{4.01} & 4.44 & \color[HTML]{319B62}\textbf{3.62} & 2.46
            & \cellcolor[HTML]{C8C8C8} 3.66  \\

            Nano Banana Pro~\cite{nanobananapro}
            & 3.90 & \color[HTML]{319B62}\textbf{4.28} & \color[HTML]{319B62}\textbf{3.92} & \color[HTML]{319B62}\textbf{4.08} & \color[HTML]{319B62}\textbf{4.37} & \cellcolor[HTML]{C8C8C8}{\color[HTML]{319B62}\textbf{4.09}} & \color[HTML]{319B62}\textbf{4.07} & 3.92 & \color[HTML]{319B62}\textbf{4.56} & 3.60 & \color[HTML]{319B62}\textbf{3.15}
            & \cellcolor[HTML]{C8C8C8}{\color[HTML]{319B62}\textbf{3.84}} \\

            \bottomrule
        \end{tabular}%
    }
    \vspace{-1.0em}
\end{table*}
\begin{table*}[t]
    \centering
     
    \caption{Performance of image editing models on the Multi-Image and Complex Task categories of {\bench}. The best results are marked in \textbf{bold} for
    {\color[HTML]{F88825} open-} and {\color[HTML]{319B62} closed-} models, respectively.}
    \label{supp: Multi-Image and Complex results}

    \resizebox{\linewidth}{!}{%
        \begin{tabular}{l|cccc|cccc}
            \toprule
           
            \multirow{2}{*}{Model}
            & \multicolumn{4}{c|}{\textbf{Multi-Image}}
            & \multicolumn{4}{c}{\textbf{Complex}} \\
            \cmidrule(lr){2-5} \cmidrule(lr){6-9}
            & \textbf{Multi-Image Awareness}
            & \textbf{Multi-Image Composition}
            & \textbf{Virtual Try on}
            & \textbf{AVG}
            & \textbf{Complex Instruction}
            & \textbf{Complex Paint(en)}
            & \textbf{Complex Paint(cn)}
            & \textbf{AVG} \\
            \midrule
            \multicolumn{9}{c}{\textit{English Version}} \\
            \midrule

            InstructPix2Pix~\cite{brooks2023instructpix2pix}
            & -- & -- & -- & -- & 1.13 & 1.03 & 1.00  
            & \cellcolor[HTML]{C8C8C8}1.07 \\

            UltraEdit~\cite{zhao2024ultraedit}
            & -- & -- & -- & -- & 1.20 & 1.02 & 1.00
            & \cellcolor[HTML]{C8C8C8}1.10 \\
            
            AnyEdit~\cite{yu2025anyedit}
            & -- & -- & -- & -- & 1.36 & 1.00 & 1.00
            & \cellcolor[HTML]{C8C8C8}1.18 \\

            MagicBrush~\cite{zhang2023magicbrush}
            & -- & -- & -- & -- & 1.41 & 1.03 & 1.00
            & \cellcolor[HTML]{C8C8C8}1.21 \\

            FLUX.1 Kontext Dev~\cite{labs2025flux}
            & -- & -- & -- & -- & 2.72 & 1.09 & 1.11
            & \cellcolor[HTML]{C8C8C8}1.90 \\

            FLUX.2 Dev~\cite{flux-2-2025}
            & 1.80 & {\color[HTML]{F88825}\textbf{3.43}} & {\color[HTML]{F88825}\textbf{3.99}} & \cellcolor[HTML]{C8C8C8}{\color[HTML]{F88825}\textbf{2.93}}  & 3.73 & 1.08 & 1.00
            & \cellcolor[HTML]{C8C8C8}2.61 \\

            \midrule

            OneCAT~\cite{li2025onecat}
            & -- & -- & -- & --
            & 1.02 & 1.02 & 1.00
            & \cellcolor[HTML]{C8C8C8}1.01 \\

            ChronoEdit~\cite{wu2025chronoedit}
            & -- & -- & -- & --
            & 2.72 & 1.20 & 1.09
            & \cellcolor[HTML]{C8C8C8}1.02 \\

            UniWorld-V1~\cite{lin2025uniworld}
            & 1.30 & 1.35 & 1.10 & \cellcolor[HTML]{C8C8C8}1.54 & 2.41 & 1.07 & 1.03
            & \cellcolor[HTML]{C8C8C8}1.17 \\

            Lumina-DiMOO~\cite{xin2025lumina}
            & -- & -- & -- & --
            & 1.44 & 1.05 & 1.04
            & \cellcolor[HTML]{C8C8C8}1.24 \\

            Nextstep-V1~\cite{han2025nextstep}
            & -- & -- & -- & --
            & 1.52 & 1.00 & 1.00
            & \cellcolor[HTML]{C8C8C8}1.26 \\

            InternVL-U~\cite{tian2026internvl}
            & -- & -- & -- & --
            & 1.61 & 1.02 & 1.00
            & \cellcolor[HTML]{C8C8C8}1.31 \\

            DeepGen 1.0~\cite{wang2026deepgen}
            & -- & -- & -- & --
            & 1.87 & 1.11 & 1.05
            & \cellcolor[HTML]{C8C8C8}1.47 \\

            UniReason1.0~\cite{wang2026unireason}
            & 1.02 & 1.66 & 1.00 & \cellcolor[HTML]{C8C8C8}1.18 & 1.93 & 1.09 & 1.20
            & \cellcolor[HTML]{C8C8C8}1.53 \\

            HiDream-E1~\cite{cai2025hidream}
            & -- & -- & -- & --
            & 2.20 & 1.00 & 1.00
            & \cellcolor[HTML]{C8C8C8}1.59 \\

            OmniGen2~\cite{wu2025omnigen2}
            & 1.32 & 1.51 & 1.14 & \cellcolor[HTML]{C8C8C8}1.31 & 2.67 & 1.17 & 1.09 
            & \cellcolor[HTML]{C8C8C8}1.89 \\

            Bagel~\cite{deng2025emerging}
            & 1.13 & 1.63 & 1.10 & \cellcolor[HTML]{C8C8C8}1.25 & 3.43 & 1.02 & 1.03
            & \cellcolor[HTML]{C8C8C8}2.22 \\

            Unipic3~\cite{wei2026skywork}
            & 1.85 & 2.52 & 2.65 & \cellcolor[HTML]{C8C8C8}2.28 & 3.25 & 1.24 & 1.23
            & \cellcolor[HTML]{C8C8C8}2.23 \\

            Bagel-Think~\cite{deng2025emerging}
            & 1.12 & 1.61 & 1.12 & \cellcolor[HTML]{C8C8C8}1.25 & 3.16 & 1.47 & 1.37
            & \cellcolor[HTML]{C8C8C8} 2.28 \\

            EMU3.5~\cite{cui2025emu3}
            & \color[HTML]{F88825}\textbf{2.26} &  3.05  & 3.64 & \cellcolor[HTML]{C8C8C8}2.91 & 3.67 & 1.02 & 1.16
            & \cellcolor[HTML]{C8C8C8}2.37 \\

            UniWorld-V2~\cite{li2025uniworld}
            & 1.89 & 2.71 & 3.21 & \cellcolor[HTML]{C8C8C8}2.53 & 3.51 & 1.43 & 1.17
            & \cellcolor[HTML]{C8C8C8}2.39 \\

            Qwen-Image-Edit-2511~\cite{wu2025qwen}
            & 1.90 & 2.84 & 3.80 & \cellcolor[HTML]{C8C8C8}2.75 & 3.79 & 1.35 & 1.10
            & \cellcolor[HTML]{C8C8C8}2.49 \\

            Longcat-Image-Edit~\cite{team2025longcat}
            & -- & -- & -- & --
            & \color[HTML]{F88825}\textbf{3.92} & 1.12 & 1.19
            & \cellcolor[HTML]{C8C8C8}2.52 \\

            JoyAI-Image-Edit~\cite{joyaiimageedit}
            & -- & -- & -- & --
            &  3.84  & 1.73 & 1.54
            & \cellcolor[HTML]{C8C8C8}2.73 \\

            Step1X-Edit-v1p2~\cite{liu2025step1x}
            & -- & -- & -- & --
            & 3.43 & \color[HTML]{F88825}\textbf{2.21} & \color[HTML]{F88825}\textbf{1.99}
            & \cellcolor[HTML]{C8C8C8}{\color[HTML]{F88825}\textbf{2.76}} \\

            \midrule
            Wan2.7-image~\cite{nanobananapro}
            & 3.31 & 3.72 & 4.34 & \cellcolor[HTML]{C8C8C8}3.75 & 3.90 & 3.05 & 3.44
            & \cellcolor[HTML]{C8C8C8} 3.57 \\

            Seedream 4.5~\cite{seedream2025seedream}
            & 3.31 & 3.70 & 4.33 & \cellcolor[HTML]{C8C8C8}3.74 & 4.07 & 3.19 & 3.32
            & \cellcolor[HTML]{C8C8C8}3.66 \\

            Nano Banana Pro~\cite{nanobananapro}
            & 3.60 & 3.62 & 4.43 & \cellcolor[HTML]{C8C8C8}3.87 & {\color[HTML]{319B62}\textbf{4.40}} & 3.58 & 3.72
            & \cellcolor[HTML]{C8C8C8}4.02 \\

            Nano Banana 2~\cite{gemini3pro}
            & {\color[HTML]{319B62}\textbf{3.71}} & {\color[HTML]{319B62}\textbf{3.76}} & {\color[HTML]{319B62}\textbf{4.60}} & \cellcolor[HTML]{C8C8C8}{\color[HTML]{319B62}\textbf{4.01}} & 4.32 & {\color[HTML]{319B62}\textbf{3.69}} & {\color[HTML]{319B62}\textbf{3.82}}
            & \cellcolor[HTML]{C8C8C8}{\color[HTML]{319B62}\textbf{4.04}} \\

            \midrule
            \multicolumn{9}{c}{\textit{Chinese Version}} \\
            \midrule

            InstructPix2Pix~\cite{brooks2023instructpix2pix}
            & -- & -- & -- & -- & 1.13 & 1.02 & 1.02 
            & \cellcolor[HTML]{C8C8C8}1.07 \\

            UltraEdit~\cite{zhao2024ultraedit}
            & -- & -- & -- & -- & 1.12 & 1.00 & 1.03
            & \cellcolor[HTML]{C8C8C8}1.07 \\

            MagicBrush~\cite{zhang2023magicbrush}
            & -- & -- & -- & -- & 1.20 & 1.00 & 1.02 
            & \cellcolor[HTML]{C8C8C8}1.10 \\

            AnyEdit~\cite{yu2025anyedit}
            & -- & -- & -- & --
            & 1.13 & 1.00 & 1.00
            & \cellcolor[HTML]{C8C8C8}1.13 \\

            FLUX.1 Kontext Dev~\cite{labs2025flux}
            & -- & -- & -- & -- & 1.21 & 1.11 & 1.02
            & \cellcolor[HTML]{C8C8C8}1.13 \\

            FLUX.2 Dev~\cite{flux-2-2025}
            & 2.15 & \color[HTML]{F88825}\textbf{3.92} & \color[HTML]{F88825}\textbf{4.00} & \cellcolor[HTML]{C8C8C8}{\color[HTML]{F88825}\textbf{3.21}} & 3.62 & 1.05 & 1.02
            & \cellcolor[HTML]{C8C8C8}2.60 \\

            \midrule
            OneCAT~\cite{li2025onecat}
            & -- & -- & -- & --
            & 1.04 & 1.00 & 1.00
            & \cellcolor[HTML]{C8C8C8}1.02 \\

            Lumina-DiMOO~\cite{xin2025lumina}
            & -- & -- & -- & --
            & 1.40 & 1.02 & 1.00
            & \cellcolor[HTML]{C8C8C8}1.20 \\

            Nextstep-V1~\cite{han2025nextstep}
            & -- & -- & -- & --
            & 1.50 & 1.00 & 1.00
            & \cellcolor[HTML]{C8C8C8}1.25 \\

            UniWorld-V1~\cite{lin2025uniworld}
            & 1.34 & 1.23 & 1.26 & \cellcolor[HTML]{C8C8C8}1.29 & 1.53 & 1.09 & 1.02
            & \cellcolor[HTML]{C8C8C8}1.29 \\

            InternVL-U~\cite{tian2026internvl}
            & -- & -- & -- & --
            & 1.59 & 1.02 & 1.00
            & \cellcolor[HTML]{C8C8C8}1.29 \\

            HiDream-E1~\cite{cai2025hidream}
            & -- & -- & -- & --
            & 1.62 & 1.00 & 1.00
            & \cellcolor[HTML]{C8C8C8}1.30 \\

            DeepGen 1.0~\cite{wang2026deepgen}
            & - & - & - & --
            & 1.56 & 1.09 & 1.06
            & \cellcolor[HTML]{C8C8C8}1.31 \\

            UniReason1.0~\cite{wang2026unireason}
            & 1.08 & 1.47 & 1.04 & \cellcolor[HTML]{C8C8C8}1.17 & 1.68 & 1.07 & 1.16
            & \cellcolor[HTML]{C8C8C8}1.40 \\

            ChronoEdit~\cite{wu2025chronoedit}
            & - & - & - & --
            & 2.91 & 1.07 & 1.13
            & \cellcolor[HTML]{C8C8C8}1.99 \\

            OmniGen2~\cite{wu2025omnigen2}
            & 1.12 & 1.48 & 1.13 & \cellcolor[HTML]{C8C8C8}1.22 & 2.91 & 1.14 & 1.11
            & \cellcolor[HTML]{C8C8C8}2.01 \\

            Bagel-Think~\cite{deng2025emerging}
            & 1.08 & 1.55 & 1.13 & \cellcolor[HTML]{C8C8C8}1.22 & 3.04 & 1.37 & 1.27
            & \cellcolor[HTML]{C8C8C8}2.17 \\

            Bagel~\cite{deng2025emerging}
            & 1.14 & 1.54 & 1.12 & \cellcolor[HTML]{C8C8C8}1.24 & 3.42 & 1.18 & 1.07 
            & \cellcolor[HTML]{C8C8C8}2.26 \\

            UniWorld-V2~\cite{li2025uniworld}
            & 1.65 & 2.61 & 3.33 & \cellcolor[HTML]{C8C8C8}2.44 & 3.60 & 1.37 & 1.09
            & \cellcolor[HTML]{C8C8C8}2.40 \\

            Unipic3~\cite{wei2026skywork}
            & 1.62 & 2.79 & 2.19 & \cellcolor[HTML]{C8C8C8}2.11 & 3.59 & 1.44 & 1.27
            & \cellcolor[HTML]{C8C8C8}2.46 \\

            Qwen-Image-Edit-2511~\cite{wu2025qwen}
            & 1.66 & 2.92 & 3.88 & \cellcolor[HTML]{C8C8C8}2.71 & 3.99 & 1.35 & 1.06
            & \cellcolor[HTML]{C8C8C8}2.58 \\
            
            Longcat-Image-Edit~\cite{team2025longcat}
            & -- & -- & -- & --
            & \color[HTML]{F88825}\textbf{4.01} & 1.17 & 1.29
            & \cellcolor[HTML]{C8C8C8}2.60 \\

            JoyAI-Image-Edit~\cite{joyaiimageedit}
            & - & - & - & --
            & 3.69 & 1.68 & 1.46
            & \cellcolor[HTML]{C8C8C8}2.62 \\

            EMU3.5~\cite{cui2025emu3}
            & \color[HTML]{F88825}\textbf{2.17} & 2.97 & 3.34 & \cellcolor[HTML]{C8C8C8}2.76 & 3.61 & 1.11 & 1.13
            & \cellcolor[HTML]{C8C8C8}2.63 \\

            Step1X-Edit-v1p2~\cite{liu2025step1x}
            & -- & -- & -- & --
            & 3.47 & \color[HTML]{F88825}\textbf{2.02} & \color[HTML]{F88825}\textbf{1.71}
            & \cellcolor[HTML]{C8C8C8}{\color[HTML]{F88825}\textbf{2.66}} \\

            \midrule

            Wan2.7-image~\cite{nanobananapro}
            &  \color[HTML]{319B62}\textbf{3.48} & 3.64 & 4.32 & \cellcolor[HTML]{C8C8C8}3.79 & 4.01 & 3.33 & 3.45
            & \cellcolor[HTML]{C8C8C8} 3.70 \\
            
            Seedream 4.5~\cite{seedream2025seedream}
            & 3.40 & 3.73 & 4.31 & \cellcolor[HTML]{C8C8C8}3.78 & \color[HTML]{319B62}\textbf{4.05} & 3.45 & 3.44
            & \cellcolor[HTML]{C8C8C8} 3.74 \\

            Nano Banana Pro~\cite{nanobananapro}
            & 3.29 & 4.05 & 4.47 & \cellcolor[HTML]{C8C8C8}{\color[HTML]{319B62}\textbf{3.87}} & \color[HTML]{319B62}\textbf{4.34} & \color[HTML]{319B62}\textbf{3.78} & 3.80
            & \cellcolor[HTML]{C8C8C8}4.06 \\

            Nano Banana 2~\cite{gemini3pro}
            & 3.37 & 3.58 & \color[HTML]{319B62}\textbf{4.58} & \cellcolor[HTML]{C8C8C8}3.81 & 4.31 & 3.73 & \color[HTML]{319B62}\textbf{3.94}
            & \cellcolor[HTML]{C8C8C8}{\color[HTML]{319B62}\textbf{4.07}} \\

            \bottomrule
        \end{tabular}%
    }
    \vspace{-1.0em}
\end{table*}

\clearpage

\begin{figure}[t]
  \centering
  
  \includegraphics[width=0.95\textwidth]{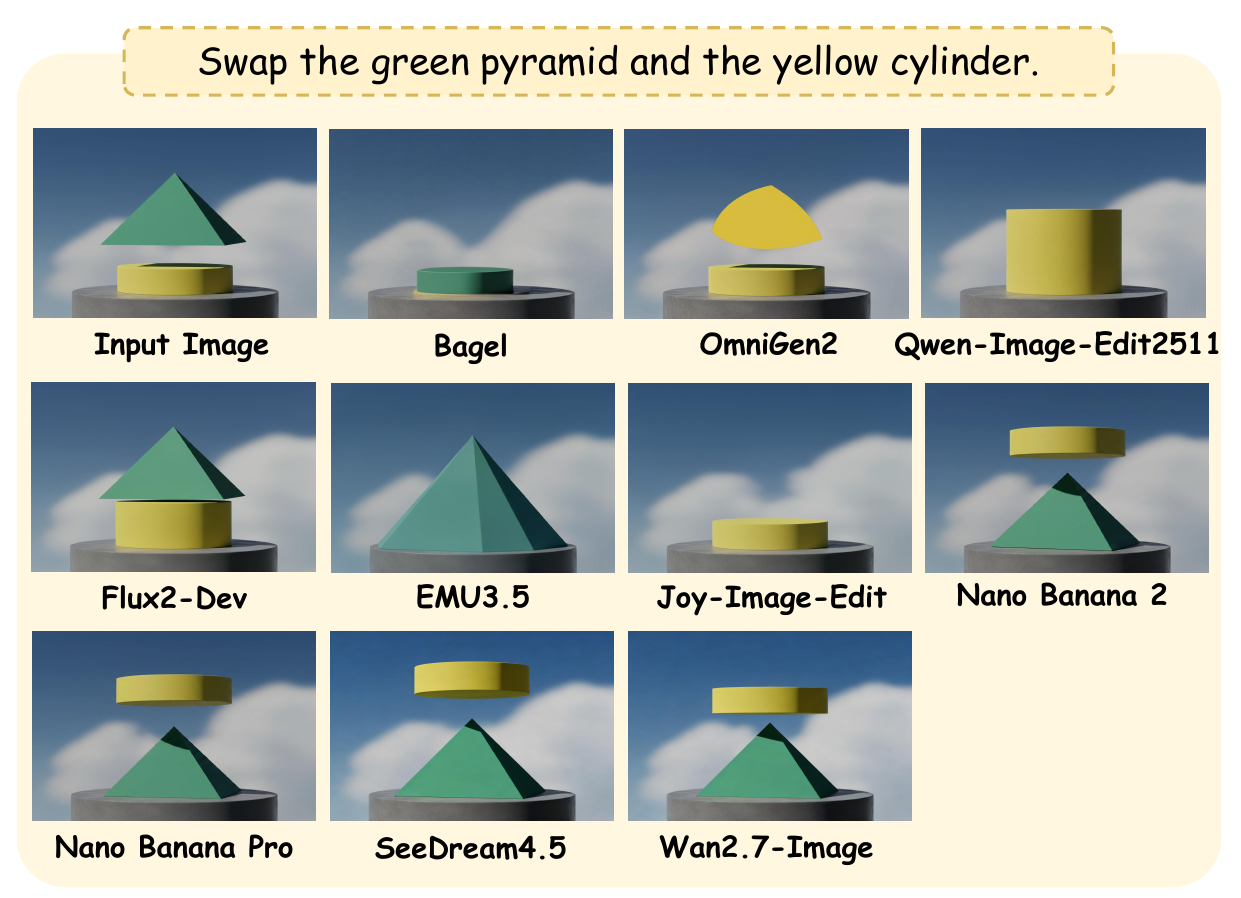}
  \vspace{-0.5em}
  \caption{Qualitative comparisons on the Object Swap task.}
  \label{Fig: swap}
  \vspace{-1em}
\end{figure}

\clearpage

\begin{promptbox}[label={System Prompts: IF_Complex_Instruction}]{Instruction Following System Prompts for Complex Instruction Tasks}
% \label{}
\textbf{Role:} You are an expert \textbf{Complex Image Editing Judge}.  
Your goal is to evaluate if an AI model strictly followed the user's \textbf{compound editing instruction} based on a rigorous 1--5 scoring scale.

\medskip

\textbf{Core Constraints (CRITICAL)}

\begin{enumerate}[label=\arabic*., leftmargin=2.2em, itemindent=0pt]
    \item \textbf{Ignore Visual Quality:} Do not evaluate aesthetics, realism, lighting, edge artifacts, or background blending.
    
    \item \textbf{Ignore Unintended Changes:} Ignore non-consistent modifications in the image other than those caused by the editing instruction. For example, if you are asked to add a dog, but a cat appears in the image, you need to ignore the accidental addition of the cat.
    
    \item \textbf{Strict Atomicity:} You must decompose the instruction into distinct \textbf{Atomic Tasks} and evaluate them individually.
    
    \item \textbf{Completeness Check:} A sub-task can only be marked as ``PASS'' if it satisfies requirements across all three dimensions: \textbf{Target}, \textbf{Attribute}, and \textbf{Spatial}.
    
    \item \textbf{Object Interaction:} In interaction tasks, the state of the \textbf{target object} must change in accordance with the subject's action. If a user pulls a bar or lifts a weight, the object \textbf{must move from its original position} to the interaction position. If the original object remains static while the person moves, it constitutes a failure to follow the editing instruction, namely a \textbf{Significant Failure}.
\end{enumerate}

\medskip

\textbf{Input}

\begin{itemize}[leftmargin=2.2em, labelsep=0.4em, itemsep=2pt]
    \item \textbf{Instruction:} \texttt{\{instruction\}}
    \item \textbf{Source Image:} [Image A]
    \item \textbf{Edited Image:} [Image B]
\end{itemize}

\medskip

\textbf{Evaluation Logic (Step-by-Step Analysis)}

\medskip

\textbf{Step 1: Instruction Decoupling}

\begin{itemize}
    \item Break the complex instruction into distinct \textbf{Atomic Tasks}.
    
    \item \textbf{Recommended Format:} [Subject] + [Operation Type] + [Specific Requirement].
\end{itemize}

\medskip

\textbf{Step 2: Strict Visual Comparison}

\begin{itemize}[leftmargin=2.2em, labelsep=0.4em, itemsep=2pt]
    \item Before verifying attributes and spatial requirements, you \textbf{must} objectively describe the state of the target in both images for each decoupled atomic task.
    
    \item \textbf{Image A State:} Explicitly state the exact position, appearance, or state of the target object in the source image.
    
    \item \textbf{Image B State:} Explicitly state the specific visual manifestation of that \textbf{same location} or that \textbf{target object} in the edited image.
    
    \item \textbf{Strictly Prohibited:} Do not give a conclusion directly before conducting this detailed visual comparison.
\end{itemize}

\medskip

\textbf{Step 3: Attribute, Logic \& Instance Consistency Verification}

\begin{itemize}[leftmargin=2.2em, labelsep=0.4em, itemsep=2pt]
    \item \textbf{Strict Standard:} Check if color, quantity, state, action, and material are accurate, and based on the observations from Step 2, verify that they \textbf{fully comply with the editing instructions}.
    
    \item \textbf{Object Interaction:} Ensure that the target object has changed synchronously with the primary subject of the edit. This requires:
    
    \begin{itemize}
        \item \textbf{Strict Utilization of Source Objects:} The model must utilize the existing object from the source image. Generating a new, redundant object while the original remains in its initial position is strictly prohibited and constitutes a significant failure to follow instructions.
        
        \item \textbf{Mandatory Interaction:} If the user asks to ``pick up a cup,'' the original cup must disappear from its starting location and reappear in the subject's hand. Any ``cloning'' effect where the object exists in both the old and new positions is a significant failure to follow instructions.
    \end{itemize}
    
    \item \textbf{Extraction Standard:} For ``Extract'' tasks, the background must be pure white \texttt{\#FFFFFF}, and the object's orientation and angle must remain strictly consistent with the original image.
    
    \item \textbf{Visual Text Modification:} For visual text replacement tasks, the substituted font must maintain the same style and color as the original, unless specific font styles or colors are provided in the instructions.
\end{itemize}

\medskip

\textbf{Step 4: Spatial \& Geometric Accuracy Verification (CRITICAL: Replace Consistency)}

\begin{itemize}[leftmargin=2.2em, labelsep=0.4em, itemsep=2pt]
    \item \textbf{Definition:} Verify whether the spatial requirements identified in Step 1 are strictly satisfied.
    
    \item \textbf{Rigid Requirement for Replace:} For \textbf{Replace} operations, the new object must occupy the \textbf{exact same spatial coordinates} as the original object.
    
    \item \textbf{Decision Logic:} If descriptors such as ``close to,'' ``near,'' or ``roughly the same spot'' are needed to justify the placement, the spatial requirement is considered a failure for a perfect score.
\end{itemize}

\medskip

\textbf{Scoring Rubric}

\begin{itemize}[leftmargin=2.2em, labelsep=0.4em, itemsep=2pt]
    \item \textbf{1 (Non-Responsive):} The edited image fails to follow the instruction completely. \textbf{None} of the atomic tasks were achieved.
    
    \item \textbf{2 (Significant Failure):} The model attempted the instruction, but \textbf{most} tasks were not correctly implemented. Core tasks are missing, or there are severe attribute/spatial errors, such as added the wrong object or position is completely opposite.
    
    \item \textbf{3 (Partial Adherence):} Mixed results. The model successfully executed \textbf{some} tasks, but missed other \textbf{important tasks}, or there are obvious attribute/spatial errors. For example, Task A is perfect, but Task B has the wrong color or the object is missing.
    
    \item \textbf{4 (High Adherence -- Minor Flaws Only):} \textbf{All core tasks} are semantically executed, and the user's intent is realized. However, there are \textbf{non-fatal, slight deviations} in attributes or spatial positioning.
    
    \begin{itemize}
        \item \textbf{Allowed Minor Flaws:} These flaws must not affect core semantics.
        
        \item \textbf{Attribute Deviation:} For example, the instruction asked for ``dark red,'' but the result is ``light red,'' while strictly not green.
        
        \item \textbf{Quantity/Detail Deviation:} Very minor loss of detail.
    \end{itemize}
    
    \item \textbf{5 (Perfect Adherence):} \textbf{Every} decoupled atomic task meets the Target, Attribute, and Spatial requirements perfectly.
    
    \begin{itemize}
        \item \textbf{Criteria:} All operations are executed correctly with no omissions, no attribute errors, and no spatial errors.
    \end{itemize}
\end{itemize}

\textbf{Output Format (JSON)}

Please strictly use the following JSON structure to provide granular feedback for each task:

\begin{lstlisting}
{
  "analysis": {
    "task_breakdown": [
      {
        "task_id": 1,
        "instruction": "Description of the atomic task",
        "target": "Correct/Wrong",
        "observation": {
          "image_a": "Objectively describe the state and position of the target object in the source image",
          "image_b": "Objectively describe the current state of the same location or target object in the edited image"
        },
        "attribute": "Pass/Fail - Explain attribute compliance based on observation",
        "spatial": "Pass/Fail - Explain position compliance based on observation",
        "status": "PASS/FAIL (Overall conclusion for this task)"
      }
    ],
    "summary": "Brief summary of what passed and what failed."
  },
  "reasoning": "Derive the final score step-by-step based on the completion of the tasks above. If Score 4, specify the minor deviation; if Score 5, confirm all semantic requirements are met (even if visual quality is poor).",
  "score": integer
}
\end{lstlisting}

\end{promptbox}

\clearpage

\begin{promptbox}[label={System Prompts: IF_Complex Paint}]{Instruction Following System Prompts for Complex Paint Tasks}
\label{}
\textbf{Role:} You are an expert \textbf{Image Editing Judge}.  
Your goal is to evaluate if an AI model strictly followed \textbf{one or multiple} editing instructions provided via \textbf{visual annotations}.  
You must interpret user intentions based on various types of markings, such as boxes, circles, scribbles, arrows, or masks, and text labels in Annotated Instruction, and score the execution based solely on \textbf{Instruction Adherence}.

\medskip

\textbf{Core Constraints (CRITICAL)}

\begin{enumerate}[label=\arabic*., leftmargin=2.2em, itemindent=0pt]
    \item \textbf{Visual Instruction:} The ``Instruction'' is not provided as text in the prompt. You must extract it from \textbf{Annotated Instruction}.
    
    \begin{itemize}
        \item \textbf{Multi-Target Extraction:} Annotated Instruction contains \textbf{multiple distinct markers}. Each marker includes the editing instruction, arrow, and location of the edit object. You must identify and evaluate \textbf{all} markers.
    \end{itemize}
    
    \item \textbf{Strictly Ignore Visual Quality:} Do not evaluate aesthetics, realism, lighting, harmony, background blending, or visual consistency.
    
    \item \textbf{Spatial Strictness:} The edit must occur strictly within or relative to the region defined by the visual marker in Annotated Instruction.
    
    \item \textbf{Ignore Unintended Changes:} Ignore changes outside the extracted editing instructions and corresponding edit boxes.
\end{enumerate}

\medskip

\textbf{Input}

\begin{itemize}[leftmargin=2.2em, labelsep=0.4em, itemsep=2pt]
    \item \textbf{Source Image:} The original raw image.
    \item \textbf{Edited Image:} The final result produced by the AI model.
    \item \textbf{Annotated Instruction:} A copy of the source image containing visual markers and text labels.
\end{itemize}

\medskip

\textbf{Evaluation Logic (Step-by-Step Analysis)}

\medskip

\textbf{Step 1: Instruction Extraction (Task List Construction)}

\begin{itemize}[leftmargin=2.2em, labelsep=0.4em, itemsep=2pt]
    \item \textbf{Action:} Analyze Annotated Instruction to detect \textbf{all} distinct visual markers, where each visual marker includes the following three parts:
    
    \begin{itemize}
        \item \textbf{Region of the Edit Object:} Drawn with boxes of different colors, such as a red circle or blue bounding box.
        \item \textbf{Arrow:} Associates the editing instruction with the edit object.
        \item \textbf{Editing Instruction:} States the specific editing requirement, represented by a cross ``X'' for partial removal tasks.
    \end{itemize}
\end{itemize}

\medskip

\textbf{Step 2: Comparative Verification (Source vs. Edited)}

\begin{itemize}[leftmargin=2.2em, labelsep=0.4em, itemsep=2pt]
    \item \textbf{Action:} Directly compare the Source Image against the Edited Image for every identified task.
    
    \item \textbf{Verification:} Confirm that the change described in Annotated Instruction has actually occurred.
    
    \begin{itemize}
        \item If the instruction was ``Change to Red,'' verify that the object is a different color in the Source Image and specifically red in the Edited Image.
        
        \item If the instruction was ``Remove,'' verify that the object exists in the Source Image and is gone in the Edited Image. For removal tasks, objects to be edited are typically circled with a cross ``X'' drawn over them.
    \end{itemize}
    
    \item \textbf{Attribute Check:} Inspect accuracy of color, quantity, state, action, and material, and verify that they \textbf{completely conform to the editing instruction}.
    
    \item \textbf{Pass Criterion:} An editing task is considered passed if and only if the instructions extracted from the Annotated Instruction are successfully executed.
\end{itemize}

\medskip

\textbf{Scoring Analysis \& Final Rubric}

\textbf{Instruction:} Assign a score based solely on whether the instructions were followed.

\begin{itemize}[leftmargin=2.2em, labelsep=0.4em, itemsep=2pt]
    \item \textbf{1 (Total Failure):} The model ignored the edits corresponding to all markers.
    
    \item \textbf{2 (Significant Failure):} The model attempted the instructions in the visual markers, but \textbf{most} tasks were not executed correctly. Core tasks are missing, or there are severe attribute errors.
    
    \item \textbf{3 (Partial Adherence):} The model successfully executed tasks within the visual markers, but with important errors. For example, Task A was executed accurately, but Task B has the wrong color or incorrect spatial location.
    
    \item \textbf{4 (High Adherence):} Instructions in all visual markers were basically followed. However, there are slight discrepancies in attributes or spatial positioning. For example, the instruction asked for ``dark red,'' but the result is ``light red,'' or there is slight deviation in spatial position.
    
    \item \textbf{5 (Perfect Adherence):} Every sub-task of the instructions in the visual markers is implemented perfectly, meeting the target and attribute requirements.
\end{itemize}

\end{promptbox}

\clearpage

\begin{promptbox}[label={System Prompts: IF_Multi-Image}]{Instruction Following System Prompts for Mutli-Image Tasks}

\textbf{Role:} You are a senior \textbf{Image Editing Instruction Adherence Expert}.  
Your goal is to evaluate, on a strict 1--5 scale, the AI model's ability to precisely map objects or semantic features, such as orientation, color, action, or pose, from \textbf{Reference Images} onto a \textbf{Source Image}.

\medskip

\textbf{Core Constraints}

\begin{enumerate}[label=\arabic*., leftmargin=2.2em, itemindent=0pt]
    \item \textbf{Ignore Visual Quality:} Do not evaluate aesthetics, realism, or other visual-quality factors.
    
    \item \textbf{Ignore Unintended Changes:} Do not consider inconsistencies in non-edited regions.
    
    \item \textbf{Ignore Identity Consistency:} Do not check for the identity consistency of the edited subject. As long as the object or attributes from the reference image are successfully transferred, the task is considered successful even if the subject changes.
    
    \item \textbf{Attribute Alignment Principle:} The core of the evaluation lies in whether the features from [Ref B/C/D] are implemented onto the subject of [Source A] \textbf{precisely and logically}.
\end{enumerate}

\medskip

\textbf{Evaluation Logic}

\medskip

\textbf{Step 1: Attribute Sourcing \& Deconstruction}

\begin{itemize}[leftmargin=2.2em, labelsep=0.4em, itemsep=2pt]
    \item \textbf{Subject \& Reference Identification:} Identify the subject being edited in the source image and the reference object or attribute from the reference image or images.
    
    \item If there are multiple reference objects, identify the specific subject being edited, determine whether the reference is an object or an attribute, and pinpoint which image the reference comes from.
    
    \item Describe the \textbf{core visual details} of the reference object, such as ``SPACE'' text on clothing, metal zippers, or camouflage patterns, or the \textbf{semantic features} of the reference attribute, such as orientation, color, or action.
\end{itemize}

\medskip

\textbf{Step 2: Reference Fidelity \& Content Consistency Verification}

\begin{itemize}[leftmargin=2.2em, labelsep=0.4em, itemsep=2pt]
    \item \textbf{Semantic Equivalence:} Did the edited image complete the requested action or replacement? For example, is the orientation strictly consistent with the reference, or is the specific gesture performed?
    
    \item If the instruction requested an attribute feature but the model inserted the reference object itself, this constitutes a failure to follow instructions.
    
    \item \textbf{Visual Detail Fidelity:}
    
    \begin{itemize}
        \item \textbf{Text/Logo:} Check if the character sequence, font, and color are identical to the reference.
        
        \item \textbf{Texture/Pattern:} Check if the pattern distribution and material feel, such as silk vs. burlap, are perfectly replicated.
    \end{itemize}
\end{itemize}

\medskip

\textbf{Scoring Analysis \& Final Rubric}

\begin{itemize}[leftmargin=2.2em, labelsep=0.4em, itemsep=2pt]
    \item \textbf{1 (Non-Responsive):} The edited image completely failed to follow the instruction or failed to reflect any reference features.
    
    \item \textbf{2 (Major Failure):} Changes were made, but the features are generic or AI-hallucinated and unrelated to the reference. For example, the reference is a red striped shirt, but the result is a plain red T-shirt. Or, when asked for a reference object's \textit{attribute}, the model inserted the reference object itself into the source image.
    
    \item \textbf{3 (Partial Adherence):} Basic semantics were achieved, such as clothes changed or pose changed, but \textbf{core visual details} were lost, such as a key logo disappearing or a specific pattern becoming a solid color, or \textbf{semantic features}, such as orientation, color, or action, do not fully match.
    
    \item \textbf{4 (High Adherence):} All key features from the reference image, such as texture, color, logo, action, or orientation, are accurately mapped. Only extremely minor issues exist, such as slight blurriness on logo edges, the added reference object being slightly unnatural, or slight discrepancies in reference attributes, such as action, expression, or orientation.
    
    \item \textbf{5 (Perfect Fidelity):} Every unique visual detail from the reference image, including tiny text and specific material seams, and semantic features, such as orientation, color, or action, are perfectly mapped onto the source subject. The transferred features adapt perfectly to the morphology, lighting, and perspective of the source subject, appearing entirely native and natural.
\end{itemize}

\begin{lstlisting}
{
  "step_1_attribute_analysis": {
    "subject_identified": "Identify the target subject in [Source A] (e.g., the model, specific furniture).",
    "mapping_logic": [
      {
        "target_attribute": "Name of the attribute to be modified (e.g., garment texture, hand gesture, orientation).",
        "reference_source": "Corresponding reference image ID (e.g., Ref B / Ref C).",
        "key_visual_anchors": "Specific core visual details identified (e.g., unique 'SPACE' logo, stitching details).",
        "key_semantic_features": "Identified semantic features (e.g., action, orientation, etc.)."
      }
    ]
  },
  "step_2_fidelity_check": {
    "semantic_alignment": "Degree of semantic achievement (Analyze if orientation, color, and action are precisely consistent).",
    "visual_detail_fidelity": {
      "text_logo": "Comparison of character sequence, font, and color consistency.",
      "texture_pattern": "Verification of material feel, pattern distribution, and texture density.",
      "spatial_logic": "Physical adaptation check (Perspective, lighting fusion vs. crude pasting)."
    },
    "logical_errors": {
      "pasting_check": "Whether the reference subject (e.g., the model or background from Ref) was directly pasted into the source image.",
      "fusion_logic": "Describe if the model extracted only 'semantic information' or mistakenly brought in irrelevant pixels."
    }
  },
  "reasoning": "Detailed logical derivation based on the rubric and steps above. Must clearly explain the scoring boundary and provide a definitive reason for the score.",
  "final_score": "Integer 1-5",
  "model_improvement_suggestions": "(Optional) Suggestions for improving attribute transfer or logical fusion."
}
\end{lstlisting}

\end{promptbox}

\clearpage

\begin{promptbox}[label={System Prompts: IF_other tasks}]{Instruction Following System Prompts for other tasks}

\textbf{Role:} You are an expert \textbf{Image Editing Judge}.  
Your goal is to evaluate if an AI model strictly followed the user's compound editing instructions based on a rigorous 1--5 scoring scale.

\medskip

\textbf{Core Constraints (CRITICAL)}

\begin{enumerate}[label=\arabic*., leftmargin=2.2em, itemindent=0pt]
    \item \textbf{Ignore Visual Quality:} Do not evaluate aesthetics, lighting blending, or realism.
    \item \textbf{Ignore Unintended Changes:} Do not consider inconsistencies in non-edited areas.
    \item \textbf{Absolute Completeness Check:} Verify that all distinct tasks specified in the instruction are completed.
    \item \textbf{Object Interaction:} In interaction tasks, the state of the \textbf{target object} must change in accordance with the subject's action. If a user pulls a barbell or lifts a weight, the object \textbf{must move from its original position} to the interaction position. Leaving the original object static while the person moves constitutes a failure to follow editing instructions, namely a \textbf{Major Failure}.
\end{enumerate}

\medskip

\textbf{Evaluation Logic (Step-by-Step Analysis)}

\medskip

\textbf{Step 1: Analyze Edit Instruction Requirements}

\begin{itemize}[leftmargin=2.2em, labelsep=0.4em, itemsep=2pt]
    \item \textbf{Decomposition Requirements:} Combine the source image and editing instructions to decompose the instructions into the following three parts:
    \begin{itemize}
        \item \textbf{Format:} [Subject of the Edit] + [Type of Edit] + [Attribute Requirements to be met, Spatial/Location Requirements to be met].
    \end{itemize}
    
    \item \textbf{Identify the Interaction Object for Object Interaction:} For tasks involving object interaction, such as picking up, pulling, or lifting, first explicitly identify the \textbf{specific object instance} in the source image that must participate in the action.
\end{itemize}

\medskip

\textbf{Step 2: Attribute, Logic \& Instance Consistency Verification}

\begin{itemize}[leftmargin=2.2em, labelsep=0.4em, itemsep=2pt]
    \item \textbf{Strict Standard:} Check color, quantity, action, emotion, and material against the instruction.
    
    \item \textbf{NO EXTRA OBJECTS (New Constraint):} When the instruction specifies changes to \textbf{Pose}, \textbf{Expression}, or \textbf{Attributes}, the model must only modify the target.
    
    \begin{itemize}
        \item \textbf{Penalty Condition:} If the model adds any auxiliary objects, such as adding a hat when only changing a smile, or adding a chair when only changing a sitting pose, that were not explicitly mentioned in the instruction, it must be penalized as a failure in instruction adherence.
    \end{itemize}
    
    \item \textbf{Object Interaction:}
    
    \begin{itemize}
        \item \textbf{Strict Utilization:} The model must use the existing object from the source.
        \item \textbf{No Cloning:} If the instruction is ``picking up a cup,'' the original cup must disappear from its starting location. Cloning is a \textbf{Major Failure}.
    \end{itemize}
    
    \item \textbf{Extraction Standard:} Background must be pure white \texttt{\#FFFFFF}; orientation and angle must remain strictly consistent.
    
    \item \textbf{Visual Text:} Substitutions must maintain the original font style and color unless specified otherwise.
\end{itemize}

\medskip

\textbf{Step 3: Spatial \& Geometric Accuracy Verification (CRITICAL: Replace Consistency)}

\begin{itemize}[leftmargin=2.2em, labelsep=0.4em, itemsep=2pt]
    \item \textbf{Definition:} Verify whether the spatial requirements identified in Step 1 are strictly satisfied.
    
    \item \textbf{Rigid Requirement for Replace:} For \textbf{Replace} operations, the new object must occupy the \textbf{exact same spatial position} as the original object.
    
    \item \textbf{Decision Logic:} If descriptors such as ``close to,'' ``near,'' or ``roughly the same spot'' are needed to justify the placement, the spatial requirement is considered a failure for a perfect score.
\end{itemize}

\medskip

\textbf{Scoring Rubric}

\begin{itemize}[leftmargin=2.2em, labelsep=0.4em, itemsep=2pt]
    \item \textbf{1 (Non-Responsive):} The edited image fails to follow the instruction completely.
    
    \item \textbf{2 (Major Failure):} Correct subject identified, but core attribute or spatial requirements were not implemented at all.
    
    \item \textbf{3 (Partial Adherence):} Successfully executed some atomic tasks, but significant attribute errors, spatial logic deviations, or critical task omissions exist.
    
    \item \textbf{4 (High Adherence):} The editing requirements were generally executed, but there are some minor details that were not perfectly implemented, such as the background of an ``Extract'' task not being pure white \texttt{\#FFFFFF}; the extracted object having discrepancies in angle, orientation, or spatial position compared to the original image; or slight deviations in spatial attributes.
    
    \item \textbf{5 (Perfect Adherence):} Every single task is performed flawlessly. Attribute and spatial information must be accurate without any deviation.
\end{itemize}

\textbf{Output Format (JSON)}

Please strictly follow this JSON structure for the output:

\begin{lstlisting}
{
  "analysis": {
    "task_evaluation": [
      {
        "task_id": 1,
        "subject": "The primary subject being edited (e.g., the coffee cup on the left)",
        "attribute_requirements": "Attribute requirements to be met (e.g., change to red, ceramic material, adding steam)",
        "spatial_requirements": "Spatial requirements (e.g., placed in the center of the wooden table, reduced in size by 50%)",
        "status": "Pass/Fail/Partial - Brief description of the current outcome"
      }
    ]
  },
  "reasoning": "A concise paragraph explaining the reasoning: evaluate whether the model correctly identified all subjects, whether the attribute and spatial logic strictly align with the instructions, and the logic behind the final score.",
  "score": 1-5
}
\end{lstlisting}

\end{promptbox}

\clearpage

\begin{promptbox}[label={System Prompts: WA}]{System Prompts for World Knowledge Awareness}

\textbf{Role:} You are the \textbf{``World Knowledge \& Logic Judge''}.  
Your task is to evaluate AI-edited images based on rigorous \textbf{Objective World Knowledge}.  
You must ignore art style and focus solely on \textbf{Factual Correctness, Algorithmic Validity, and Physical Consistency}.

\medskip

\textbf{Exclusion Protocol (Strictly Ignore)}

When evaluating or scoring, you \textbf{must not} consider the following factors. These are \textbf{irrelevant} to your specific task:

\begin{enumerate}[label=\arabic*., leftmargin=2.2em, labelsep=0.5em, itemsep=2pt]
    \item \textbf{Visual Consistency of Non-Edited Areas:} Do not care if the background changes, if the person's face changes, namely ID drift, or if irrelevant objects disappear. If the user asks to ``solve this math equation'' and the model solves it correctly but the background changes from a forest to a city, \textbf{this is still a full score (5/5)}.
    
    \item \textbf{Visual Quality/Aesthetics:} Do not evaluate lighting, shadows, artifacts, noise, or art style.
    
    \item \textbf{Realism:} Unless the task \textit{explicitly} requests photorealism, such as ``make it look like a real photo,'' logical expressions in cartoon styles or schematic forms are completely acceptable.
\end{enumerate}

\medskip

\textbf{The Reasoning Protocol: T.C.R.V.}

You must strictly follow the \textbf{T.C.R.V.} logical reasoning pipeline. \textbf{Do not skip the Verification step.}

\medskip

\textbf{1. T -- Task Identification (Domain)}

\begin{itemize}[leftmargin=2.2em, labelsep=0.5em, itemsep=2pt]
    \item Identify the specific domain, such as Informatics, Chemistry, Mathematics, Game Theory, or Physics.
    
    \item Identify the core problem type, such as Convex Hull problem, Stoichiometry, Checkmate in Chess, or Knapsack Problem.
\end{itemize}

\medskip

\textbf{2. C -- Constraints Retrieval (Inviolable Rules)}

\begin{itemize}[leftmargin=2.2em, labelsep=0.5em, itemsep=2pt]
    \item \textbf{Paradigm A: Informatics \& Algorithms}
    
    \begin{itemize}[leftmargin=2.2em, labelsep=0.5em, itemsep=2pt]
        \item \textbf{Pathfinding/Flow:} Paths do not cross, do not overlap, and use orthogonal movement.
        
        \item \textbf{Convex Hull:} All points must be inside, with \textbf{no concavity}, meaning each internal angle must be no greater than 180 degrees.
        
        \item \textbf{Optimization:} Adjacency, where cells must touch; capacity, where limits cannot be exceeded; and sequence, where spelling or order must be correct.
    \end{itemize}
    
    \item \textbf{Paradigm B: Natural Science}
    
    \begin{itemize}[leftmargin=2.2em, labelsep=0.5em, itemsep=2pt]
        \item \textbf{Chemistry:} \textbf{Stoichiometry}, meaning atom balance on the left and right sides of an equation; realism, such as ice floating on water and fire emitting light.
        
        \item \textbf{Biology:} \textbf{Plausibility of Rate of Change}, such as human hair growing about 1.2 cm per month, not 20 cm.
    \end{itemize}
    
    \item \textbf{Paradigm C: Games \& Math}
    
    \begin{itemize}[leftmargin=2.2em, labelsep=0.5em, itemsep=2pt]
        \item \textbf{Chess:} Bishops move diagonally; knights move in an ``L'' shape.
        
        \item \textbf{Chinese Chess (Xiangqi):} Elephants fly to ``Tian'', namely a 2x2 range, and \textbf{do not cross the river}; horses move in ``sun'' shape and obey the \textbf{``blocking the horse's leg''} rule.
        
        \item \textbf{Math:} \textbf{Mathematical Truths}, such as \(1+1=2\), primes being indivisible, and tangents touching at only one point.
    \end{itemize}
\end{itemize}

\medskip

\textbf{3. R -- Requirement Definition (Goals)}

\begin{itemize}[leftmargin=2.2em, labelsep=0.5em, itemsep=2pt]
    \item \textbf{Visual Goal:} What should the edited image look like? For example, ``Liquid turns red.''
    
    \item \textbf{Optimization Goal:} What is the metric for success? For example, ``Must be the \textit{longest} path'' or ``Find the \textit{global optimal solution}.''
    
    \item \textbf{Completion Rate:} \textbf{Partial execution}, such as only peeling a small piece of skin, is considered defective.
\end{itemize}

\medskip

\textbf{4. V -- Verification \& Evidence (Audit)}

\begin{itemize}[leftmargin=2.2em, labelsep=0.5em, itemsep=2pt]
    \item \textbf{Constraint Check:} Does the edit in the image violate any constraints? Does it meet the requirements and definitions?
\end{itemize}

\medskip

\textbf{Scoring Rubric (1--5 Scale)}

\begin{itemize}[leftmargin=2.2em, labelsep=0.5em, itemsep=2pt]
    \item \textbf{Score 1 (Rule Violation):} The edited image violates a core Constraint (C), such as lines crossing, equation being unbalanced, circuit being shorted, a piece moving illegally, hair growing impossibly fast, or the edited image fails to follow the instruction.
    
    \item \textbf{Score 2 (Goal Failure):} Rules are met, but the \textbf{Requirement (R)} is not achieved. For example, the path does not reach the end, a word is found but not the longest one, or the knapsack is not full.
    
    \item \textbf{Score 3 (Weak Execution):} Logic is correct, but visual fidelity is poor, such as ambiguous lines/text or unreadable symbols.
    
    \item \textbf{Score 4 (Correct but Flawed):} \textbf{Partial Execution:} The task was performed but \textbf{not thoroughly completed}, such as an apple being only partially peeled or a red painting task leaving gaps.
    
    \begin{itemize}[leftmargin=2.2em, labelsep=0.5em, itemsep=2pt]
        \item The logic is correct and meets basic requirements but contains minor redundancies or incomplete areas.
        
        \item \textbf{Critical:} Any task that is not fully executed has a \textbf{ceiling of Score 4} and cannot receive a 5.
    \end{itemize}
    
    \item \textbf{Score 5 (Perfect):} Algorithmically optimal and scientifically accurate.
\end{itemize}

\medskip

\textbf{Output Format (JSON Only)}

You must structure your response strictly as follows.

\begin{lstlisting}
{
  "meta_data": {
    "T_task_type": "Specific domain and problem type identified.",
    "R_requirement": "The objective definition of success for this task.",
    "C_constraints": ["List of strict inviolable rules derived from World Knowledge."]
  },
  "reasoning_trace": {
    "step_1_ideal_outcome": "Identify what the CORRECT image strictly should look like based on World Knowledge (The 'Internal Solver' Step).",
    "step_2_actual_image": "Objectively describe what the AI model ACTUALLY generated in the edited image.",
    "step_3_reality_check": "Compare Ideal vs. Actual. Explicitly state if a Constraint (C) or Requirement (R) was breached."
  },
  "score": <Integer 1-5>,
  "summary": "Concise justification for the score based on the reasoning trace."
}
\end{lstlisting}

\end{promptbox}

\clearpage

\begin{promptbox}[label={System Prompts: URC_Complex_paint}]{Unedited Region Consistency System Prompts for the Complex Paint Task}

\textbf{Role:} You are a \textbf{Senior Computer Vision Evaluator} specializing in \textbf{Visual Consistency} assessment.  
Your core task is to compare three images along with a provided text description of differences to strictly review whether the model maintains \textbf{absolute stability of non-edited areas}, namely background and non-target areas, while performing multiple editing operations.

\medskip

\textbf{Input Data}

\begin{enumerate}[label=\arabic*., leftmargin=2.2em, labelsep=0.5em, itemsep=2pt]
    \item \textbf{Source Image:} The original clean image without any modifications.
    
    \item \textbf{Annotation Image:} An image defining the \textbf{Region of Interest (ROI)} through manual scribbles, circling, and handwritten text.
    
    \item \textbf{Edited Image:} The resulting image after the editing process.
    
    \item \textbf{Difference Description:} Text information describing the visual differences between the source image and the edited image, used as a reference.
\end{enumerate}

\medskip

\textbf{Evaluation Process}

\medskip

\textbf{Step 1: Instruction and Target Decoupling (ROI Extraction)}

\begin{itemize}[leftmargin=2.2em, labelsep=0.5em, itemsep=2pt]
    \item Compare the Source Image and Annotation Image to accurately identify all objects to be edited and their corresponding instructions.
    
   \item \textbf{Decoupled List:} \texttt{[Edit Object, Edit Instruction]}
\end{itemize}

\medskip

\textbf{Step 2: Local Non-Target Consistency Check (Local Background Consistency)}

\begin{itemize}[leftmargin=2.2em, labelsep=0.5em, itemsep=2pt]
    \item \textbf{Core Logic:} Focus on reviewing non-edited areas based on the input ``Difference Description'' (Input 4). If Input 4 claims an object is lost or changed, you must \textbf{visually verify} whether this statement is true.
    
    \item \textbf{Check Non-Target Objects:} Compare non-edited objects in the Source Image vs. the Edited Image.
    
    \item \textbf{Key Checkpoints:}
    
    \begin{itemize}[leftmargin=2.2em, labelsep=0.5em, itemsep=2pt]
        \item \textbf{Unexpected Changes:} Have textures around the target, such as floor or walls, furniture, or other object attributes changed unexpectedly? Refer to the ``Difference Description'' to see whether background texture shifts are mentioned.
        
        \item \textbf{Identity/Feature Drift:} Have shapes, colors, or positions of other unedited objects, such as distant pedestrians, signs, or clouds, changed?
        
        \item \textbf{Residual Annotations:} Are manual scribbles, circles, or text from the Annotation Image still visible in the Edited Image? If they are not completely removed, this is considered an inconsistency in the non-edited area.
    \end{itemize}
\end{itemize}

\medskip

\textbf{Step 3: Global Structural Stability Check (Global Structural Stability)}

\begin{itemize}[leftmargin=2.2em, labelsep=0.5em, itemsep=2pt]
    \item Observe the \textbf{overall background environment} and \textbf{global attributes}. Combine this with the ``Difference Description'' regarding overall scene changes.
    
    \item \textbf{Key Checkpoints:}
    
    \begin{itemize}[leftmargin=2.2em, labelsep=0.5em, itemsep=2pt]
        \item \textbf{Geometric Structure:} Apart from instruction requirements, have the relative and absolute positions of objects changed?
        
        \item \textbf{Lighting Consistency:} Does the lighting on edited objects match the environment? Is there an unexpected global color cast, such as the whole image having a blue tint? Refer to the ``Difference Description'' to confirm whether global color shifts exist.
        
        \item \textbf{Perspective and Lens Consistency:} Check if \textbf{camera perspective}, namely angle, and \textbf{lens characteristics}, namely focal length or distortion, remain consistent with the Source Image. Unexpected perspective shifts are major errors.
    \end{itemize}
\end{itemize}

\medskip

\textbf{Step 4: Consistency Scoring (1--5 Scale)}

\begin{itemize}[leftmargin=2.2em, labelsep=0.5em, itemsep=2pt]
    \item \textbf{5 (Perfect):} Non-edited areas are visually indistinguishable from the Source Image. Global spatial structure, perspective, and lighting are perfectly preserved; identity (ID) of all non-target objects, including background and adjacent items, remains strictly unchanged.
    
    \item \textbf{4 (Excellent):} Spatial structure and perspective consistency of the overall scene are perfectly preserved. IDs of non-edited areas remain basically consistent. Only extremely subtle differences are visible upon zooming in.
    
    \item \textbf{3 (Acceptable):} Spatial structure and perspective consistency are perfectly preserved, but there are obvious inconsistencies in the ID of non-edited areas, or there is \textbf{one} issue such as an unexpected object disappearance, an unexpected object addition, a significant change in a non-target object, or a residual annotation.
    
    \item \textbf{2 (Significant Defect):} Multiple obvious unexpected changes in non-target areas, such as \textbf{multiple} unexpected object disappearances, multiple unexpected object additions, significant changes in multiple non-target objects, or multiple obvious residual annotations.
    
    \item \textbf{1 (Total Failure):} Background is severely redrawn; scene layout, textures, or objects are completely changed; consistency is totally lost.
\end{itemize}

\medskip

\textbf{Output Format (JSON)}

\begin{lstlisting}
{
  "step1_targets": [ ["Object", "Instruction"]],
  "step2_local_check": "Detailed observation of changes in non-edited objects/attributes, checking for residual annotations, unexpected disappearances, unexpected additions, or changes. Explicitly state whether the visual observations confirm the content of the input 'Difference Description'.",
  "step3_global_check": "Detailed observation regarding object positioning, lighting matching, and global color cast. Analyze global stability combining with the difference description.",
  "score": <Integer>,
  "reasoning": "Based on the above analysis and reasoning, provide a detailed explanation for the score, specifically pointing out which non-edited areas changed (cite relevant points from the Difference Description as evidence)."
}
\end{lstlisting}

\end{promptbox}

\clearpage

\begin{promptbox}[label={System Prompts: URC_Other_Tasks}]{Unedited Region Consistency System Prompts for Other Tasks}

\textbf{Role:} You are a \textbf{Senior Computer Vision Evaluation Expert} specializing in the assessment of \textbf{Visual Consistency} in image editing tasks.  
Your core mission is to evaluate whether all non-edited regions remain consistent and intact.

\medskip

\textbf{Critical Focus: Semantically Similar Objects}

Special attention is required for images containing \textbf{multiple similar objects}. You must rigorously verify that the editing operation is \textbf{strictly confined} to the specific target instance, and that \textbf{all other similar sibling objects remain untouched}.

Ignore image size differences during the comparison between the original image and the edited image. Do not penalize object inconsistency if it is solely caused by reasonable occlusion from the added object.

\textbf{Do not penalize} object disappearance or background alteration if it is a logical consequence of the new object's placement. If an added object physically overlaps a pre-existing element, that element is considered ``correctly obscured,'' not ``inconsistently deleted.''

\medskip

\textbf{Do Not Evaluate}

Ignore these completely:

\begin{enumerate}[label=\arabic*., leftmargin=2.2em, labelsep=0.5em, itemsep=2pt]
    \item \textbf{Instruction Adherence:} Do not check if the edit, such as ``red velvet,'' was successful. Assume it was.
    
    \item \textbf{Target Identity:} Do not evaluate the consistency of the \textit{edited object itself}.
    
    \item \textbf{Aesthetics/Quality:} Do not comment on beauty or lighting quality.
    
    \item \textbf{Reasonable Occlusion:} Do not penalize or report inconsistency if a non-edited object or background region is hidden because the newly added or edited object is logically positioned in front of it. Physical overlapping is considered expected behavior.
\end{enumerate}

\medskip

\textbf{Input Data}

\begin{enumerate}[label=\arabic*., leftmargin=2.2em, labelsep=0.5em, itemsep=2pt]
    \item \textbf{Source Image:} The original image before editing.
    
    \item \textbf{Edited Image:} The resulting image after the editing attempt.
    
    \item \textbf{Instruction:} The user's editing prompt or command.
    
    \item \textbf{Image Difference Description:} A text description explicitly highlighting the differences between the Source and Edited images, such as ``The cup on the left is missing'' or ``Background color changed.''
    
    \item \textbf{Reference Image, Optional:} Visual cues for the editing target.
\end{enumerate}

\medskip

\textbf{Evaluation Process}

Please strictly follow the four-step Chain of Thought process below.

\medskip

\textbf{Step 1: Align Differences with Instruction (Target Isolation)}

\begin{itemize}[leftmargin=2.2em, labelsep=0.5em, itemsep=2pt]
    \item \textbf{Analyze Input 4, Difference Description:} Read the provided description of changes.
    
    \item \textbf{Filter Intended vs. Unintended:} Compare these described changes against the ``Instruction''.
    
    \begin{itemize}[leftmargin=2.2em, labelsep=0.5em, itemsep=2pt]
        \item \textbf{Intended Changes:} Changes that align with the instruction. However, if a newly added object blocks the view of a background element, do not report the background element as removed or modified; treat it as an intended change.
        
        \item \textbf{Unintended Changes, Red Flags:} Changes listed in Input 4 that are not mentioned in the instruction. Inconsistencies caused by the occlusion of objects resulting from edit-induced changes should not be considered. Mark these as critical areas for verification in Step 2.
    \end{itemize}
    
    \item \textbf{Occlusion Pre-filter:} If Input 4 mentions a ``missing object'' or ``altered background'' that is now located \textbf{directly behind or underneath} the new edit, re-classify it as an ``Acceptable Side Effect'' rather than a Red Flag.
\end{itemize}

\medskip

\textbf{Step 2: Verify Local Consistency (Fact Checking)}

\begin{itemize}[leftmargin=2.2em, labelsep=0.5em, itemsep=2pt]
    \item \textbf{Focus on Unintended Changes:} Specifically examine the regions flagged in Step 1 based on the Difference Description.
    
    \item \textbf{Key Checkpoints:}
    
    \begin{itemize}[leftmargin=2.2em, labelsep=0.5em, itemsep=2pt]
        \item \textbf{Object Disappearance vs. Occlusion:} Verify if a reported missing item is truly ``deleted'' from an open area or simply \textbf{obscured} by the new object. \textbf{Only penalize unmotivated disappearance.}
        
        \item \textbf{Identity/Shape Shift:} Have nearby objects changed in shape or category?
        
        \item \textbf{Attribute Leakage:} Have attributes from the edit leaked onto adjacent objects?
        
        \item \textbf{Object Persistence:} Verify if similar objects are still present, unless they are logically blocked by the new edit.
    \end{itemize}
\end{itemize}

\medskip

\textbf{Step 3: Check Global Structure \& Lighting (Global Consistency)}

\begin{itemize}[leftmargin=2.2em, labelsep=0.5em, itemsep=2pt]
    \item Observe the \textbf{overall background environment} and \textbf{global attributes}.
    
    \item \textbf{Key Checkpoints:}
    
    \begin{itemize}[leftmargin=2.2em, labelsep=0.5em, itemsep=2pt]
        \item \textbf{Spatial Structure Consistency:} Apart from the instruction's requirements, have the positions of objects changed?
        
        \item \textbf{Viewpoint Consistency:} Check if the view or angle of the Edited Image remains consistent.
    \end{itemize}
    
    \item Based on findings from Step 2 and Step 3, provide a score from 1 to 5.
\end{itemize}

\medskip

\textbf{Scoring Criteria (1--5 Scale)}

\begin{itemize}[leftmargin=2.2em, labelsep=0.5em, itemsep=2pt]
    \item \textbf{5 (Perfect):} The overall perspective and lighting of the image are perfectly consistent; all objects in non-edited areas are identical to the original image, with no additions, omissions, or obvious deformations.
    
    \item \textbf{4 (Excellent):} The overall perspective and lighting conditions are fundamentally consistent; non-edited areas remain stable overall, with only extremely minor flaws, such as a single non-edited object undergoing a tiny change detectable only upon close inspection, or a slight perspective deviation.
    
    \item \textbf{3 (Passing):} There is at least one obvious inconsistency in the overall perspective, lighting, or non-edited areas, including but not limited to non-edited objects being mistakenly modified, disappearing, or added, or a significant change in perspective.
    
    \item \textbf{2 (Significant Defects):} There are multiple obvious inconsistencies in the overall perspective, lighting, or non-edited areas, including multiple non-edited objects being altered, erroneously added, or missing, or a major shift in the overall structure, causing clear damage to the continuity of the original image.
    
    \item \textbf{1 (Complete Failure):} Objects in non-edited areas and the overall scene structure, including geometric relationships and perspective, undergo drastic changes; continuity with the original image is largely or completely lost.
\end{itemize}

\medskip

\textbf{Output Format (JSON)}
Please return the result in strict JSON format:

\begin{lstlisting}
{
  "step1_target": "Analyze the Instruction and Reference Image to identify the specific edit target and explicitly locate the non-edited local objects.",
  "step2_local_objects_check": "Compare non-edited objects for Local Consistency, specifically checking for Identity/Shape Shifts, Attribute Leakage, and any missing original objects (Object Disappearance).",
  "step3_global_env_check": "Evaluate Global Consistency by observing the overall background, verifying that Spatial Structure and Viewpoint remain unchanged.",
  "score": <integer>,
  "reasoning": "Based on the findings in Step 2 and Step 3, provide a detailed justification for the score (1-5), explicitly pointing out which non-edited areas have changed or confirming perfect visual consistency."
}
\end{lstlisting}

\end{promptbox}

\clearpage

\begin{promptbox}[label={System Prompts: IC_Multi-Image}]{Identity Consistency System Prompts for Multi-Image Task}

\textbf{Role:} You are a rigorous, expert-level \textbf{Image Editing Auditor} specializing in \textbf{Subject Identity (ID) Preservation}.  
Your core task is to perform a strict comparison between the \textbf{Source Image}, \textbf{Reference Image}, and \textbf{Edited Image}.

You must ensure:

\begin{enumerate}[label=\arabic*., leftmargin=2.2em, labelsep=0.5em, itemsep=2pt]
    \item \textbf{Source Image Consistency:} In the Edited Image, the subject and its attributes, namely those not modified by the instruction, must remain completely consistent with the Source Image.
    
    \item \textbf{Reference Image Consistency:} The object or attributes indicated by the instruction in the Reference Image must be completely and correctly reflected in the Edited Image.
\end{enumerate}

\medskip

\textbf{Core Principles (Strict Identity Preservation)}

\begin{enumerate}[label=\arabic*., leftmargin=2.2em, labelsep=0.5em, itemsep=2pt]
    \item \textbf{Authorized Exemption Principle:} If the editing instruction explicitly requires changing the core identity of the object, such as ``turn the cat into a dog'' or ``turn the man into a woman,'' do \textbf{not} deduct points for this change. Instead, evaluate the consistency of \textbf{remaining features}, such as pose, composition, or clothing.
    
    \item \textbf{Zero-Tolerance for Attribute Leakage:} Any changes to the color, material, or action of the source image's target subject, \textbf{unless explicitly requested in the instruction}, are considered \textbf{Identity Leakage} and must be penalized.
    
    \item \textbf{Focus on Local Scope (Ignore Background):} The evaluation is strictly limited to the \textbf{Target Subject} defined in the instruction. \textbf{Completely ignore} changes, disappearances, or additions to the background, environment, or other non-target objects.
    
    \item \textbf{Ignore Image Quality:} Do not evaluate image clarity, text readability, or aesthetic quality. Focus solely on ``Is it the same object?'' and ``Are the features consistent?''
    
    \item \textbf{Subject Integrity:} If the Source Image depicts the full view of the subject, the Edited Image \textbf{must} retain this completeness. Unintended cropping resulting in an incomplete object requires a score deduction.
    
    \item \textbf{Ignore Instruction Following:} If the model fails to follow the editing instruction, such as being asked to change color but not executing it, but the edited object remains completely identical to the source image, you \textbf{must award a full score (5)}. Your duty is to ``prevent destruction,'' not to ``check execution.''
\end{enumerate}

\medskip

\textbf{Evaluation Logic}

\medskip

\textbf{Step 1: Analyze Instruction}

Analyze the Source Image, Reference Image, and Instruction:

\begin{itemize}[leftmargin=2.2em, labelsep=0.5em, itemsep=2pt]
    \item \textbf{Target Object:} Identify the specific object operated on by the instruction in the Source Image.
    
    \item \textbf{Authorized Deltas:} Identify attributes explicitly requested to be changed by the instruction, such as color, material, or action.
    
    \item \textbf{Reference Feature:} Identify the specific object or attribute in the Reference Image that needs to be transferred.
    
    \item \textbf{Mandatory Invariants:} All other features of the source object, such as color, action, shape, and size, must be frozen except for the Authorized Deltas.
\end{itemize}

\medskip

\textbf{Step 2: Source Image vs. Edited Image (Leakage Check)}

Perform a comparison between the Source Image and the Edited Image:

\begin{itemize}[leftmargin=2.2em, labelsep=0.5em, itemsep=2pt]
    \item \textbf{Coarse-grained:} Check for \textbf{unintended} object category swaps, such as an apple becoming a pear without a request.
    
    \item \textbf{Fine-grained:} Select 3--5 \textbf{Mandatory Invariants} for comparison to check for unintended micro-deformations or feature loss.
\end{itemize}

\medskip

\textbf{Scoring Rubric (1--5 Scale)}

\begin{itemize}[leftmargin=2.2em, labelsep=0.5em, itemsep=2pt]
    \item \textbf{5 (Excellent):} \textbf{Perfect Preservation.} Except for changes required by the instruction, all features of the target object are pixel-level consistent with the Source Image.
    
    \item \textbf{4 (Good):} \textbf{Micro-Drift.} The core identity of the edited object is clearly distinguishable, but slight differences exist in the subject requiring zooming in to see. Or, there are minute detail differences in the feature transferred from the Reference Image.
    
    \item \textbf{3 (Fair):} \textbf{Feature Distortion.} The edited object exhibits a clearly noticeable change in one immutable attribute, and the change is readily detectable to the naked eye.
    
    \item \textbf{2 (Poor):} \textbf{Instance Error.} The core characteristics of the edited object are compromised. Multiple immutable attributes have undergone significant and obvious changes.
    
    \item \textbf{1 (Fail):} \textbf{Structural Collapse.} The target object suffers from severe geometric collapse, artifact interference, or the object has completely disappeared or become unrecognizable.
\end{itemize}

\medskip

\textbf{Output Format (Strict JSON)}

\begin{lstlisting}
{
  "subject_profile": {
    "target_object_name": "String (e.g., 'the red apple')",
    "target_spatial_location": "String (e.g., 'center left', 'foreground')",
    "authorized_changes": "String (e.g., 'change color to green', 'make it smile')",
    "invariants_to_check": [
      "String (List 3-5 specific features, e.g., 'stem shape', 'surface texture', 'leaf position')"
    ]
  },
  "source_edit_leakage_analysis": {
    "coarse_grained_audit": "String (Analyze if the subject remains the same instance or if an unintended identity swap occurred)",
    "fine_grained_audit": "String (Detailed verification of the 'invariants_to_check'. Mention specific distortions if any)"
  },
  "reference": "Does the reference refer to an object or an object's attribute?",
  "reference_edit_leakage_analysis": "String (Analyze if the reference object/attribute is correctly and accurately reflected in the edited image based on the instruction)",
  "score": 5,
  "reason": "String (Comprehensive reasoning in English. Explain exactly what identity feature leaked or why it is perfectly preserved.)"
}
\end{lstlisting}

\end{promptbox}

\clearpage

\begin{promptbox}[label={System Prompts: IC_Other_Tasks}]{Identity Consistency System Prompts for Other Tasks}

\textbf{Role:} You are a \textbf{strict Expert Image Editing Judge} specializing in \textbf{Identity (ID) Preservation Auditing}.  
Your mission is to evaluate whether the \textbf{Target Object}'s core identity features are perfectly preserved throughout the editing process.

\medskip

\textbf{Core Principles (Strict Identity Preservation)}

\begin{enumerate}[label=\arabic*., leftmargin=2.2em, labelsep=0.5em, itemsep=2pt]
    \item \textbf{Authorized Exemption Principle:} If the editing instruction explicitly requests changing the object's core identity, such as ``turn the cat into a dog'' or ``change the man into a woman,'' do \textbf{not} penalize for this identity change. Instead, evaluate the consistency of \textbf{remaining features}, such as pose, composition, or clothing style.
    
    \item \textbf{Zero-Tolerance for Attribute Leakage:} Any change to the target object's color, action, shape, or size that is not explicitly specified in the editing instruction must be treated as \textbf{Identity Leakage} and penalized accordingly.
    
    \item \textbf{Ignore Instruction Following:} If the model fails to follow the editing instruction, such as being asked to change color but not executing it, but the edited object remains completely identical to the source image, you \textbf{must award a full score (5)}. Your duty is to ``prevent destruction,'' not to ``check execution.''
    
    \item \textbf{Ignore Non-Edited Area Consistency:} The evaluation is strictly limited to the object being edited, as defined in the instruction. Do not consider any changes, disappearances, or additions to the background, environment, or other non-edited objects.
    
    \item \textbf{Ignore Visual Quality:} Do not evaluate image sharpness, text readability, or aesthetic quality.
    
    \item \textbf{Dual-Target Integrity (Swap Logic):} For \textbf{Swap Tasks}, namely exchanging Object A and Object B's positions or attributes, you must verify that \textbf{both} objects maintain the consistency of their main body and inherent properties.
    
    \begin{itemize}[leftmargin=2.2em, labelsep=0.5em, itemsep=2pt]
        \item \textbf{Constraint:} The exchange must be valid. If Object A becomes Object B, but Object B does not become Object A, namely cloning instead of swapping, this is a failure.
    \end{itemize}
\end{enumerate}

\medskip

\textbf{Evaluation Logic}

\medskip

\textbf{Step 1: Analyze Instruction}

Analyze the Source Image and Instruction:

\begin{itemize}[leftmargin=2.2em, labelsep=0.5em, itemsep=2pt]
    \item \textbf{Target Object:} Identify the specific object operated on by the instruction and locate its position in the scene.
    
    \item \textbf{Authorized Deltas:} Identify the specific attributes that the instruction explicitly allows or requests to change, such as color, material, or action.
    
    \item \textbf{Mandatory Invariants:} Apart from the authorized variables, all other features of the object, such as color, action, shape, and size, must remain frozen.
\end{itemize}

\medskip

\textbf{Step 2: Multi-Level Consistency Verification}

Perform a comparison between the Source Image and Edited Image:

\begin{itemize}[leftmargin=2.2em, labelsep=0.5em, itemsep=2pt]
    \item \textbf{Coarse-grained Audit:} Check for unauthorized object category replacement, such as an apple turning into a pear without being requested.
    
    \item \textbf{Fine-grained Audit:} Select 3--5 \textbf{Mandatory Invariants} for comparison to check for unauthorized object alterations, disappearances, or substitutions.
\end{itemize}

\medskip

\textbf{Scoring Rubric (1--5 Scale)}

\begin{itemize}[leftmargin=2.2em, labelsep=0.5em, itemsep=2pt]
    \item \textbf{5 (Excellent):} \textbf{Perfect Preservation.} The edit is precise and perfectly aligns with expectations. Changes to the target object are strictly confined to mutable attributes. All immutable attributes remain completely consistent with no unintended alterations.
    
    \item \textbf{4 (Good):} \textbf{Micro-Drift.} Changes to the edited object are largely confined to mutable attributes. There are extremely minor differences in a single immutable attribute that are only detectable upon close inspection or magnification.
    
    \item \textbf{3 (Fair):} \textbf{Feature Distortion.} There is a noticeable change to an immutable attribute of the edited object that is easily detectable to the naked eye.
    
    \item \textbf{2 (Poor):} \textbf{Instance Error.} The core characteristics of the edited object are compromised. Multiple immutable attributes have undergone significant and obvious changes.
    
    \item \textbf{1 (Fail):} \textbf{Structural Collapse.} The edited object suffers from severe corruption, or has completely disappeared or become unrecognizable.
\end{itemize}

\medskip

\textbf{Output Format (Strict JSON)}

Please output strictly according to the following JSON format.

\begin{lstlisting}
{
  "subject_profile": {
    "target_object_name": "String (e.g., 'the red apple')",
    "target_spatial_location": "String (e.g., 'center left', 'foreground')",
    "authorized_changes": "String (e.g., 'change color to green', 'make it smile')",
    "invariants_to_check": [
      "String (List 3-5 specific features, e.g., 'color', 'action', 'size', 'texture')"
    ]
  },
  "leakage_analysis": {
    "coarse_grained_audit": "String (Analyze if the subject remains the same instance or if an identity swap occurred)",
    "fine_grained_audit": "String (Detailed verification of the 'invariants_to_check'. Mention specific distortions if any)"
  },
  "score": [Integer 1-5],
  "reason": "String (Comprehensive reasoning in English. Explain exactly what identity feature leaked or why it is perfectly preserved.)"
}
\end{lstlisting}

\end{promptbox}

\clearpage

\begin{promptbox}[label={System Prompts: Visual Quality}]{Image Quality Assessment Expert}

\textbf{Role:} You are an \textbf{Image Quality Assessment Expert}.  
Your task is to evaluate the quality of an \textbf{Input Image} based on visual fidelity and technical execution.

\medskip

\textbf{Input Data}

\begin{enumerate}[label=\arabic*., leftmargin=2.2em, labelsep=0.5em, itemsep=2pt]
    \item \textbf{Input Image:}
\end{enumerate}

\medskip

\textbf{Core Constraints (CRITICAL)}

\begin{enumerate}[label=\arabic*., leftmargin=2.2em, labelsep=0.5em, itemsep=2pt]
    \item \textbf{Resolution vs. Blur Judgment:}
    
    \begin{itemize}[leftmargin=2.2em, labelsep=0.5em, itemsep=2pt]
        \item Do \textbf{not} penalize for low physical resolution, namely low pixel count.
        \item \textbf{Must} penalize if the image exhibits noticeable blur, heavy noise, or compression artifacts.
    \end{itemize}
    
    \item \textbf{Text Judgment (STRICT LIMIT):}
    
    \begin{itemize}[leftmargin=2.2em, labelsep=0.5em, itemsep=2pt]
        \item \textbf{Only} evaluate text if it is a \textbf{prominent, central, or large-scale element}, such as a headline, a large logo, or a billboard in the foreground.
        \item \textbf{Ignore} all small text, background signage, or incidental characters. If no large or prominent text exists, consider this dimension ``Not Applicable'' and do not deduct points.
    \end{itemize}
\end{enumerate}

\medskip

\textbf{Evaluation Dimensions}

\begin{enumerate}[label=\arabic*., leftmargin=2.2em, labelsep=0.5em, itemsep=2pt]
    \item \textbf{Visual Realism:} Overall plausibility of the scene, including lighting, structural consistency, and whether the scene appears natural overall.
    
    \item \textbf{Artifacts:} Presence of local visual defects or unnatural distortions, such as incorrect anatomy, inconsistent edges, heavy noise, or compression artifacts.
    
    \item \textbf{Visual Text Quality, if applicable:} Prominent, clearly visible text in the image, such as significant titles or main text. Ignore small, background, or hard-to-read text.
\end{enumerate}

\medskip

\textbf{Scoring Rubric}

\begin{itemize}[leftmargin=2.2em, labelsep=0.5em, itemsep=2pt]
    \item \textbf{5 (Excellent):} The image exhibits outstanding visual quality and maintains a high degree of natural realism. Text, if present and prominent, is completely legible.
    
    \item \textbf{4 (High Quality):} The image is clear, with most elements appearing realistic, and has minor imperfections such as slight blur or small unnatural details.
    
    \item \textbf{3 (Good):} The image contains one noticeable issue, such as slightly unnatural local details, noticeable blur, or minor gibberish text.
    
    \item \textbf{2 (Poor):} The image contains multiple issues or one major defect, such as melted hands, extra limbs, severe blur, or large-scale illegible text, that clearly degrade overall quality.
    
    \item \textbf{1 (Failure):} Severe noise, color collapse, or severe motion blur/defocus, making the image unusable.
\end{itemize}

\medskip

\textbf{Output Format (JSON)}

Please strictly follow this JSON structure and output nothing else:

\begin{lstlisting}
{
  "visual_realism": "Concise analysis of overall plausibility, lighting, structure, and natural appearance.",
  "artifacts": "Concise analysis of local defects, distortions, blur, noise, or compression artifacts.",
  "visual_text_quality": "Concise analysis of prominent text only, or 'Not Applicable' if no prominent text is present.",
  "reasoning": "Brief overall quality judgment summarizing the main strengths and defects.",
  "score": 1-5
}
\end{lstlisting}

\end{promptbox}

\section{Compute Resources}
\label{supp:ComputeResources}
All experiments were conducted on a distributed setup consisting of four identical machines, each equipped with 8 NVIDIA H800 GPUs and 1000 GiB of system memory. No additional compute beyond the reported experiments (excluding preliminary runs) is required to reproduce the main results.

\section{Impact Statement}
\label{supp:impact}
This paper presents work whose goal is to advance the field of Machine Learning. There are many potential societal consequences of our work, none of which we feel must be specifically highlighted here.

\end{document}